%% file: main.tex
\pdfoutput=1

\documentclass[11pt]{article}

\usepackage[final]{acl}
\usepackage{times}
\usepackage{latexsym}

\usepackage[T1]{fontenc}
\usepackage[utf8]{inputenc}

\usepackage{microtype}

\usepackage{inconsolata}

\usepackage{graphicx}

\input{math_commands.tex}
\usepackage{hyperref}

\usepackage{amsmath}
\usepackage{amssymb}
\usepackage{mathtools}
\usepackage{amsthm}
\newcommand{\highlight}[1]{\cellcolor{green!10}{#1}}

\input{packs}
\input{defs}

\usepackage[textsize=tiny]{todonotes}
\usepackage{colortbl}

\newif\ifshowedits
\showeditstrue  %

\title{TiC-LM: A Web-Scale Benchmark for Time-Continual LLM Pretraining}

\author{Jeffrey Li$^{1*\dagger}$\And
Mohammadreza Armandpour$^{2\dagger}$\And
Iman Mirzadeh$^2$\AND
Sachin Mehta$^{\circ}$\And %
Vaishaal Shankar$^{\circ}$\And
Raviteja Vemulapalli$^2$\And
Samy Bengio$^2$\AND
Oncel Tuzel$^2$\And
Mehrdad Farajtabar$^2$\And
Hadi Pouransari$^2$\And
Fartash Faghri$^{2\ddagger}$\AND
\vspace*{-20pt}\\
$^1$University of Washington, $^2$Apple\\ %
\texttt{jwl2162@cs.washington.edu},\texttt{fartash@apple.com}\\
{\footnotesize
$^*$Work done during an internship at Apple.
$^{\circ}$Work done while at Apple.
$^\dagger$Equal contribution
$^\ddagger$Project lead
}
}

\begin{document}
\maketitle

\input{sec/abstract}
\input{sec/intro_v2}

\input{sec/related}
\input{sec/dataset}
\input{sec/evals}
\input{sec/method}
\input{sec/experiments}
\input{sec/conclusion}

\section*{Limitations}
\vspace*{-3pt}

An important limitation of our work is that we were not able to find a method that outperfroms Oracle re-training on all evaluations. To close these remaining gaps, we hope that future efforts can use our assets to explore more creative ways to balance forgetting and plasticity. Some potentially promising directions could be, but are not limited to: (1) \textit{adaptive} strategies for determining replay ratios; (2) adding timestamps to training documents as in \citet{rosin2022time, dhingra2022time}; (3) time-aware \textit{data-centric} interventions such as deduplicating/filtering replay buffers to keep only the most unique/useful knowledge from older dumps.

Another set of limitations relates to our benchmark design and bringing it closer to current common practices for LLM pre-training. First, we focus on general web-data but LLMs are typically trained on a mix of other separately curated sources (e.g. Wikipedia, ArXiv, etc.). It would thus be interesting to expand our training setup to include time-stratified versions of such sources. Another practical consideration is related to the evolution of \textit{tokenizers} over time. Our benchmark fixes the tokenizer across all timesteps and methods but it would be interesting to explore whether it would be beneficial to co-adapt models and tokenizers as language changes over time. Finally, our current dynamic evaluations were limited to perplexity-based metrics. Given the scale of our training runs, we had found that it was difficult to find accuracy-based variants of our evaluations that resulted in meaningful signal compared to noise.

\section*{Ethical Considerations}
\vspace*{-3pt}

This paper presents work whose goal is to advance the efficiency of training
and updating large language models.
These improvements could help democratize LM development by enabling resource-constrained organizations to participate in research and development, while reducing the environmental footprint of model training through decreased energy consumption.
However, increased accessibility of LM training also increases any risks that pertain to language models as a technology, such as the potential for malicious actors to spread disinformation.

Since our training data comes from Common Crawl and the general web, there may exist offensive and personally identifying content. This risk carries over from the original datasets we built ours from (Common Crawl, DCLM-Pool). Removing such data may affect the performance of models. Future work exploring the filtering of the original datasets may be directly applied to our training and evaluations as well.

\section*{Acknowledgements}
\vspace*{-2pt}
We are highly grateful to the many developers of the OpenLM library which was used extensively in this work. In particular, we'd like to thank George Smyrnis, Achal Dave, and Samir Yitzhak Gadre for helpful discussions regarding its usage. We'd also like to thank Chun-Liang Li, Pavan Kumar Anasosalu Vasu, Preetum Nakkiran, and Russ Webb for helping review earlier versions of the manuscript.

\bibliography{main}

\newpage
\appendix
\onecolumn
\input{sec/sup_dataset}

\input{sec/sup_evals}

\input{sec/sup_hparams}

\input{sec/sup_results}
\input{sec/sup_related}

\input{sec/future}
\end{document}

%% file: math_commands.tex
\usepackage{amsmath,amsfonts,bm}

\def\eqref#1{equation~\ref{#1}}

\def\1{\bm{1}}

\DeclareMathAlphabet{\mathsfit}{\encodingdefault}{\sfdefault}{m}{sl}
\SetMathAlphabet{\mathsfit}{bold}{\encodingdefault}{\sfdefault}{bx}{n}

%% file: packs.tex
\usepackage[utf8]{inputenc} %
\usepackage[T1]{fontenc}    %
\usepackage{xcolor}
\usepackage{algorithm}
\usepackage{amsfonts}       %
\usepackage{amsmath}
\usepackage{amssymb}
\usepackage{amsthm}
\usepackage{array}
\usepackage{bm}
\usepackage{booktabs}       %
\usepackage{colortbl}
\usepackage{empheq}
\usepackage{enumitem}
\usepackage{etoolbox}
\usepackage{float}
\usepackage{makecell}
\usepackage{mathrsfs}
\usepackage{mdframed}
\usepackage{microtype}      %
\usepackage{multirow}
\usepackage{multicol}
\usepackage{nccmath}
\usepackage{nicefrac}       %
\usepackage{pifont}
\usepackage{ragged2e}
\usepackage{rotating}
\usepackage{subcaption}
\usepackage{tabularx}
\usepackage{tikz}
\usetikzlibrary{tikzmark}
\usepackage{url}            %
\usepackage{wrapfig}
\usepackage{xcolor, color, colortbl}
\usepackage{xspace}
\usepackage{thm-restate}
\usepackage{xfrac}

\newcounter{rowcntr}[table]
\renewcommand{\therowcntr}{\arabic{chapter}.\the\numexpr\arabic{table}+1.\arabic{rowcntr}}

\AtBeginEnvironment{tabular}{\setcounter{rowcntr}{0}}
\newcolumntype{H}{>{\setbox0=\hbox\bgroup}c<{\egroup}@{}}

\newcommand{\PreserveBackslash}[1]{\let\temp=\\#1\let\\=\temp}
\newcolumntype{C}[1]{>{\centering\arraybackslash}m{#1}}
\newcolumntype{R}[1]{>{\raggedleft\arraybackslash}m{#1}}
\newcolumntype{L}[1]{>{\raggedright\arraybackslash}m{#1}}

\usepackage[capitalize]{cleveref}
\usepackage{pifont}
\crefname{section}{Sec.}{Secs.}
\Crefname{section}{Section}{Sections}
\Crefname{table}{Table}{Tables}
\crefname{table}{Tab.}{Tabs.}
\Crefname{appendix}{Appendix}{Appendices}
\crefname{appendix}{Appx.}{Apps.}
\usepackage{algorithm}
\usepackage{algorithmic}
\usepackage{amsmath}

\usepackage{listings}

\definecolor{jsondelim}{RGB}{0,128,0}   %
\definecolor{jsonkey}{RGB}{0,0,255}     %
\definecolor{jsonstring}{RGB}{160,32,240} %

\lstdefinelanguage{JSON}{
    basicstyle=\ttfamily\small,
    numbers=left,
    numberstyle=\scriptsize,
    stepnumber=1,
    numbersep=15pt,
    showstringspaces=false,
    breaklines=true,
    frame=single,
    morestring=[b]",
    stringstyle=\color{jsonstring},
    keywordstyle=\color{jsonkey},
    morekeywords={},
    literate=
     *{:}{{{\color{jsondelim}{:}}}}{1}
      {,}{{{\color{jsondelim}{,}}}}{1}
      {\{}{{{\color{jsondelim}{\{}}}}{1}
      {\}}{{{\color{jsondelim}{\}}}}}{1}
      {[}{{{\color{jsondelim}{[}}}}{1}
      {]}{{{\color{jsondelim}{]}}}}{1}
}

%% file: defs.tex
\definecolor{colorYes}{RGB}{51,160,44}
\definecolor{colorNo}{RGB}{228,26,28} %
\newcommand{\cmark}{\textcolor{colorYes}{\ding{51}}}%
\newcommand{\xmark}{\textcolor{colorNo}{\ding{55}}}%
\newcommand{\stdev}[1]{{\scalebox{0.7}{\color{gray} #1}}}

\newcommand{\ticcc}{\textsc{TiC-CC}\xspace}
\newcommand{\ticccwiki}{\textsc{TiC-CC-Wiki}\xspace}
\newcommand{\ticccnews}{\textsc{TiC-CC-News}\xspace}
\newcommand{\ticwiki}{\textsc{TiC-Wiki}\xspace}
\newcommand{\ticstack}{\textsc{TiC-StackE}\xspace}
\newcommand{\ticstackmath}{\textsc{TiC-StackE-Math}\xspace}
\newcommand{\ticstackoverflow}{\textsc{TiC-StackOverflow}\xspace}

\newcommand{\ticdocs}{\textsc{TiC-CodeDocs}\xspace}
\newcommand{\ticdocstorch}{\textsc{TiC-CodeDocs-PyTorch}\xspace}
\newcommand{\ticdocsnumpy}{\textsc{TiC-CodeDocs-NumPy}\xspace}
\newcommand{\staticevals}{\textsc{Core}\xspace}

\newcommand{\ppltoken}{\text{ppl}_{\text{token}}}
\newcommand{\pplanswer}{\text{ppl}_{\text{answer}}}
\newcommand{\pplnoun}{\text{ppl}_{\text{noun}}}

%% file: sec/abstract.tex
\begin{abstract}
   Large Language Models (LLMs) trained on historical web data inevitably become outdated.
   We investigate evaluation strategies and update methods for LLMs as new data becomes available.
   We introduce a web-scale dataset for \textit{time-continual pretraining} of LLMs derived from 114 dumps of Common Crawl (CC) -- orders of magnitude larger than previous continual language modeling benchmarks. 
    We also design time-stratified evaluations 
    across both general CC data and specific domains (Wikipedia, StackExchange, and code documentation) to assess how well various continual learning methods adapt to new data while retaining past knowledge.\footnote{Our code: \url{https://github.com/apple/ml-tic-lm}} Our findings demonstrate that, on general CC data, autoregressive meta-schedules combined with a fixed-ratio replay of older data can achieve comparable held-out loss to re-training from scratch, while requiring significantly less computation (2.6$\times$). 
    However, the optimal balance between incorporating new data and replaying old data differs as replay is crucial to avoid forgetting on generic web data but less so on specific domains.
\end{abstract}

%% file: sec/intro_v2.tex
\section{Introduction}
Large language models (LLMs) rely on massive amounts of data, most of which comes from large-scale web-crawls run over the past 
10--20 years. 
Common Crawl (CC), the most well-known source of such data, has been active since 2007 and continues to release monthly dumps of data.
While typically LLMs are trained from scratch on many (or all) previous dumps jointly \citep{refinedweb, li2024datacomp}, they also suffer from the knowledge cutoffs of their training sets causing their performance to deteriorate on newer data \citep{cheng2024dated}. Combined with the vast costs of re-training LLMs from scratch, a natural question is how LLMs can 
best be \textit{continually} reused and updated over many months and years.

\begin{figure}[t]
    \centering
    \includegraphics[width=0.95
    \linewidth]{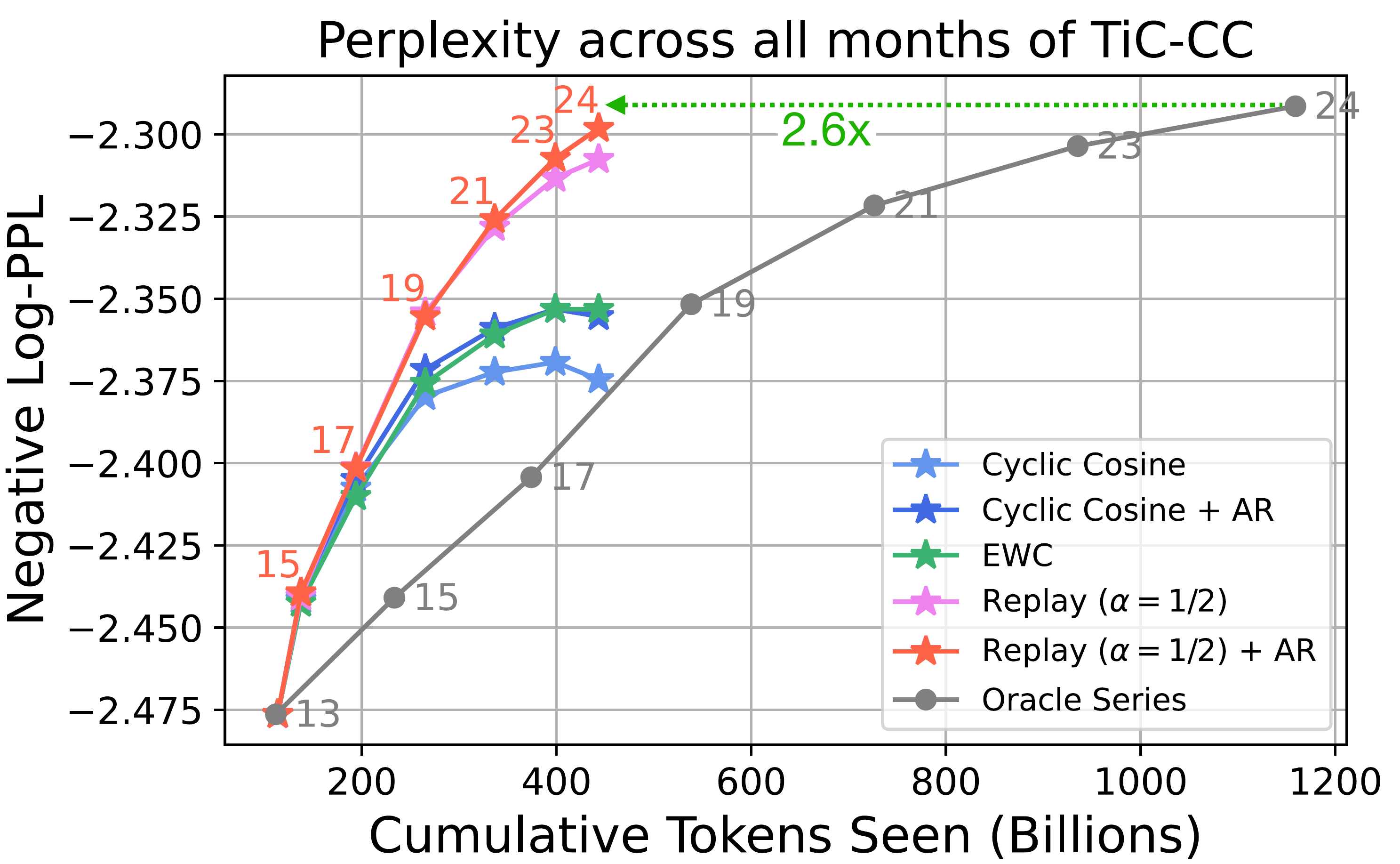}
    \caption{\textbf{Replay allows for matching repeated from-scratch training.} Each line traces a continually trained 3B model where the annotations indicate the data cutoff year of the checkpoint (e.g., ``24'' = 2024). We find that combining autoregressive (AR) learning rate schedules and data replay (red) can nearly match the perplexity on all months achieved by the Oracle series which re-trains from scratch every two years (gray), despite requiring 2.6$\times$ less compute. Meanwhile, methods that only modify the optimizer (blue) or loss (green) insufficiently prevent forgetting and plateau. }
    \vspace{-6.5mm}
    \label{fig:ticcc_efficiency}
\end{figure}

\begin{figure*}[t!]
    \centering
    \includegraphics[width=0.95\linewidth]{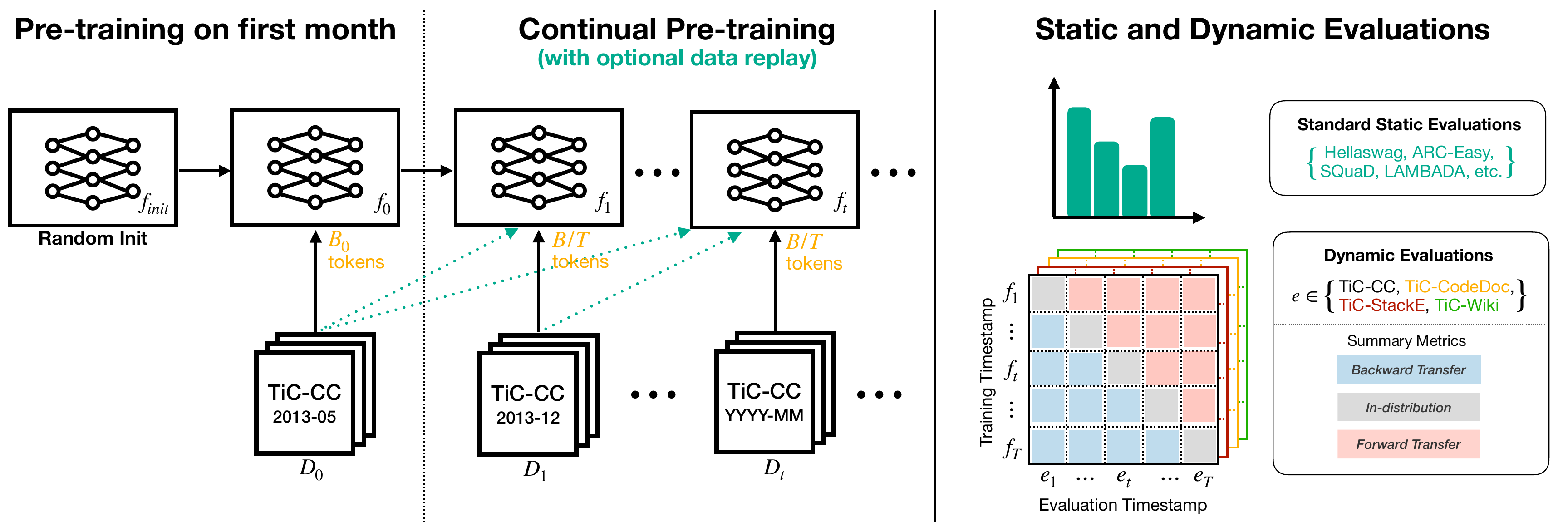}
\vspace*{-6pt}
    \caption{\textbf{The TiC-LM benchmark.} We simulate a setup where each Common Crawl dump $D_0, \cdots, D_T$ is revealed one-at-a-time. An LLM $f_0$ is first pre-trained for $B_0$ tokens on the initial month $D_0$ and then continually updated for a fixed budget of $B/T$ tokens in each following month (optionally replaying older data). The goal is for each monthly model $f_1, \cdots, f_T$ to perform well on both standard static downstream tasks as well as dynamic evaluations that evolve over time, requiring the balance of learning (gray/red) with preventing forgetting (blue). }
    \label{fig:summary-figure}
\vspace*{-4pt}
\end{figure*}

To study this question, our take is that a benchmark of appropriate scale and scope is a key prerequisite missing from the current literature. In particular, many prior works on continual language modeling train and evaluate on single domains such as Wikipedia, News, or Twitter/X \citep{jang2022temporalwiki, liska2022streamingqa, luu2021time}. However in practice, LLMs are trained on \textit{general} web-scale data (in which many implicit domains and their relative presence both evolve over time) with the explicit goal of performing well across many types of tasks. Further, while more recent efforts \citep{gupta2023continual, parmar2024reusedontretrain, ibrahim2024simple} do study continual LLM training at web-scale, their setups do not focus on \textit{temporal} distribution shifts and contain generally less than three training rounds. This limits their potential generalizabiltiy  for studying time-continual learning across longer horizons in a truly lifelong sense.

\vspace{-0.2mm}
Taking inspiration from TiC-CLIP~\citep{garg2024tic}, our work aims to address these gaps by introducing the \textbf{TiC-LM} (\textbf{Ti}me-\textbf{C}ontinual Learning of \textbf{L}anguage \textbf{M}odels) \textbf{benchmark} (see \cref{fig:summary-figure}). Our setup centers on TiC-CommonCrawl (\ticcc{}), a massive time-stratified set 
of training and held-out evaluation data created using 114 months (May-2013 to July-2024) of CC data, where during training, each month is revealed only one at a time. In total, \ticcc{} contains 2.9T possible training tokens spread among the 114 timesteps, providing 100$\times$ more potential tokens and 10$\times$ more timesteps compared to prior time-continual LLM benchmarks. For evaluation, TiC-LM also provides several domain-specific \textit{dynamic} evaluations sourced from diverse domains including TiC-Wikipedia (\ticwiki{}), TiC-StackExchange (\ticstack{}), and \ticdocs{}.

Using our benchmark, we run over 150 experiments to evaluate the effectiveness of different optimization, replay, and regularization-based continual learning strategies. These results provide insights into the following key questions:

\vspace*{-2mm}

\begin{itemize}[leftmargin=*]
\itemsep0em
    \item \textit{Can continual pretraining match periodic re-training at lower cost?}
    We find that a mix of learning rate and data replay strategies allows us to be competitive with a series of models that retrains from scratch every two years requiring 2.6$\times$ less total compute (see \cref{fig:ticcc_efficiency}). However, trade-offs remain between continual methods and re-training across  domains and evaluations.

    \item \textit{Is forgetting a challenge when continually pretraining on web-data?} 
    We observe that on general web-data in \ticcc{}, older CC dumps are significantly forgotten when only training on new data and replay is essential to retain performance on these earlier dumps (see \cref{fig:ticcc_main}). 

    \item \textit{Is the impact of forgetting domain-dependent?} Forgetting older CC dumps need not always be detrimental. Replaying old data can actually hurt when evaluating on rapidly evolving domains like StackOverflow and PyTorch, while still benefiting more stable ones where older dumps are more useful such as Math and NumPy (see \cref{fig:downstream_heatmaps}).    

\end{itemize}

We release all code for our training and evaluation setups. Our hope is that the broader research community will use these assets to further realize the potential of continual LLM pretraining.

%% file: sec/related.tex
\section{Related Work}\label{sec:related}

Learning new capabilities from multiple, sequentially observed, distributions has long been 
an active area of ML research~\citep{wang2024survey}. More recently, several 
works have studied this setting for  LLMs~\citep{wu2024surveyllm},  
targeting improvements on: (1) general capabilities (via higher-quality datasets) \citep{parmar2024reusedontretrain, ibrahim2024simple, 
gupta2023continual}; (2) specific domains~\citep{jin2022lifelong, 
gururangan2020dont, chen2024effectiveefficientcontinualpretraining}; (3) newer 
data as the world evolves~\citep{jin2022lifelong, jang2021towards, 
jang2022temporalwiki, lazaridou2021mind, nylund2023timeencoded, 
loureiro2022timelms, qin2022elle, liska2022streamingqa}. Works in this third 
category have shown that in many domains, the performance of LLMs decays 
as training and evaluation sets grow farther apart in time, motivating the need 
for methods to \textit{efficiently} adapt to 
new data while retaining existing knowledge. Our work scales up these previous efforts to more closely 
match current LLM training practices. While older works typically focus on 
continual training and evaluation over individual sources (e.g., news, Wikipedia, 
and social media) and $\leq$10 timesteps, we consider training on a \textit{generic 
web-crawl} (i.e., Common Crawl) spanning 114 different months.
In turn, the generality of our training data allows us to go beyond 
single-domain evaluations. \Cref{tab:cl_dataset_comparison} summarizes our proposed datasets compared with
the most related continual benchmarks.
With 2.9T tokens, \ticcc{} is the \textit{largest and most diverse} continual 
learning benchmark for language modeling. We provide an extended discussion of related works 
in \cref{sec:related_sup}.

\begin{table}[t]
    \centering
    \caption{Comparison of TiC-LM (bottom) with previous continual learning studies for LLMs. The ``Temporal'' column refers to whether the continual training rounds are defined across temporal distribution shifts. Training set size is in terms of tokens unless otherwise specified (i.e., Art. = Articles, QA = Question-Answer pairs). }
    \vspace*{-3pt}
    \label{tab:cl_dataset_comparison}
    \resizebox{0.48\textwidth}{!}{
    \begin{tabular}{lcccc}
        \toprule[1.2pt]
        \textbf{Benchmark}
        & \textbf{Domain}
        & \textbf{Temporal}
        & \textbf{Steps}
        & \textbf{\# Train} \\
        \midrule[1.2pt]
        \citet{gupta2023continual} & Web & \xmark & 2-3 & 297B \\
        \citet{parmar2024reusedontretrain} & Web & \xmark & 2 & 3.8B-500B \\
        \citet{ibrahim2024simple} & Web & \xmark & 2--7 & 100B\\ \hline
        Chrono. Tweet \citeyearpar{jin2022lifelong} & \makecell[l]{Science, Tweets} & \cmark & 4 & 25M \\
        TempEL~\citeyearpar{zaporojets2022tempel} & Wiki & \cmark & 10 & <138k Art. \\
        TemporalWiki~\citeyearpar{jang2022temporalwiki} & Wiki & \cmark & 4 & 23B \\
        StreamingQA~\citeyearpar{liska2022streamingqa} & News & \cmark & 12 & 99k Art. \\
        EvolvingQA~\citeyearpar{kim2023carpe} & Wiki & \cmark & 6 & 390k Art. \\ 
        TIQ~\citeyearpar{jia2024tiq} & Wiki & \cmark & --- & 6k QA \\
        TAQA~\citeyearpar{zhao2024set} & Wiki & \cmark & --- & 9k QA \\
        \citet{luu2021time} & \makecell[c]{Science, News\\Reviews, Tweets} & \cmark & 4--7 & 695k Art. \\
        \midrule
        \ticcc{} & Web & \cmark & 114 & \makecell[c]{220B-440B \\ (up to 2.9T)} \\
        \ticwiki{} & Wiki & \cmark & 62 & --- \\
        \ticstack{} & \makecell[c]{Diverse QA} & \cmark & 8--170 & --- \\
        \ticdocs{} & Code & \cmark & 11--16 & --- \\
        \bottomrule[1.2pt]
    \end{tabular}
    }
    \vspace{-5pt}
\end{table}

%% file: sec/dataset.tex
\section{\ticcc{}: over 10 years of web data}\label{sec:tic_cc_data}

We create a large English \textit{time-stratified} dataset of 2.9T tokens based upon 
Common Crawl (CC), a free and open corpus of web-crawled data that has been online 
since 2007.
We collect all dumps between May-2013 and July-2024, resulting in 114 
corresponding splits that we refer to by the month of their release date.
For each split, we then apply a pre-processing pipeline based on that of 
DataComp-LM~\citep{li2024datacomp} to create a corpus representative of existing pre-training datasets.
Notably, to retain causality, we do not perform 
any operations on older months that depend on future months. 

\textbf{Data processing.} We build upon the assets from DataComp-LM
\citep{li2024datacomp}, 
starting with DCLM-Pool~\citep{li2024datacomp}, which contains all CC dumps up to Dec-2022 and parsed to extract 
plaintext from webpages via the \texttt{resiliparse} 
library~\citep{bevendorff2018,bevendorff2021c}.
We split this data by month and reuse the 
same download and processing scripts to extend DCLM-Pool until 
July-2024. Next, we follow DCLM-Baseline's pipeline by applying 
heuristic filters from RefinedWeb~\citep{refinedweb} and a fuzzy-deduplication 
step which we modify to run only \textit{within} each month rather than across all months. Alternatively, as in TiC-CLIP, one could deduplicate newer months against older ones. Here, we do not do this by default given that it may not always be helpful (e.g., it may remove updated or edited pages where a few key facts have changed but most text remains the same). We perform an initial exploration of cross-month deduplication in \Cref{app:deduplication} and overall allow for exploring the benefits/pitfalls of such data-centric interventions as part of method design.
Finally, we do not use the final classifier-based filter in DCLM-Baseline, as this classifier was trained on data from all months, violating causality. For more details about the data pipeline, see \cref{app:data_pipeline}.

In \cref{fig:token_counts} (left), we show the number of tokens we have for 
each month of \ticcc{}. In total, the dataset spans 29T tokens, with 
individual months ranging between 100B to 500B tokens.
Our experiments train on subsets of 220-440B tokens from a single global shard that contains 2.9T
while future work can expand to the full 29T.

%% file: sec/evals.tex
\section{Evaluations}\label{sec:evaluations}

In this section, we discuss various time-continual evaluations that are 
designed both with and independent of \ticcc{}.
As our focus is on pretraining, we focus on evaluations without 
instruction-tuning.

\textbf{Perplexity metrics.}
We employ three distinct perplexity metrics for different evaluations:

\begin{equation}
\vspace*{-2pt}
\ppltoken = \exp\left(\frac{\sum_{d \in \mathcal{D}} \sum_{t \in T_d} -\log P(t|c_{<t})}{\sum_{d \in \mathcal{D}} |T_d|}\right)\,
\end{equation}
where $\mathcal{D}$ is a set of pages, $T_d$ is the set of tokens in page $d$, and $c_{<t}$ is the context prior to token $t$.
\vspace*{-5pt}
\begin{equation}
\vspace*{-5pt}
\pplanswer = \frac{1}{|\mathcal{Q}|} \sum_{q \in \mathcal{Q}} \exp\left(-\log P(a_q|c_q)\right)\,
\end{equation}
where $\mathcal{Q}$ is a set of question-answer pairs, $a_q$ is the gold answer for question $q$, and $c_q$ is the context.
\begin{equation}
\vspace*{-2pt}
\pplnoun = \exp\left(\frac{\sum_{d \in \mathcal{D}} \sum_{n \in N_d} -\log P(n|c_{<n})}{\sum_{d \in \mathcal{D}} |N_d|}\right)\,
\end{equation}
where $N_d$ is the set of proper noun tokens found
by a PoS 
tagger \citep{honnibal2017spacy} in page $d$, and $c_{<n}$ is the context prior to noun $n$.

\subsection{TiC-CommonCrawl (\ticcc{})}\label{sec:loss_evals}

We compute  $\ppltoken$ on three monthly subsets of our CC data which 
were held out from training:
\vspace*{-2mm}
\begin{itemize}[leftmargin=*]
\itemsep -0.4mm
\item \ticcc{}: Held-out pages coming from the full distribution for each month of \ticcc{}.
\item \ticccwiki{}: Pages in \ticcc{} from English Wikipedia and Simple English Wikipedia 
\item \ticccnews{}: Pages in \ticcc{} from a set of news sites based on   
    WMT competitions ~\citep{wmt}. 
\end{itemize}
\vspace*{-2mm}

These evaluations align exactly with the training objective and data, providing direct signal on how well different months and subsets were learned.  

\subsection{TiC-Wikipedia (\ticwiki{})}

\ticccwiki{} in \cref{sec:loss_evals} uses a representative subset of Wikipedia pages found in CC dumps. In contrast, \ticwiki{} aims to (1) comprehensively build examples using complete Wikipedia dumps; (2) isolate changed/unchanged factual knowledge through focusing on proper nouns ($\pplnoun${}) as opposed to all tokens ($\ppltoken${}), following \citep{jang2022temporalwiki, lazaridou2021mind}. To construct \ticwiki{}, we build upon TemporalWiki~\citep{jang2022temporalwiki}, expanding its coverage from four months to a decade (2014--2024) while improving the parsing of Wikipedia (see \cref{sec:tic_wiki_sup}). This results in \ticwiki{}-Diff and \ticwiki{}-Unchanged which capture the changed and unchanged facts across dumps, respectively.

\subsection{TiC-StackExchange (\ticstack{})}
We design another
evaluation based on the historical data 
from StackExchange.
StackExchange has 182 communities 
that share knowledge by posting questions and answers. We measure answer-perplexity ($\pplanswer$) on high-quality answers from selected sites by collecting answers that have been accepted by the question author (using the accepted answer timestamp to bin examples by month). The resulting 
evaluation contains examples from 2008--2024
(see \cref{sec:tic_stack_sup} for more details).

\subsection{\ticdocs{}}
Our \ticdocs{} evaluation is based on the official documentation from major releases of NumPy~\citep{numpy} (16 releases from 2017-2024, \texttt{v1.13.0} to \texttt{v2.1.0}) and PyTorch~\citep{Pytorch} (11 releases from 2021-2024, \texttt{v1.8.0} to \texttt{v2.4.0}). 
We build documentation from each library's Git repository by reverting to release commits and generating HTML documentation from source, which we convert to plain text using \texttt{readability.js} to retain only the main content. We evaluate model performance using perplexity ($\ppltoken{}$) across version snapshots.

\subsection{\textbf{Static downstream evaluations.}}
We evaluate models on a variety of downstream zero-shot and few-shot 
tasks suitable for base models. Specifically, we use the \staticevals{} average from the DCLM 
benchmark~\citep{li2024datacomp} which includes 22 zero-shot and few-shot 
in-context learning tasks. These evaluations, which include benchmarks such as 
ARC-Easy~\citep{arc} and Hellaswag~\citep{hellaswag},
assess general capabilities of base 
models via a variety of world knowledge and natural language understanding 
tasks. While they are not designed to be time-dependent, we use 
them to assess whether continually trained models match the general 
capabilities of models trained on all dumps. 

%% file: sec/method.tex
\section{Continual Learning Baseline Methods}\label{sec:methods}

We now go over the continual methods that we test on our benchmark as well as relevant non-continual baselines. For continual methods, we consider the following three categories: optimization-based, data replay, and 
regularization.

\textbf{Optimization-based methods.} In non-continual settings, LLMs are 
often trained with a cosine-decayed learning rate schedule which requires 
knowledge of total training steps ahead of time. In a continual setup, however, 
the number of total tokens grows over time and we care about the performance after each month. We benchmark the following optimization approaches in our work:

\begin{itemize}[leftmargin=*]
\itemsep-1pt
    \item \textit{Cyclic Cosine} is the simplest alternative which applies cosine decay \textit{within} 
        each training month using the same maximum learning rate and warmup (\cref{fig:cyclic_lr}, green). This was found to be most effective in TiC-CLIP \citep{garg2024tic}.
    \item \textit{Cyclic Cosine + AR (autoregressive)} uses Cyclic Cosine but also decays 
        the maximum learning rate across rounds, regressed 
        from a global cosine schedule (\cref{fig:cyclic_lr}, blue). It was shown to offer improvements by 
        \citet{roth2024practitioner}.
    \item \textit{Rsqrt} (\textit{reciprocal-$\sqrt{\ }$}) are \textit{infinite} schedules that decay the learning rate slowly in a global training run and branch off of this trajectory with linear cooldowns \citep{Zhai2022scaling}. To keep training steps fixed compared to other methods, we follow \citet{roth2024practitioner} and implement a version that maintains only a single trajectory by re-warming up from the previous cooldown.
    \item \textit{Schedule-Free} is an optimizer proposed by 
        \citet{defazio2024roadscheduled} which aims to circumvent the need for 
        defining a learning rate schedule by using iterate averaging and has 
achieved promising results in {i.i.d.} non-continual settings.
\end{itemize}
\vspace*{-5pt}

\textbf{Data replay methods.}
A classical strategy to prevent forgetting is to mix the current timestep's data with data from earlier timesteps~\citep{lopez2017gradient,rebuffi2017icarl,chaudhry2018efficient}.  
Defining a replay method therefore boils down to defining the mixture 
ratios across rounds. Based on TiC-CLIP ~\citep{garg2024tic}, we mostly consider variants of the form: 

\vspace*{-5pt}
\begin{itemize}[leftmargin=*]
\itemsep-1pt
    \item For the current timestep $t$, we allocate a ratio $0 \le \alpha_t \le 
        1$ of the monthly token budget $B_t$ to data from the current month, 
        seeing $\alpha_t B_t$ tokens from that month.
    \item For previous months, we redistribute the remaining $(1-\alpha_t)B_t$ tokens equally, i.e., each month contributing $\frac{1-\alpha_t}{t-1}B_t$ tokens to this round's data.
\end{itemize}
\vspace*{-3pt}

In our setup, $B_t$ is fixed across months. We first try $\alpha_t = 1/t$, which sees an equal number of tokens from all observed months.
We also consider the constant value $\alpha_{t} = 1/2$ which always allocates half the 
token budget to the current month. In \Cref{app:extended_results}, we try other fixed settings of $\alpha_{t}$ as well as exponentially down-weighting older months based on distance from the current timestep. One potential downside of replay-based 
methods is the cost of retaining old data, especially if 
old data expires and needs to be removed. Methods with larger values of $\alpha_{t}$ are less affected by such 
limitations. We do not consider such costs in our work (as we assume they are likely to be dominated by training costs). 

\textbf{Regularization-based methods.}
These methods alter the training objective by 
adding a regularization term which encourages newer model updates to stay close 
to the model weights learned after the previous month. Following TiC-CLIP, we 
try two notable methods: LwF~\citep{li2018lwf} and 
EWC~\citep{kirkpatrick2017overcoming}. 
\vspace*{-5pt}
\begin{itemize}[leftmargin=*]
\itemsep-1pt
   \item \textit{LwF} adds a KL divergence loss term to 
       penalize differences in model outputs between the previous checkpoint 
       and the current model.
   \item \textit{EWC} attempts to slow down updates to particular model 
       parameters which are highly influential for performing well on older 
       months as measured by the (approximate) Fisher information 
       matrix.
\end{itemize}
\vspace*{-3pt}

Because both LwF and EWC involve extra loss terms and model copies, it is 
important to note that they induce larger GPU memory footprints and run-times 
compared to optimizer and replay-based methods. That being said, we do not try 
to adjust the token counts to account for this given that our 
re-implementations may not be optimally efficient.

\textbf{Non-continual Oracles.} The alternative to continual learning is to simply retrain a model from scratch on an equal number of tokens from all currently available months. We refer to such a model as \textit{Oracle-$t$} where $t$ is particular cutoff date. We then consider as a baseline a \textit{series of Oracle models} re-trained roughly every two years (i.e., \{2013-05, 2015-01, 2017-01, 2019-01, 2021-01, 2023-01, 2024-07\}), where for any timestamped evaluation, the most recently trained \textit{Oracle} is always used. 

%% file: sec/experiments.tex
\section{Experiments}\label{sec:experiments}

\begin{figure*}[t]
    \centering
\includegraphics[height=0.242\linewidth]{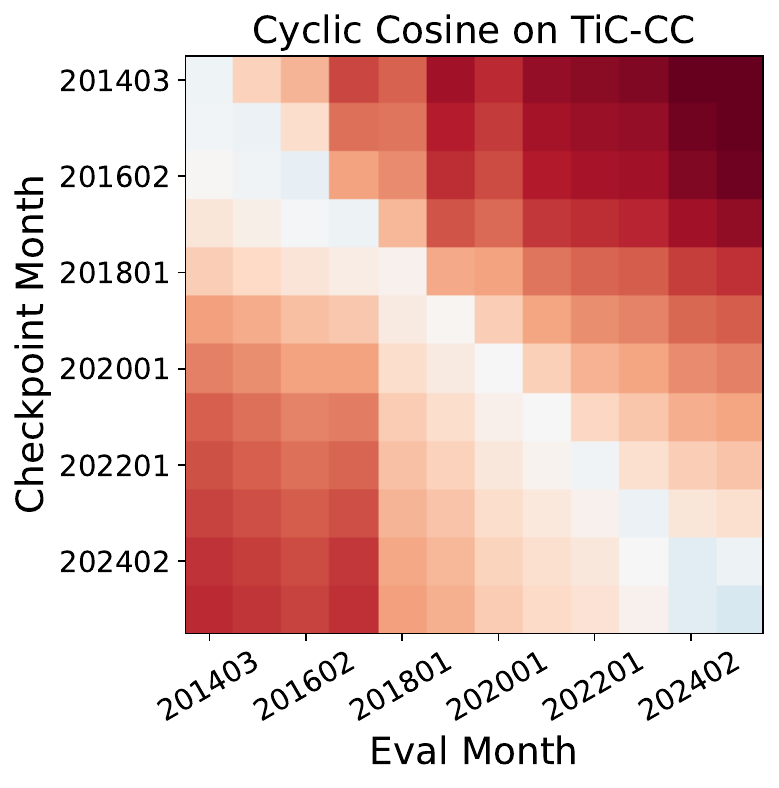} 
\includegraphics[height=0.242\linewidth]{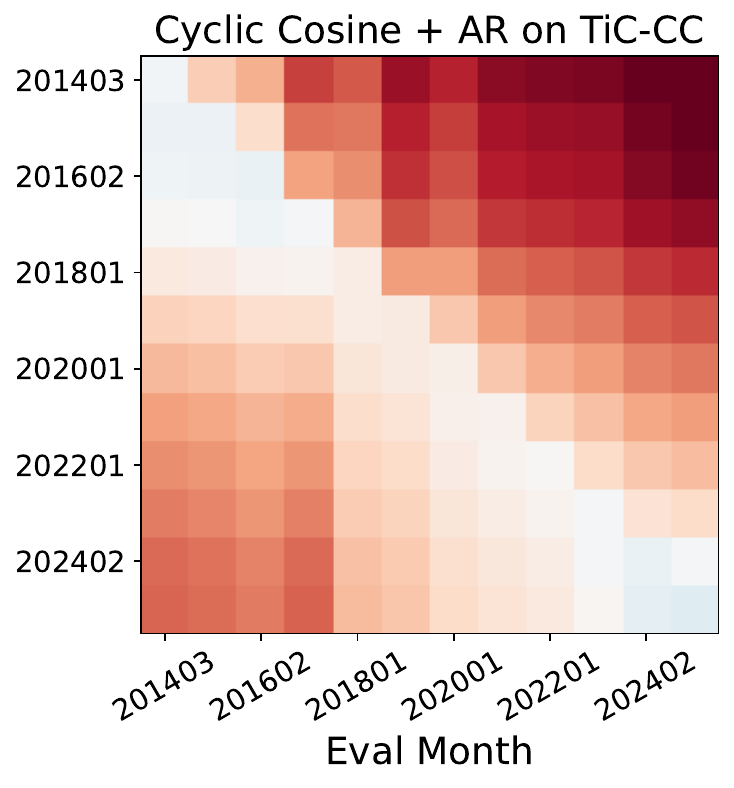} 
\includegraphics[height=0.242\linewidth]{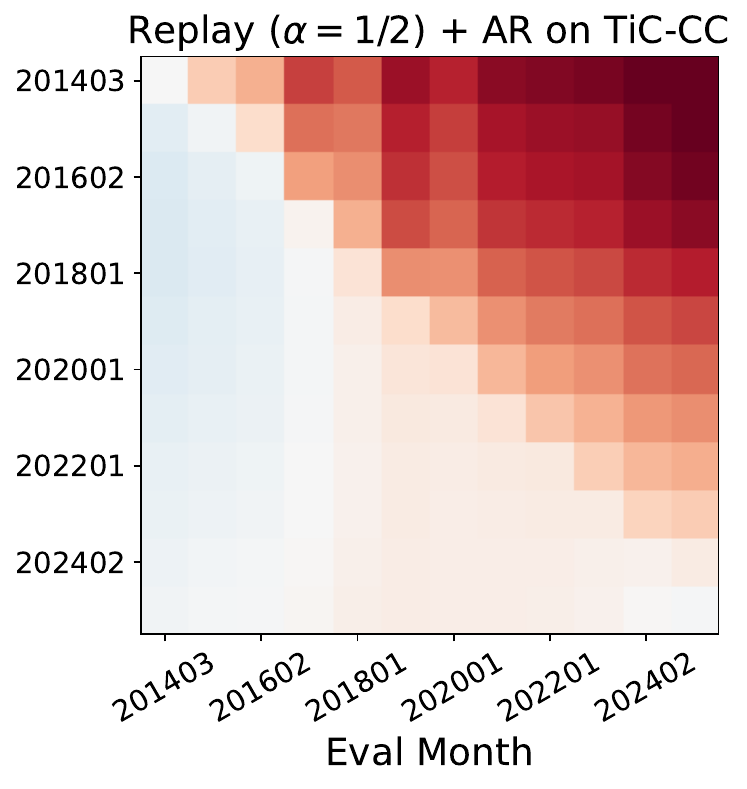} \includegraphics[height=0.242\linewidth]{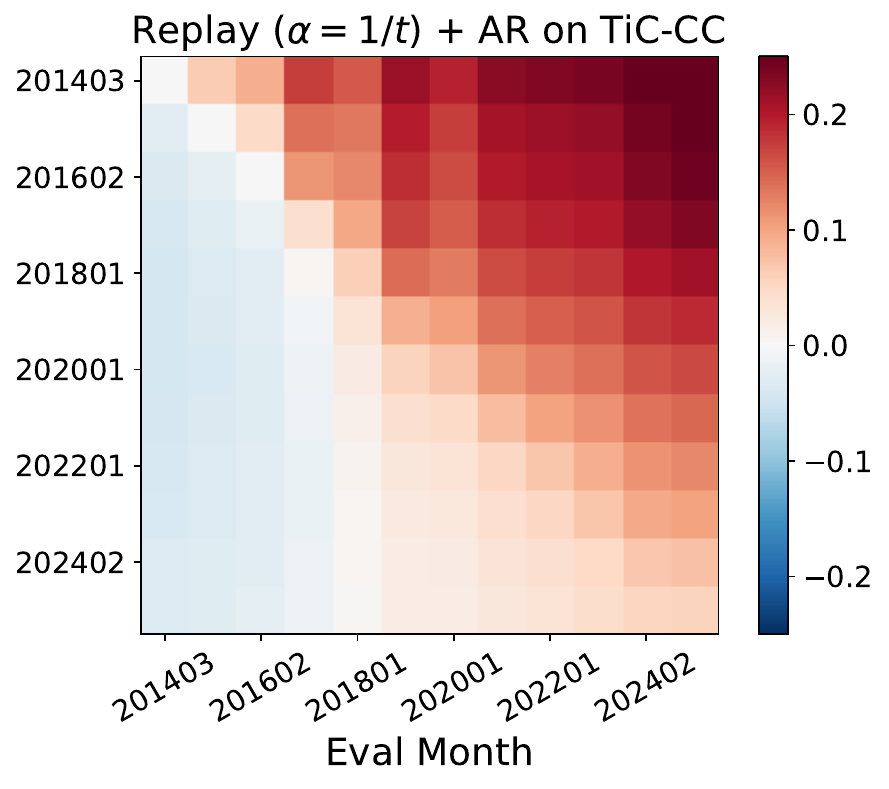}
\vspace*{-3mm}
    \caption{
    \textbf{Continual methods incur different trade-offs between ID and Backward performance on \ticcc{}.} 
    We plot selected regret matrices for 3B models and 440B training tokens (lower is better). Overall, Cyclic Cosine leads to strong ID performance (along the diagonal) but also significant forgetting after a few years. This can be partially addressed by using an AR meta-schedule and more significantly by replay. However, too much replay such as with $\alpha_t = 1/t$ scales poorly with a large number of timestamps, significantly sacrificing ID performance.
    }
    \label{fig:ticcc_main}
    \vspace{-2mm}
\end{figure*}

\begin{table*}[t!]
    \scriptsize
    \centering
    \caption{\textbf{Loss-based evaluations for various methods.} We report log-perplexity relative to \textit{Oracle-2024-07} models of the same size. While optimizer (top) and regularization-based (bottom) methods trade-off backward transfer with ID performance, replay (middle) is required to obtain the least amount of forgetting.
    \textbf{Bold} values are within one standard deviation of the best in each column for a given token budget, with standard deviations estimated from three runs of Cyclic Cosine. \colorbox{green!15}{Highlighted} values indicate when a continual run outperforms the  \colorbox{gray!25}{Oracle series}.}
    \vspace*{-2mm}
    
    \begin{tabular}{c|cc|ccc|ccc|ccc}
        \toprule[1.2pt]
        \multirow{2}{*}{\textbf{Method}} & \multirow{2}{*}{\textbf{Tokens}} & \multirow{2}{*}{\textbf{Params}}  & \multicolumn{3}{c|}{\textbf{\ticcc{} 
        $\downarrow$}} & \multicolumn{3}{c|}{\textbf{ \ticccwiki{} 
        $\downarrow$}} & \multicolumn{3}{c}{\textbf{\ticccnews{} $\downarrow$}} 
        \\
        & & & Bwd. & ID & Fwd. & Bwd. & ID & Fwd. & Bwd. & ID & Fwd.  \\
        \midrule[1pt]
        \multirow{2}{*}{Cyclic Cosine \stdev{(std)}} & \multirow{2}{*}{220B} & \multirow{2}{*}{1B} & 0.079	& \highlight{\textbf{0.026}} & \highlight{\textbf{0.154}} &  0.041 & 0.033 & 0.072 &	0.065 &	\highlight{0.016} &	0.105 \\[-3pt] 
        & & &  \stdev{(0.000)} & \stdev{(0.000)} & \stdev{(0.000)} & \stdev{(0.000)} & \stdev{(0.000)} & \stdev{(0.000)} & \stdev{(0.000)} & \stdev{(0.000)} & \stdev{(0.000)} \\[+1pt]
        \multirow{1}{*}{Cyclic Cosine + AR} & 220B & 1B & 0.065 & 0.040 & 0.158 &	0.036 & 0.033 & 0.073 &	0.047 & \highlight{0.017} & 0.106 \\
        \multirow{1}{*}{Cyclic Rsqrt} & 220B & 1B & 0.073 & \highlight{0.031} & \highlight{\textbf{0.154}} & 0.039 & 0.032 & 0.071 & 0.057 & \highlight{\textbf{0.015}} & 0.104  \\
        \multirow{1}{*}{Schedule-Free} & 220B & 1B & 0.075 & 0.036 & 0.158 & 0.043 & 0.037 & 0.075 & 0.058 & \highlight{0.019} & 0.107 \\ \hline
        \multirow{1}{*}{Replay ($\alpha=1/t$)} & 220B & 1B & \textbf{0.026} & 0.072 & 0.170 & 0.024 & 0.036 & 0.076  & 0.008 & 0.034 & 0.112 \\
        \multirow{1}{*}{Replay ($\alpha=1/2$)} & 220B & 1B & 0.028 & 0.042 & 0.159 & 0.026 & 0.032 & 0.073  & 0.017 & \highlight{0.019} & 0.107 \\
        \multirow{1}{*}{Replay ($\alpha=1/t$) + AR} & 220B & 1B & 0.029 & 0.080 & 0.173 & \textbf{0.022} & 0.038 & 0.076 & \textbf{0.007} & 0.038 & 0.113 \\
        \multirow{1}{*}{Replay ($\alpha=1/2$) + AR} & 220B & 1B & 0.029 & 0.054 & 0.163 & 0.025 & 0.033 & 0.073 & 0.012 & 0.022 & 0.107 \\
        \hline
        \multirow{1}{*}{LwF} & 220B & 1B & 0.079 & \highlight{0.027} & \highlight{\textbf{0.154}} & 0.042 & 0.034 & 0.073 & 0.066 & \highlight{0.016} & 0.105  \\
        \multirow{1}{*}{EWC} & 220B & 1B & 0.067 & \highlight{0.035} & 0.155 & 0.035 & \textbf{0.031} & \textbf{0.070} & 0.050 & \highlight{\textbf{0.015}} & \textbf{0.103 } \\ \midrule[1pt]
        \rowcolor{lightgray!25} \multirow{1}{*}{Oracle Series} & 1.16T & 1B & -0.003 & 0.035 & 0.154 & 0.004 & 0.016 & 0.062 & -0.008 & 0.019 & 0.100  \\ 
        \bottomrule[1.2pt] \bottomrule[1.2pt]

        \multirow{2}{*}{Cyclic Cosine \stdev{(std)}} & \multirow{2}{*}{220B} & \multirow{2}{*}{3B} & 0.072 & \highlight{\textbf{0.027}} & \highlight{\textbf{0.161}} & 0.038 & 0.032 & 0.074 & 0.058 & \highlight{0.015} & 0.109 \\[-3pt] 
        & & &  \stdev{(0.000)} & \stdev{(0.000)} & \stdev{(0.000)} & \stdev{(0.000)} & \stdev{(0.000)} & \stdev{(0.000)} & \stdev{(0.000)} & \stdev{(0.000)} & \stdev{(0.000)} \\[+1pt]
        \multirow{1}{*}{Cyclic Cosine + AR} & 220B & 3B & 0.058 & 0.040 & 0.166 & 0.032 & 0.031 & 0.074 & 0.041 & \highlight{0.017} & 0.110 \\
        \multirow{1}{*}{Cyclic Rsqrt} & 220B & 3B  & 0.065 & \highlight{0.030} & \highlight{0.162} & 0.033 & 0.030 & 0.073 & 0.049 & \highlight{0.015} & \textbf{0.108} \\
        \multirow{1}{*}{ScheduleFree} & 220B & 3B & 0.065 & \highlight{0.036} & 0.164 & 0.036 & 0.033 & 0.076 & 0.049 & \highlight{0.017} & 0.110 \\ \hline
        \multirow{1}{*}{Replay ($\alpha=1/t$)} & 220B & 3B  & \textbf{0.023} & 0.074 & 0.178 & 0.020 & 0.036 & 0.078 & 0.005 & 0.035 & 0.117 \\
        \multirow{1}{*}{Replay ($\alpha_t=1/2$)} & 220B & 3B & 0.024 & 0.042 & 0.167 & 0.024 & 0.031 & 0.074 & 0.013 & \highlight{0.019} & 0.111
        \\ 
        \multirow{1}{*}{Replay ($\alpha=1/t$) + AR} & 220B & 3B &  0.026 & 0.083 & 0.181 & \textbf{0.019} & 0.037 & 0.079 & \textbf{0.004} & 0.039 & 0.119 \\
        \multirow{1}{*}{Replay ($\alpha=1/2$) + AR} & 220B & 3B & 0.025 & 0.055 & 0.171 & 0.022 & 0.032 & 0.076 & 0.009 & 0.022 & 0.112 \\ \hline
        \multirow{1}{*}{LwF} & 220B & 3B & 0.072 & \highlight{\textbf{0.027}} & \highlight{\textbf{0.161}} & 0.038 & 0.032 & 0.074 & 0.058 & \highlight{0.015} & 0.109 \\
\multirow{1}{*}{EWC} & 220B & 3B & 0.061 & \highlight{0.032} & \highlight{0.162} & 0.031 & \textbf{0.029} & \textbf{0.071} & 0.046 & \highlight{\textbf{0.014}} & \textbf{0.108} \\ \bottomrule[1.2pt] 
        \multirow{2}{*}{Cyclic Cosine \stdev{(std)}} & \multirow{2}{*}{440B} & \multirow{2}{*}{3B} & 0.082 & \highlight{\textbf{-0.011}} & \highlight{\textbf{0.145}} & 0.029 & \highlight{0.015} & \highlight{0.059} & 0.071 & \highlight{-0.001} & \highlight{0.099} \\[-3pt] 
        & & & \stdev{(0.000)} & \stdev{(0.000)} & \stdev{(0.000)} & \stdev{(0.000)} & \stdev{(0.000)} & \stdev{(0.000)} & \stdev{(0.000)} & \stdev{(0.000)} & \stdev{(0.000)} \\[+1pt]
        \multirow{1}{*}{Cyclic Cosine + AR} & 440B & 3B & 0.058 & \highlight{-0.002} & \highlight{0.148} & 0.014 & \highlight{\textbf{0.009}} & \highlight{\textbf{0.057}} & 0.044 & \highlight{\textbf{-0.005}} & \highlight{\textbf{0.097}} \\
        \multirow{1}{*}{Cyclic Rsqrt} & 440B & 3B & 0.067 & \highlight{-0.007} & \highlight{0.146} & 0.018 & \highlight{0.010} & \highlight{\textbf{0.057}} & 0.055 & \highlight{-0.004} & \highlight{\textbf{0.097}} \\
        \multirow{1}{*}{Schedule-Free} & 440B & 3B& 0.063 & \highlight{-0.004} & \highlight{0.147} & 0.017 & 0.011 & \highlight{0.059} & \highlight{0.049} & \highlight{-0.004} & \highlight{0.098} \\ \hline
        \multirow{1}{*}{Replay ($\alpha=1/t$)} & 440B & 3B & 0.001 & 0.044 & 0.164 & \highlight{0.003} & \highlight{0.016} & \highlight{0.062} & \highlight{-0.009} & \highlight{0.013} & \highlight{0.105} \\
        \multirow{1}{*}{Replay ($\alpha=1/2$)} & 440B & 3B & 0.007 & \highlight{0.007} & \highlight{0.151} & 0.010 & \highlight{0.013} & \highlight{0.059} & 0.005 & \highlight{-0.000} & \highlight{0.099} \\
        \multirow{1}{*}{Replay ($\alpha=1/t$) + AR} & 440B & 3B & \highlight{\textbf{-0.003}} & 0.050 & 0.166 & \highlight{\textbf{-0.006}} & \highlight{0.013} & \highlight{0.061} & \highlight{\textbf{-0.017}} & \highlight{0.014} & \highlight{0.105} \\
        \multirow{1}{*}{Replay ($\alpha=1/2$) + AR} & 440B & 3B & -0.002 & \highlight{0.016} & \highlight{0.154} & \highlight{-0.001} & \highlight{\textbf{0.009}} & \highlight{\textbf{0.057}} & \highlight{-0.009} & \highlight{-0.002} & \highlight{0.098} \\
        \hline
        \multirow{1}{*}{LwF} & 440B & 3B &  0.082 & \highlight{\textbf{-0.011}} & \highlight{\textbf{0.145}}  &  0.029 & \highlight{0.015} & \highlight{0.059} & 0.072 & \highlight{-0.001} & \highlight{0.099} \\
        \multirow{1}{*}{EWC} & 440B & 3B & 0.055 & \highlight{0.011} & \highlight{0.152} & 0.017 & \highlight{0.015} & \highlight{0.061} & 0.041 & \highlight{0.001} & \highlight{0.100} \\ \midrule[1pt]
        \rowcolor{lightgray!25} \multirow{1}{*}{Oracle Series} & 1.16T & 3B & -0.003 & 0.037 & 0.163 & 0.004 & 0.017 & 0.066 & -0.007 & 0.020 & 0.107 \\ 
        \bottomrule[1.2pt]
    \end{tabular}
    \vspace*{-4mm}
    \label{tab:loss-evals}
\end{table*}

\textbf{Training details.} We train 1B and 3B parameter models using OpenLM \citep{openlm}. Unless otherwise indicated, most of our results are for 3B models, where each method sees the same total number of 220B or 440B tokens, equivalent to 4$\times$ and 8$\times$ the Chinchilla optimal amount.~\footnote{Here, token counts are given by 20 $\times$ parameters $\times$ Chinchilla multiplier with a 1$\times$ multiplier being 
a near-optimal compute allocation found by \citet{hoffmann2022chinchilla}.} For 1B models, we only train at the 220B token scale to study how model size might affect the results (see \Cref{tab:loss-evals} and \Cref{app:model_size}).
For continual runs, we further assume that current practitioners are (a) likely to have access to more than enough data to train initial models; (b)  
unlikely to wait to obtain non-trivial performance. Hence, we front-load the 
total token budget such that 110B is allocated to an \textit{initial pretraining} on the 
first month (May-2013). Then, the remaining tokens are split equally among the 
other 113 continual timesteps. For our \textit{Oracle}-$t$ runs, each is trained on the same number of tokens that a 220B continual run observes by the end of month $t$, totaling 1.16T 
tokens for all seven Oracle models together.  Finally, to perform realistic hyperparameter selection, we follow takeaways from \citep{cha2024hyperparameterscontinuallearningreality} and only use the first 10 timesteps for tuning (see \cref{app:hyperparamters} for more details).

\textbf{Evaluation metrics.} 
Each run produces a $T_t \times T_e$ matrix of evaluations 
$E$ where $T_t,T_e$ are the total number of training/evaluation timesteps, 
$E_{i,j}$ is the performance of the model trained after seeing data up to 
month $i$
and evaluated on the month $j$. 
To control for inherent difficulty differences across evaluation months, we measure the \textit{regret} $R_{i,j} = E_{i,j} - E^*_{j}$ where 
$E^*_{j}$ is the performance of \textit{Oracle-2024-07} 
on month $j$.
Following \citet{garg2024tic}, we consider the following summary metrics, first defined assuming $T_t=T_e=T$ (deferring the discussion of the $T_t \neq T_e$ case to \cref{app:eval_summary_metrics}):

\begin{itemize}[leftmargin=*]
\vspace*{-2.5mm}
\itemsep0em
    \item In-distribution (ID) performance: averages along the matrix diagonal, i.e.,  $\sum_{i=1}^T = R_{i,i} / T$.  
    \item Backward transfer: averages the lower triangular of $R$, i.e., 
        $\sum_{i=1}^T\sum_{j<i}\frac{R_{i,j}}{T(T-1)/2}$, capturing how well continual checkpoints do on older months.
        
    \item Forward transfer: averages the upper triangular of $R$ analogously to backward transfer, capturing how well methods do on unseen future months.

\end{itemize}
\vspace*{-2.5mm}

\begin{figure*}[h!]
    \centering
\includegraphics[height=0.213\linewidth]{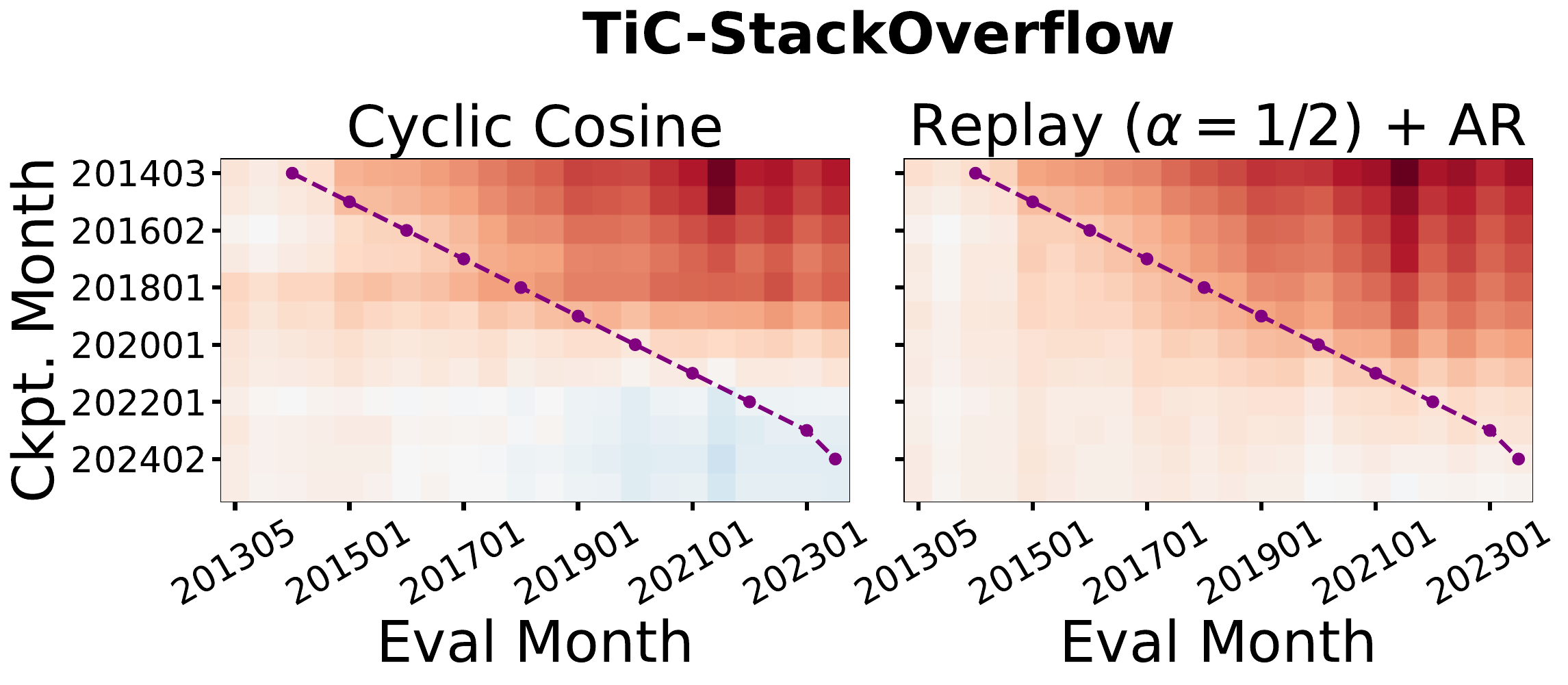} \hspace{-3mm} \includegraphics[height=0.212\linewidth]{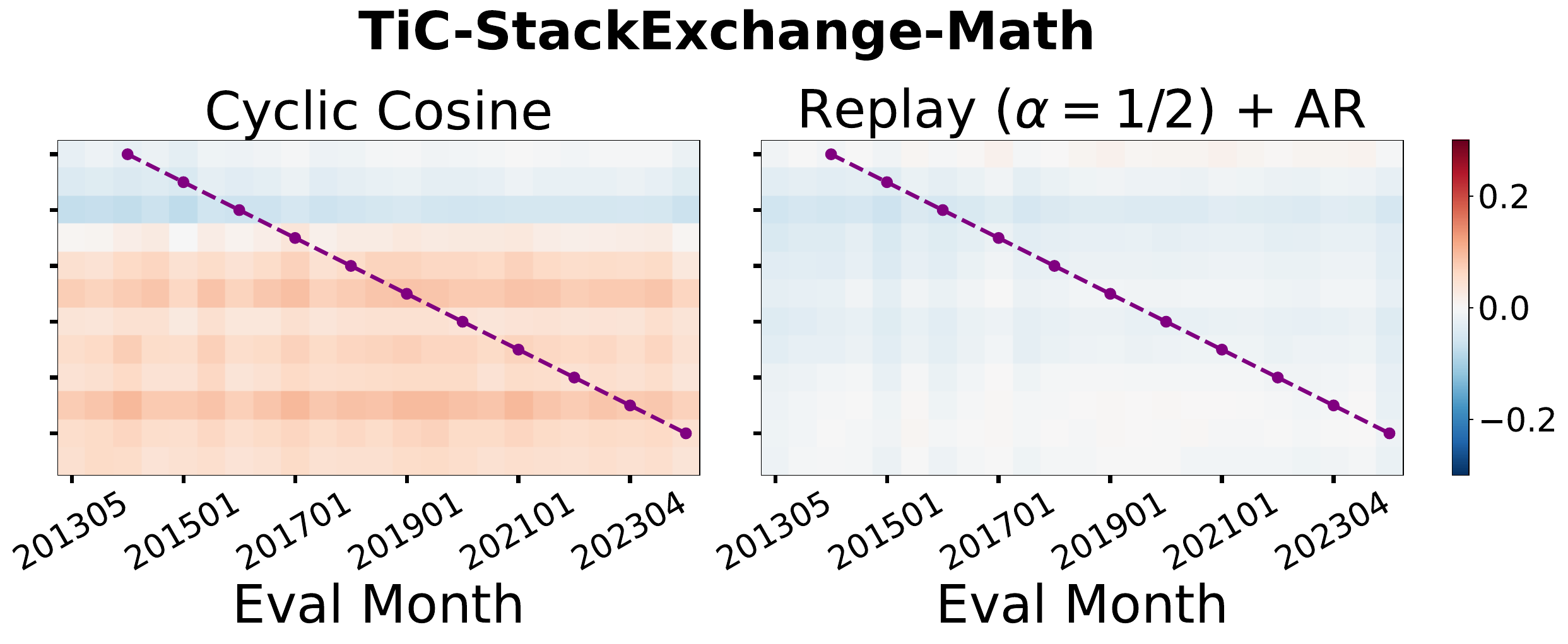}

\vspace*{-4pt}
    \caption{
    \textbf{Replay helps on \ticstackmath{} but hurts on domain that evolve more quickly such as \ticstackoverflow{}.} The purple lines trace out when the training and evaluation timestamps are closest. 
    }
    \vspace*{-5pt}
    \label{fig:downstream_heatmaps}
    \vspace*{-2pt}
\end{figure*}

\begin{figure}[h!]
    \centering
     \includegraphics[width=0.99\linewidth]{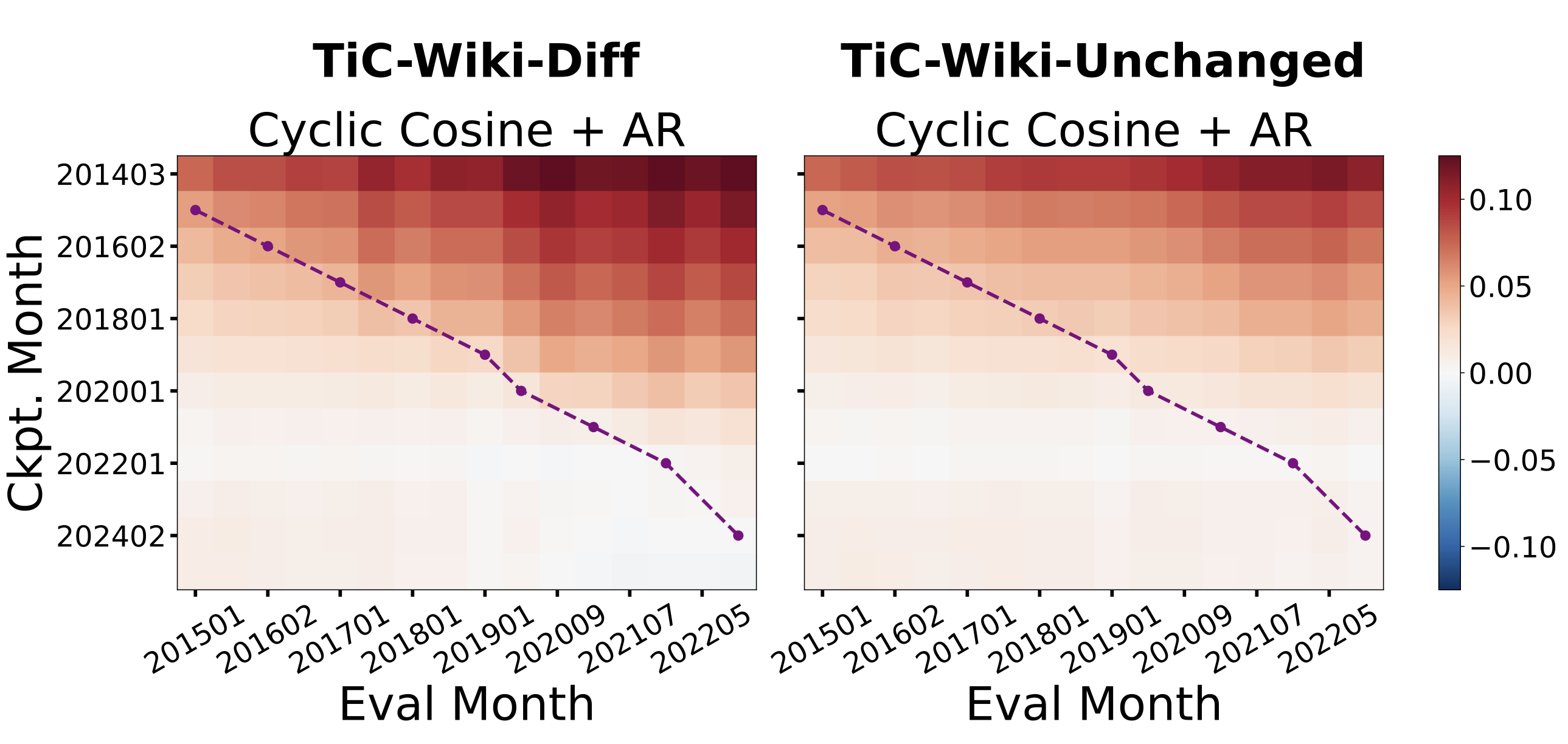}
    \caption{\textbf{On \ticwiki{}, performance on an evaluation month can peak years after the corresponding CC dump is seen.} This occurs even without replay. }
    \label{fig:ticwiki-main}
    \vspace{-3mm}
\end{figure}

\subsection{Held-out performance on \ticcc{} }

\begin{table*}[t!]
    \scriptsize
    \centering
    \caption{\textbf{Benchmarking continual methods on downstream evaluations (3B Models).} For all dynamic evaluations, we report perplexity relative to \textit{Oracle-2024-07} with log-scaling. \staticevals is an average of the accuracies of 22 downstream zero/few-shot tasks used by \citet{li2024datacomp}, evaluated only on the final model checkpoint (without scaling by an Oracle). \textbf{Bold} values are within one standard deviation (estimated with 3 runs of Cyclic Cosine) of the best in each column. \colorbox{green!15}{Highlighted} values indicate matching the  \colorbox{gray!25}{Oracle series} (within one standard deviation).}
    \vspace*{-2mm}
    \begin{tabular}{c|c|ccc|ccc|ccc|ccc}
        \toprule[1.2pt]
        \multirow{2}{*}{\textbf{Method}} & \multirow{2}{*}{\textbf{Tokens}} & \multicolumn{3}{c|}{\textbf{TiC-Wiki-Diff $\downarrow$}} & \multicolumn{3}{c|}{\textbf{TiC-Wiki-Unch. $\downarrow$}} & \multicolumn{3}{c|}{\textbf{ TiC-StackOverflow $\downarrow$}} & \multicolumn{3}{c}{\textbf{TiC-CD-PyTorch$\downarrow$}} \\ & 
        & Bwd. & ID & Fwd. & Bwd. & ID & Fwd. & Bwd. & ID & Fwd. & Bwd. & ID & Fwd.  \\
        \midrule[1pt]

        \multirow{2}{*}{Cyclic Cosine \stdev{(std)}} & \multirow{2}{*}{440B} & {0.018} & \highlight{0.026} & \highlight{0.072} & 0.026 & {0.031} & \highlight{0.056} & {0.017} & \highlight{\textbf{0.045}} & \highlight{\textbf{0.142}} & \highlight{-0.020} & \highlight{-0.002} & \highlight{0.175} \\[-3pt] & & \stdev{(0.000)} & \stdev{(0.000)} & \stdev{(0.001)} & \stdev{(0.001)} & \stdev{(0.000)} & \stdev{(0.001)} & \stdev{(0.001)} & \stdev{(0.002)} & \stdev{(0.002)} & \stdev{(0.004)} & \stdev{(0.001)} & \stdev{(0.002)} \\[+1pt]

        \multirow{1}{*}{Cyclic Cosine + AR} & 440B & \highlight{\textbf{0.010}} & \highlight{\textbf{0.023}} & \highlight{\textbf{0.071}} & \highlight{0.010} & \highlight{\textbf{0.022}} & \highlight{\textbf{0.054}} & 0.026 & \highlight{0.057} & \highlight{\textbf{0.142}} & \highlight{0.012} & \highlight{0.027} & \highlight{0.195} \\

        \multirow{1}{*}{Cyclic Rsqrt} & 440B & 
        \highlight{0.012} & \highlight{0.024} & \highlight{\textbf{0.070}} &
        \highlight{0.014} & \highlight{0.024} & \highlight{\textbf{0.053}} & 
        0.024 & \highlight{0.055} & \highlight{0.145} & 
        \highlight{-0.001} & \highlight{0.015} & \highlight{0.187} \\

        \multirow{1}{*}{ScheduleFree} & 440B & \highlight{0.012} & \highlight{0.025} & \highlight{0.073} & \highlight{0.014} & \highlight{0.025} & \highlight{0.056} & 0.038 & {0.066} & \highlight{\textbf{0.144}} & \highlight{0.030} & \highlight{0.044} & \highlight{0.197} \\ \hline

        \multirow{1}{*}{Replay ($\alpha=1/t$)} & 440B & 0.019 & 0.038 & \highlight{0.078} & \highlight{0.014} & \highlight{0.028} & \highlight{0.057} & 0.039 & 0.087 & 0.176 & 0.106 & 0.119 & 0.238 \\

        \multirow{1}{*}{Replay ($\alpha=1/2$)} & 440B & 0.015 & \highlight{0.029} & \highlight{0.073} & {0.017} & \highlight{0.027} & \highlight{0.055} & 0.027 & {0.065} & 0.158 & \highlight{0.020} & \highlight{0.032} & \highlight{0.189} \\

        \multirow{1}{*}{Replay ($\alpha=1/t$) + AR} & 440B & 0.015 & 0.036 & \highlight{0.077} & \highlight{\textbf{0.005}} & \highlight{0.023} & \highlight{0.056} & 0.057 & 0.102 & 0.179 & 0.141 & 0.151 & 0.261 \\

        \multirow{1}{*}{Replay ($\alpha=1/2$) + AR} & 440B & \highlight{\textbf{0.010}} & \highlight{0.027} & \highlight{0.073} & \highlight{0.007} & \highlight{\textbf{0.022}} & \highlight{\textbf{0.054}} & 0.036 & 0.074 & 0.160  & 0.058 & 0.070 & {0.214} \\ \hline

        \multirow{1}{*}{LwF} & 440B & {0.019} & \highlight{0.027} & \highlight{0.072} & 0.026 & {0.031} & \highlight{0.056} &{\textbf{0.014}} & \highlight{\textbf{0.047}} & \highlight{\textbf{0.142}} & \highlight{\textbf{-0.028}} & \highlight{\textbf{-0.011}} & \highlight{\textbf{0.171}}\\

        \multirow{1}{*}{EWC} & 440B & {0.015} & \highlight{0.030} & \highlight{0.074} & {0.016} & \highlight{0.029} & \highlight{0.058} & 0.039 & {0.073} & 0.155 &  0.055 & 0.069 & {0.211} \\ \midrule[1pt]

        \rowcolor{lightgray!25} \multirow{1}{*}{Oracle Series} & 1.16T & 0.014 & 0.035 & 0.080 & 0.013 & 0.030 & 0.061 & 0.012 & 0.056 & 0.146 & 0.035 & 0.057 & 0.196  \\
         
 \bottomrule[1.2pt]
 \end{tabular}

\vspace{0.25em} %

\begin{tabular}{c|c|ccc|ccc|ccc|ccc}
\toprule[1.2pt]
        \multirow{2}{*}{\textbf{Method}} & \multirow{2}{*}{\textbf{Tokens}} & \multicolumn{3}{c|}{\textbf{TiC-StackE-Math $\downarrow$}} & \multicolumn{3}{c|}{\textbf{ TiC-CD-NumPy $\downarrow$}} & \multicolumn{1}{c}{\textbf{Static Evals. $\uparrow$}} \\
        & & Bwd. & ID & Fwd. & Bwd. & ID & Fwd. & \staticevals{} (DCLM)\\
        \midrule
        \multirow{2}{*}{Cyclic Cosine} & \multirow{2}{*}{440B} & 0.048 & 0.030 & 0.013 &  0.054 & 0.070 & 0.065 & 49.6 \\[-3pt] & & \stdev{(0.002)} & \stdev{(0.002)} & \stdev{(0.000)} & \stdev{(0.002)} & \stdev{(0.001)} & \stdev{(0.002)} & \stdev{(0.3)} \\[+1pt]

        \multirow{1}{*}{Cyclic Cosine + AR} & 440B & 0.023 & 0.008 & -0.001 & 0.044 & 0.057 & 0.055 & 48.8\\

        \multirow{1}{*}{Cyclic Rsqrt} & 440B & 0.032 & 0.017 & 0.006 & 0.050 & 0.061 & 0.060 & 49.4 \\

        \multirow{1}{*}{ScheduleFree} & 440B & 0.036 & 0.023 & 0.010 & 0.081 & 0.091 & 0.072 & 49.4 \\ \hline

        \multirow{1}{*}{Replay ($\alpha=1/t$)} & 440B & \highlight{-0.025} & -0.025 & \highlight{\textbf{-0.022}} & 0.035 & 0.038 & \textbf{0.041} & 49.4 \\

        \multirow{1}{*}{Replay ($\alpha=1/2$)} & 440B & 0.000 & -0.008 & -0.011 & \textbf{0.029} & 0.038 & 0.044 & \textbf{50.1} \\

        \multirow{1}{*}{Replay ($\alpha=1/t$) + AR} & 440B & \highlight{\textbf{-0.028}} & \highlight{\textbf{-0.027}} & \highlight{\textbf{-0.022}} & 0.031 & \textbf{0.032} & 0.047 & 49.3\\

        \multirow{1}{*}{Replay ($\alpha=1/2$) + AR} & 440B & -0.015 & -0.019 & -0.017  & 0.036 & 0.039 & \textbf{0.043} &  49.3 \\ \hline

        \multirow{1}{*}{LwF} & 440B  & 0.049 & 0.031 & 0.013 & 0.059 & 0.077 & 0.070 & 49.6\\ 

        \multirow{1}{*}{EWC} & 440B & 0.014 & 0.004 & 0.000 & 0.063 & 0.066 & 0.059 & 49.0 \\ \midrule[1pt]

        \rowcolor{lightgray!25} Oracle series & 1.16T & -0.025 & -0.028 & -0.022  & 0.008 & 0.008 & 0.015 & 50.6 \\ 
        
        \bottomrule[1.2pt]
    \end{tabular}
    \label{tab:downstream-evals}
    \vspace*{-9pt}
\end{table*}

In \cref{tab:loss-evals} and \cref{fig:ticcc_main}, we first explore how well continual methods learn and retain various parts of our \textit{general web-scale training distribution}, \ticcc{}. Overall, we observe significant trade-offs between ID and backward transfer, with key findings below:

\textbf{%
Continual pretraining outperforms the Oracle series with 62\% less compute on \ticcc{}.} Replay ($\alpha_t$ = 1/2) + AR at 440B tokens  outperforms the Oracle series on almost all metrics, coming within 0.0001 on backward transfer. 
From our 220B runs, we see that gaps do remain between (token-matched) continually trained models at timestamp $t$ and Oracle-$t$. The key though is that by reusing models, continual learning allows for more frequent checkpoints (that have seen more tokens) while maintaining cheaper total costs.

\textbf{%
Cyclic Cosine achieves the best ID performance on \ticcc{} but also the most forgetting.}
As shown by \cref{tab:lr_tuning} in \cref{app:hyperparamters}, the best maximum learning rate (LR) per
cycle is 1e-4, notably 30$\times$ smaller than that used for the initial May-2013 pretraining. This differs from \citet{ibrahim2024simple, gupta2023continual} who suggest rewarming up to a similar LR, likely because our continual rounds are smaller and more frequent. Indeed, higher LRs degraded all metrics, while lower LRs improved backward transfer at the cost of ID. Compared to 1e-4, the AR schedule marked the opposite end of the spectrum, offering the best backward transfer compared to setting any fixed maximum LR.

\textbf{%
Replay is essential for addressing forgetting.} Based on backward transfer and \cref{fig:cc_extended_heatmaps_220b,fig:cc_extended_heatmaps_440b}, all non-replay methods show significant forgetting at later checkpoints. Optimizer tweaks and EWC can somewhat reduce forgetting by sacrificing some ID performance. However, additionally applying replay can further improve backward transfer by 60\% for 220B runs, a gap which only \textit{widens} as we scale up to 440B tokens. Between replay variants, $\alpha_t=1/t$ achieves the lowest forgetting but $\alpha_t = 1/2$ offers a better practical trade-off on \ticcc{}, resulting in marginally worse Backward but substantially better ID. This differs from TiC-CLIP's recommendation of $\alpha_t = 1/t$, likely since decreasing the ratio of new data becomes problematic in our setup which uses 10$\times$ more timesteps.

\textbf{%
Forgetting remedies may not sacrifice ID on specific subsets of web-data.}
Comparing ID and forward transfer, Wiki appears to evolve more slowly than News, with both changing less rapidly than \ticcc{}. Also, as shown in \cref{fig:token_counts} (\Cref{app:data_pipeline}), the \textit{prevalance} of specific domains can vary over time (in addition to their contents), with sharp drop-offs in 2017 for both subsets. This may explain why at 440B scale, AR schedules and replay can outperform Cyclic Cosine's ID performance on \ticccwiki{} and \ticccnews{}, despite being worse for ID on full \ticcc{}.

\textbf{%
For smaller models, forgetting is more of a challenge but method rankings remain similar. } We observe that ID for 1B methods (relative to the 1B Oracle) is quite close to ID for 3B methods (relative to the 3B Oracle). However, Backward metrics are worse at 1B especially when not using replay. This suggests that, perhaps as expected, models struggle more to retain older knowledge when they have fewer parameters. Overall though, we obtain similar conclusions between the two scales in terms of what methods best mitigate forgetting: increasing model size can slightly improve relative Backward regret but not nearly as much as switching from non-replay to replay methods.

\subsection{Downstream evaluations}

Here, \cref{tab:downstream-evals} and \cref{fig:downstream_heatmaps} show results for our downstream evaluations. 
In contrast to \ticcc{} evaluations, LLMs are often evaluated on curated tasks that may not be completely nor temporally aligned with CC dumps as outdated data can exist in newer dumps \citep{cheng2024dated}. Hence, we observe that methods exhibit trade-offs \textit{across different evaluations}, highlighting the challenges of performing on all domains while pretraining on general web-data. 

\vspace*{-1pt}

\textbf{%
Continual methods outperform Oracle re-training on many downstream evaluations.}
While Replay ($\alpha_t = 1/2$) + AR at 440B tokens outperforms the Oracle series on \ticcc{}, the comparison is more nuanced for downstream tasks. Many continual methods show significant gains on \ticwiki{} evaluations, while the Oracle series maintains an edge on \ticdocsnumpy and \ticstack{}-English (see \cref{app:extended_results}). Meanwhile, methods like Cyclic Cosine outperform the Oracles on \ticstackoverflow{} and \ticdocstorch{} but fall short on \ticstackmath{} where only Replay $(\alpha=1/t)$ can do so.

\textbf{%
Replay may underperform on faster-evolving domains but helps when evolution is slower and older dumps contain more relevant data.}
For some slowly-evolving domains like \ticstackmath{}, earlier CC dumps (pre-Feb-2016) provide the most value, making both replay and AR schedules beneficial (see \cref{fig:downstream_heatmaps}).
In contrast, for \ticstackoverflow{}, larger performance gradients exist across time and using less historical data improves all metrics.
A similar pattern is evident in \ticdocs, where replay improves all metrics for NumPy (released in 1995) but harms performance for PyTorch (released in 2016). The heatmaps in \cref{app:extended_results} show that Cyclic Cosine models initially improve on NumPy up to 2016 before forgetting it until 2024, suggesting replay is necessary to retain this knowledge (which features most prominently in earlier dumps). Conversely, replay hurts for PyTorch by overemphasizing data that was crawled from before its release.

\vspace{-1mm}
\textbf{%
Forgetting earlier CC dumps may have less impact on general factual knowledge.}
On \ticwiki{}, replay shows surprisingly different behavior than on \ticcc{}-Wiki. Non-replay methods remain competitive not just for ID but also for backward transfer, even on \ticwiki{}-Unchanged. \cref{fig:ticwiki-main} reveals that \ticwiki{} performance often peaks years after the corresponding CC dump, even without replay. This suggests two key insights: (1)
by focusing on specific segments and proper nouns rather than all tokens, \ticwiki{} may better isolate persistent factual knowledge  rather than temporal differences such as formatting changes; (2) knowledge captured in \ticwiki{} can be effectively learned from later CC dumps, possibly due to delayed alignment between CC's Wikipedia crawls and \ticwiki{}'s comprehensive coverage.

\textbf{%
On static evaluations, continual methods leave room for improvement.} As shown in \cref{tab:downstream-evals}, most continual runs perform similarly on \staticevals tasks from \citet{li2024datacomp}, with a persistent gap to Oracle-2024-07. 
The remaining difference could stem from either the Oracle's unrestricted data access or our initialization bias toward May-2013 data. This initialization achieves a score of 48.5 while starting from it and training simultaneously on all 113 remaining months achieves 49.9, landing between our best 220B (\cref{app:extended_results}) and 440B runs.

%% file: sec/conclusion.tex
    \vspace*{-4pt}

\section{Conclusion}
\vspace*{-3pt}
In this work, we introduced a benchmark uniquely scoped for web-scale continual LLM pretraining, consisting of 114 timesteps of Common Crawl training data and several time-stratified
evaluations in the form of \ticwiki{}, \ticstack{}, and 
\ticdocs{}.
Using these assets, we highlighted key observations about  balancing forgetting and plasticity on general web-data as well as how this translates to nuanced trade-offs across different evaluation domains. 
Overall, we view our work as a pivotal step towards understanding how to best continually pretrain LLMs and developing even more efficient continual methods.

%% file: sec/sup_dataset.tex
\section{Dataset Construction}\label{app:data_pipeline}

\begin{figure*}[h!]
\centering
    \includegraphics[width=0.9\linewidth]{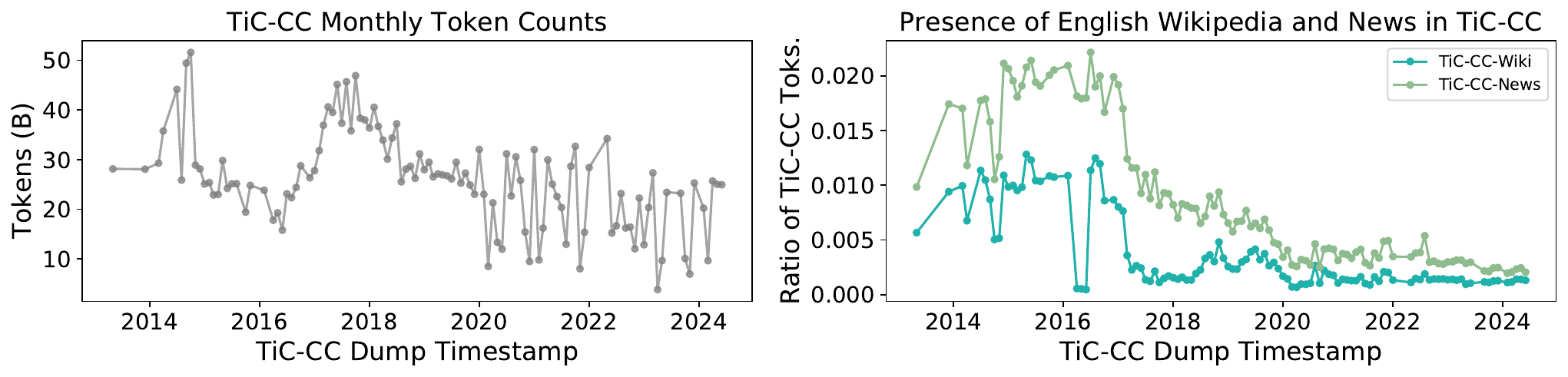}
\caption{ We plot the total number of tokens per month in \ticcc{} 
    (left) as well as the proportion of those tokens coming from our 
    \ticccwiki{} and \ticccnews{} subsets (right).}
    \vspace*{-3pt}
\label{fig:token_counts}
\end{figure*}

We build upon the existing pipeline and assets from DataComp-LM 
\citep{li2024datacomp} to build our dataset, only altering steps that rely on global operations across months.

\textbf{Initial pool and temporal splitting.}
We start with DCLM-Pool~\citep{li2024datacomp} which is available by CC-BY-4 license and contains all CC dumps 
between May-2013 and December-2022. The only pre-processing that has been done 
on this pool is to parse the HTML (contained in WARC files of CC) into 
plaintext for each webpage via the open-source \texttt{resiliparse} 
library~\citep{bevendorff2018,bevendorff2021c} with the \texttt{main\_content} flag set to \texttt{True} ~\footnote{We use 
\texttt{readability} for parsing code documentations in our \ticdocs{} (\url{https://github.com/mozilla/readability}).}.
In DCLM-Pool, documents are 
naturally grouped together into files based upon the CC dump, which is 
indicated by the file prefix.
To split the data by month, we simply group files that share the same prefix.  
Since DCLM-Pool contains data up to December-2022, we also follow their exact 
download and extraction scripts to obtain more recent data until July-2024.

\textbf{Data preprocessing and tokenization.}
Next, we follow  DCLM-Baseline's filtering procedure which starts with their 
implementation of heuristic filters from RefinedWeb \citep{refinedweb}. We apply these filters, which includes English filtering, 
independently on each page with no change. However, we have to modify 
deduplication that removes nearly identical lines and pages.  
Instead of applying deduplication globally across months as in DCLM-Baseline, we apply the 
same deduplication method \textit{only within} each month.
Finally, we also skip the final classifier-based filtering in DCLM-Baseline, as 
their classifier was trained on data that comes from all months, including 
examples generated by recent LLMs such as GPT-4.  

\begin{figure}[h!]
    \centering
    \includegraphics[width=0.6\linewidth]{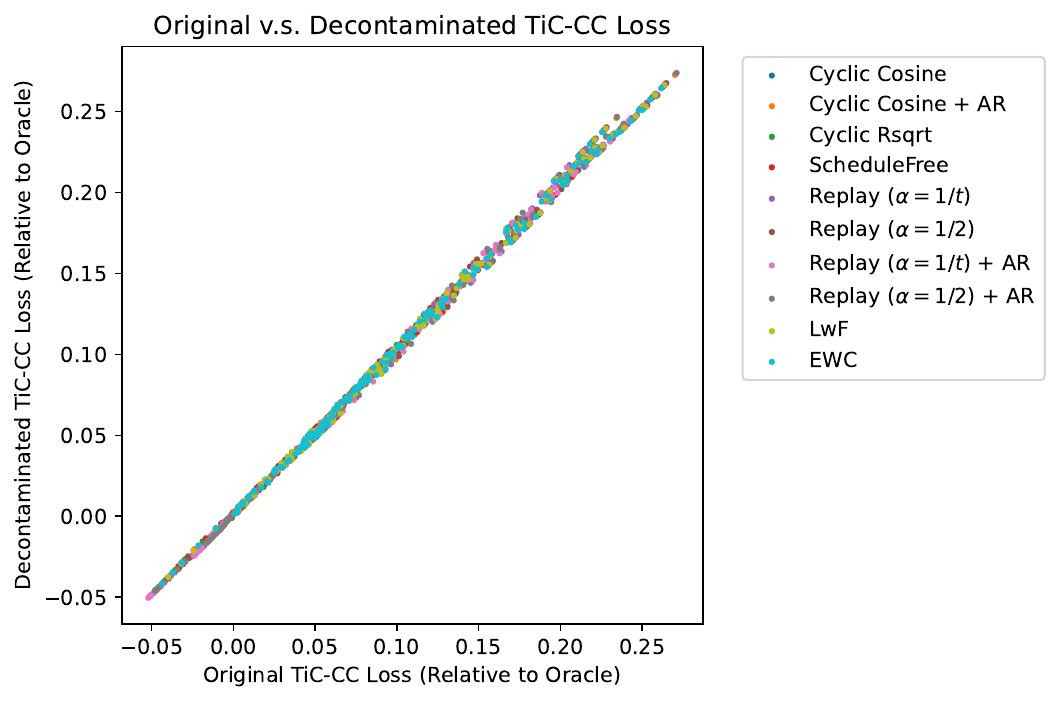}
    \caption{\textbf{Findings from \ticcc{} are robust to potential data leakages.} We create a decontaminated version of our \ticcc{} loss-based evaluation by deduplicating each month's evaluation set using a Bloom Filter pre-populated by the corresponding training set. Overall, across all the methods, checkpoints, and evaluation months we observe strong correlations between using the pre-decontamination (x-axis) and post-decontamination (y-axis) losses (relative to the Oracle).} 
    \label{fig:tic_cc_contamination}
\end{figure}

\textbf{Data sampling and held-out sets.}
DCLM-Pool was partitioned randomly into 10 equally sized ``global shards''.  
Within our monthly splits, we also maintain the original global shard 
assignments. For our training scales, using just one of these global shards 
within each month is sufficient. Notably though, when we construct evaluation sets 
such as in (\cref{sec:loss_evals}), we make sure to sample 
from a different global shard than the one used for training. This ensures the 
evaluation data is a sampled from the same distribution as the training data 
while also being \textit{mostly} held out.  Notably, since we do not deduplicate 
across globals shards or months, there could be overlap between training and 
eval sets across months. However, we observe from \cref{fig:tic_cc_contamination} that potential data leakages are unlikely significantly change relative losses values (compared to the Oracle).
For each validation set, we cap the maximum number of tokens to $16.7$M which 
corresponds to 8192 sequences for our context length of 2048. For some months 
of \ticccwiki{} and \ticccnews{}, we end up with less than this amount, but the 
smallest are 5M and 12M respectively.

%% file: sec/sup_evals.tex
\clearpage
\section{Details of Evaluations}

\subsection{\ticwiki{}}
\label{sec:tic_wiki_sup}

We construct \ticwiki{} from English Wikipedia (available under CC-BY-SA 4.0 license) and Wikidata (available under CC0 license) which are sister projects 
from the non-profit Wikimedia Foundation. Wikidata is a structured knowledge 
graph that stores the structured data of Wikipedia and other sister projects.  
Data on Wikidata is represented in the form of statements in the form of 
property-value about an item in the simplest form. For example, ``Mount Everest 
is the highest mountain in the world'' is represented as Earth (Q2) (item) 
$\rightarrow$ highest point (P610) (property) $\rightarrow$  Mount Everest 
(Q513) (value)~\footnote{\url{https://www.wikidata.org/wiki/Help:About_data}}.  
The triplet (item, property, value) can also be referred to as (subject, 
relation, object).

\textbf{TemporalWiki dataset generation.}
TemporalWiki \citep{jang2022temporalwiki} constructs two forms of evaluations from monthly snapshots of English Wikipedia 
and Wikidata, which they refer to as intrinsic and extrinsic evaluations. Overall, they are created through the following steps:
\begin{enumerate}
\itemsep-1pt
    \item Generating TWiki-Diffsets by identifying changes and additions between 
        consecutive snapshots of Wikipedia. For new articles, the entire 
        article is added to the Diffset while for existing articles, only the 
        changed or new paragraphs are added. In their study, measuring perplexity on proper nouns separately from TWiki-Diffsets and Unchanged Wikipedia pages serves as intrinsic evaluations (named as such since their study uses these datasets also as their training distributions). 
    \item Constructing TWiki-Probes by processing two consecutive snapshots of 
        Wikidata. Statements are categorized into changed if the property/value 
        has changed or categorized into unchanged otherwise. In their extrinsic evaluation, perplexity is measured on concatenated versions of the (subject, relation, object) triplets (using space delimiting).

    \begin{itemize}
    
        \item Aligning TWiki-Probes with TWiki-Diffsets to ensure changed statements in Twiki-Probes
        exist in TWiki-Diffsets and unchanged statements exist in unchanged parts of Wikipedia.
        \item Further heuristic filtering of TWiki-Probes by removing statements where the subject or object is a substring of the other or the object is more than 5 words. Moreover, a single subject is limited to maximum 1\% and relation/object is limited to maximum 5\% of the total statements. 
    \end{itemize}
    
\end{enumerate}

In \ticwiki{} we extend and modify TemporalWiki as follows:

\begin{enumerate}
\itemsep-1pt
    \item We expand the timespan from four months to a decade (2014-2024), thus capturing a broader spectrum of knowledge evolution.

    \item We construct both a perplexity-based evaluation and a QA-based evaluation. 

    \begin{itemize}

    \item Our perplexity-based evaluation follows their intrinsic evaluation protocol. In our case, this evaluation is no longer fully intrinsic since  although some Wikipedia pages exist in Common Crawl, we do not directly train on whole Twiki-Diffsets or Wikipedia dumps at each timestep. Given this, we also chose to carry out our perplexity evaluation without aligning to Wikidata, since Wikidata's knowledge graph does not have full coverage of Wikipedia~\citep{mousavi2023construction} and the temporal alignment between their dumps is not always perfect. 
    
    \item In place of their extrinsic evaluation, we create natural question-answer (QA) evaluations with an improved matching process of Wikipedia and Wikidata dumps, and enhancing the robustness of data parsing to format changes over time. 

    \end{itemize}

\end{enumerate}

For our perplexity evaluations, our \ticwiki{} evaluation consists of 61 months each with 10k changed and 10k unchanged Wikipedia sentences/paragraphs. To illustrate the characteristics of our generated dataset, we present key statistics in the following figures. Figure \ref{fig:wikipedia-changes} shows the number of Wikipedia pages with significant changes between consecutive database dumps over time. This graph provides insight into the volume and temporal distribution of our data generation process, highlighting periods of higher and lower content modification as well as distribution of our dumps. 

For QA evaluations, our models trained in \cref{sec:experiments} overall achieved fairly low performance and thus continual trends tended to be noisy. Hence, we focused on the perplexity evaluations in this work. Nevertheless, we discuss our construction for QA examples and release code for generating question-answers as we believe it may be helpful for future studies.

\subsubsection{Data Download}
\textbf{Archive dumps.} Wikimedia releases regular 
dumps~\footnote{\url{https://dumps.wikimedia.org/wikidatawiki/}}\textsuperscript{,}\footnote{\url{https://dumps.wikimedia.org/enwiki/}},
but only retains data for the most recent 4 months.
To access historical data, we utilized the Internet 
Archive~\footnote{\url{https://archive.org}}. The earliest available dump dates 
back to November 2014. It is important to note that the archived dumps do not 
cover every month, with several months missing from the record. In our study, 
we made use of all available monthly dumps.
The filenames of the dumps include the specific date of month that has been 
collected on, which is typically the 1st or 20th of the month, though this can 
vary. We include only one dump per month if multiple dumps are 
available. We check for the first date if not available look for 20th and if neither we start from beginning the month and check for the first available date in that month.%

\textbf{Wikipedia historical dumps.}
It is possible to reconstruct each version of Wikiepdia using the large history files Wikipeida provide~\footnote{\raggedright\url{https://dumps.wikimedia.org/enwiki/latest/} file names containing \texttt{pages-meta-history}.}. There are more than 200 historical dumps of English Wikipedia, each sized more than 2GB. Combined together, these files include all revisions and all pages of Wikipeida.

For Wikidata, Wikimedia does not provide historical diff files as Wikipedia except for the last three months~\footnote{\url{https://dumps.wikimedia.org/wikidatawiki/}}.
Wikidata file names are formatted similar to \texttt{wikidatawiki-20190101-pages-articles.xml.bz2}
and available at URLs similar to \url{https://dumps.wikimedia.org/wikidatawiki/20240401/}.

Each Wikidata dump is approximately 140GB whereas each Wikipeida dump is less than 22GB. Therefore, it is possible to make a version of Wikipedia that keeps track of all changes which results in 200 files of 2GB. But as far as we know there are no such files for Wikidata.

Using the dumps from \url{archive.org} has several advantages:
\vspace{-5pt}
\begin{itemize}
\itemsep0em
\item We make sure that we do not leak information from previous timesteps.
\item There exists a Wikidata dump close to each Wikipedia dump to be aligned.
\item We can use Wiki-Extractor for filtering and remove Wikipeida editorial discussions.
\end{itemize}

\subsubsection{Data preprocessing for perplexity evaluations}

We construct perplexity evaluations based on raw Wikipedia diffsets. For these evaluations, we do not rely on an alignment with Wikidata which increases the diversity of \ticwiki{} evaluation with a simpler construction.

\textbf{Data cleanup.} We utilize 
WikiExtractor~\footnote{\url{https://github.com/attardi/wikiextractor}} to 
clean up the Wikipedia data. This step extracts the main content and removes 
extraneous and non-essential characters. 

\textbf{Wikipedia diffsets.}
To construct consecutive diffs of Wikipedia, we developed a method comparing 
snapshots of articles from consecutive dumps. For comparing two snapshots of an 
article, we first remove extraneous whitespace and standardize formatting by 
preprocessing the text. This involves removing empty lines, stripping newline 
characters, and creating a normalized version of each line where punctuation is 
removed and text is converted to lowercase.

Afterward, we use a two-level comparison: first at the paragraph level, then at 
the sentence level for changed paragraphs. We utilize Python's 
\texttt{difflib.SequenceMatcher} to compare the normalized versions of 
paragraphs and sentences. This hierarchical method, coupled with normalization, 
captures substantial edits while filtering out minor or stylistic changes.

We extract and store both changed and unchanged content separately. Changed content includes replaced paragraphs with modified sentences and newly inserted paragraphs. Unchanged content preserves paragraphs and sentences that remain identical between versions. New articles are treated as entirely changed content.  This approach allows us to focus on meaningful content changes while 
maintaining the context of unchanged information, providing a comprehensive 
view of how Wikipedia articles evolve over time.
\Cref{alg:wiki_diffs,alg:wiki_changed} describe the process of constructing 
Wikipedia diffs and changed/unchanged content.

\textbf{Evaluation.} We create perplexity evaluations from sentences/paragraphs in the changed/unchanged Wikipedia diffs by implementing them in the format of LLM-Foundry \citep{mosailMLarithmetic}. We also add custom evaluation code to be able to evalute only on proper nouns.

\subsubsection{Data preprocessing for QA evaluations}

We further construct QA evaluations by extending the alignment with Wikidata.
In TemporalWiki, the evaluation was constructed from triplets (subject, relation, object) by concatenating the triplet and measuring perplexity. However, this format will often not result in natural/proper sentences (e.g., missing a verb). For example, we can have a concatenated triplet such as ``Florida State Seminoles women's basketball instance of basketball team". As such, a model trained only on natural language may assign an overall lower probability to the object. We construct question-answer pairs described in natural language. Note that the results presented in \cref{sec:experiments} are limited to our perplexity evaluations only but we release code for generating QA evaluations as well.

\textbf{Wikidata diffsets.} Next, we extract changed and unchanged Wikidata 
statements of the form (subject, relation, object) from each consecutive dump.  
Identical triplets in both dumps are marked as unchanged.  Triplets in the new 
dump not present in the old are categorized as new, with the exception that if 
a subject entity
has more than 10 triplets, the algorithm 
randomly samples 10 to represent it.  When a triplet has the same subject and 
relation as one in the old dump but a different object and the old and new 
objects differ only in case (upper/lowercase), the triplet is classified as 
unchanged; otherwise, it is categorized as new.  Triplets from the old dump not 
found in the new one are implicitly considered removed, but importantly, these 
are not included in the output sets of changed or unchanged triplets.  
Throughout this process, the algorithm filters out triplets with overly long 
object values (more than 5 words) and ensures no duplicates are added. This 
approach efficiently tracks Wikidata evolution, capturing nuanced changes while 
managing the volume of data for new entities.

\Cref{alg:wiki_triplet} describes the process of triplet extraction. In TemporalWiki, a different set of hard-coded rules are used to extract the triplets from Wikidata which perform unreliably for older data. Our approach instead systematically parses Wikidata dumps, which ensures greater efficiency and robustness. TemporalWiki also use web API requests to link Wikipedia pages with Wikidata IDs. However, in our approach we eliminate reliance on web API requests, our method significantly reduces processing time and avoids potential API limitations or downtime.

\textbf{QA construction.} We constructed question-answer pairs in natural language by utilizing Wikipedia sentences. For each triplet, we find a sentence in the Wikipedia page of the subject that mentions both the subject and the object. Then we replace the object with blank and add the additional prefix of ``Question: Fill in the blank:'' and the suffix of ``Answer:''. This process results in a natural fill-in-the-blank question for each triplet.

\textbf{Evaluation.} We implement QA evaluations again using LLM-Foundry \citep{mosailMLarithmetic}, using their \texttt{InContextLearningGenerationExactMatchAccuracy} evaluation metric. Each QA sample is contains a context, answer, and aliases. The following is an example QA:

\begin{figure}[!ht]
\begin{lstlisting}[language=JSON, frame=single]
{
    "context":
        "Question: Fill in the blank: "
        "Douglas Adams' most notable work is _______."
        "Answer:",
    "answer": "The Hitchhiker's Guide to the Galaxy",
    "aliases": ["A list of possible aliases for the answer"],
}
\end{lstlisting}
\caption{Example of \ticwiki{} for QA evaluations.}
\end{figure}

\clearpage

\begin{algorithm}
\caption{Construct Wikipedia Consecutive Diffs}
    \label{alg:wiki_diffs}
\begin{algorithmic}[1]
\STATE Input: oldSnapshot, newSnapshot
\STATE Output: changedContent, unchangedContent
    \STATE {oldArticles} $\gets$ ReadArticles({oldSnapshot})
    \STATE {newArticles} $\gets$ ReadArticles({newSnapshot})
    \STATE {changedContent} $\gets \emptyset$, {unchangedContent}$\gets \emptyset$
    \FOR{each {articleId }in {newArticles}.{keys}}
    \IF{articleId in {oldArticles}}
        \STATE oldText $\gets$ NormalizeText({oldArticles}[{articleId}].{text})
        \STATE newText $\gets$ NormalizeText({newArticles}[{articleId}].{text})
        \STATE {changed }$\gets$ ExtractChangedContent({oldText}, {newText})
        \STATE {unchanged }$\gets$ ExtractUnchangedContent({oldText}, {newText})
        \STATE Add ({articleId}, {changed}) to {changedContent}
        \STATE Add ({articleId}, {unchanged}) to {unchangedContent}
        \ELSE
        \STATE Add ({articleId}, {newArticles}[{articleId}].{text}) to {changedContent}
        \ENDIF
\ENDFOR
    \RETURN {changedContent}, {unchangedContent}
\end{algorithmic}
\end{algorithm}

\begin{algorithm}
\caption{Extract Changed Content}
    \label{alg:wiki_changed}
\begin{algorithmic}[1]
\STATE Input: $oldText$, $newText$
\STATE Output: $changedContent$
\STATE oldParagraphs  $\gets$ SplitIntoParagraphs(oldText)
\STATE newParagraphs  $\gets$ SplitIntoParagraphs(newText)
\STATE changedContent $\gets \emptyset$
\FOR{each (oldPara, newPara) in Zip(oldParagraphs, newParagraphs)}
    \IF{IsDifferent(oldPara, newPara)}
        \STATE oldSentences $\gets$ SplitIntoSentences(oldPara)
        \STATE newSentences $\gets$ SplitIntoSentences(newPara)
        \FOR{each (oldSent, newSent) in Zip(oldSentences, newSentences)}
            \IF{IsDifferent(oldSent, newSent)}
                \STATE Add newSent to changedContent
            \ENDIF
        \ENDFOR
    \ENDIF
\ENDFOR
\RETURN changedContent
\end{algorithmic}
\end{algorithm}

\begin{algorithm}
\caption{Wikidata Triplet Extraction and Categorization}
    \label{alg:wiki_triplet}
\begin{algorithmic}
\REQUIRE oldDump, newDump
\ENSURE unchanged, new

\STATE unchanged $\leftarrow \{\}$
\STATE new $\leftarrow \{\}$
\STATE newEntities $\leftarrow \{\}$

\FORALL{triplet $\in$ newDump}
    \IF{triplet $\in$ oldDump}
        \STATE Add triplet to unchanged
    \ELSIF{hasSameSubjectPredicate(triplet, oldDump)}
        \STATE oldObject $\leftarrow$ getObject(triplet.subject, triplet.predicate, oldDump)
        \IF{equalsIgnoreCase(triplet.object, oldObject)}
            \STATE Add triplet to unchanged
        \ELSE
            \STATE Add triplet to new
        \ENDIF
    \ELSE
        \IF{triplet.subject $\notin$ oldDump}
            \STATE Add triplet to newEntities[triplet.subject]
        \ELSE
            \STATE Add triplet to new
        \ENDIF
    \ENDIF
\ENDFOR

\STATE sampleNewEntityTriplets(newEntities, new)
\STATE filterLongObjects(unchanged, new)
\STATE removeDuplicates(unchanged, new)

\RETURN unchanged, new
\end{algorithmic}
\end{algorithm}

\begin{figure}[htbp]
    \centering
    \includegraphics[width=\textwidth]{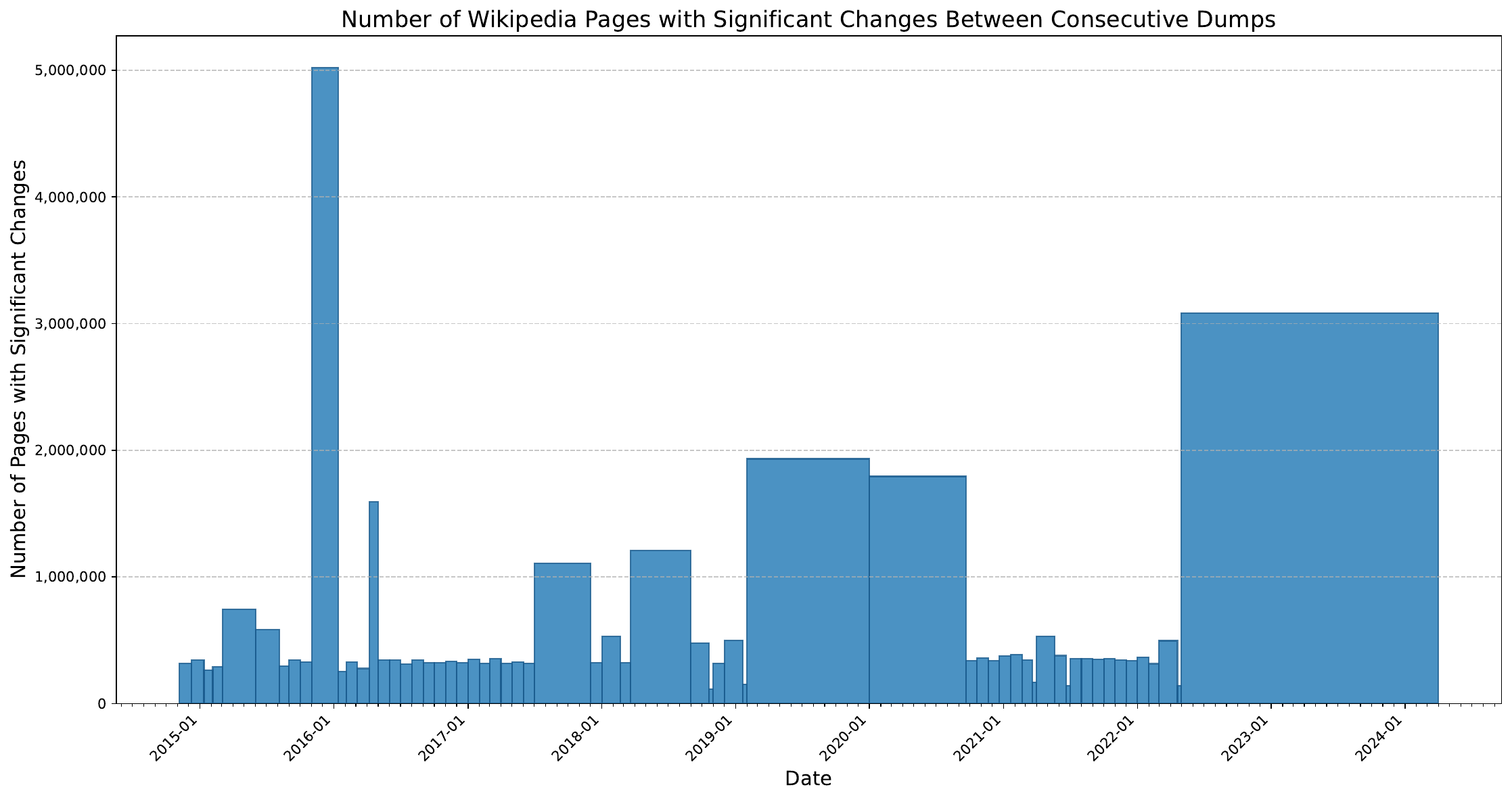}
    \caption{Number of Wikipedia pages with significant Changes between 
    consecutive \texttt{archive.org} dumps.}
    \label{fig:wikipedia-changes}
\end{figure}

\clearpage

\subsection{\ticstack{}}
\label{sec:tic_stack_sup}

\subsubsection{Data preprocessing}

\ticstack{} spans data from July 2008 through April 2024. The data was sourced from \url{archive.org} using the April 2024 dump of 
StackExchange (available by CC-BY-SA 4.0). 
Each category 
in the dump comes with two key files: \texttt{Post.xml} and 
\texttt{PostHistory.xml}.  \texttt{Post.xml} contains information on how 
answers and questions relate to each other and includes the latest text for 
each post entry. \texttt{PostHistory.xml} records the changes to each post, 
whether it is a question or an answer.

To construct our dataset, we first build the graph of question-answer 
relationships based on the \texttt{Post.xml}. We then use 
\texttt{PostHistory.xml} to reconstruct exact snapshots of posts at specific 
timestamps. This allowed us to capture the state of each post at the end of 
each month, ensuring our data reflected the actual content available at those 
points in time. 
We pick out the answers that were accepted by the user who posted the original question and which received at least 4 $\times$ the number of up-votes compared to another answer.

We use the StackOverflow, Mathematics, and English Language \& Usage categories in this work but our code can also be used to process any additional categories from the overall 182 categories.
Some categories had insufficient 
questions in a single month to provide statistically significant results. In 
such cases, we combined data from consecutive months, ensuring that each time 
frame contains at least 500 questions.

Given the question title and body as the query, we measure answer-perplexity ($\pplanswer$) on high-quality selected answers.
Here is an example QA for \ticstack{}:

\begin{figure}[!ht]
\begin{lstlisting}[language=JSON, frame=single]
{
    "query":
        "Question Title: If squaring a number means multiplying that number"
        " with itself then shouldn't taking square root of a number mean "
        " to divide a number by itself?"
        "\n\n"
        "Question Body: If squaring a number means multiplying that number"
        " with itself then shouldn't taking square root of a number mean"
        " to divide a number by itself?"
        "\n\n"
        "For example the square of $2$  is $2^2=2 \\cdot 2=4 $ ."
        "\n\n"
        "But square root of $2$ is not $\\frac{2}{2}=1$ .",
    "answer": 
        "`taking square root` means _reversing_ the effect of `squaring`."
        " Dividing a number by itself does not do that"
        " (but rather always returns 1 as you noted)."
        "\n\n"
        "Compare your question to: if doubling a number means adding it to "
        "itself, shouldn't halving a number mean subtracting it from itself?"
        "Answer: obviously not."
}
\end{lstlisting}
\caption{Example of \ticstack{} question and answers for Math.}
\end{figure}

\subsubsection{Analysis of StackExchange Data}

\begin{figure}[h]
\centering
\includegraphics[width=\textwidth]{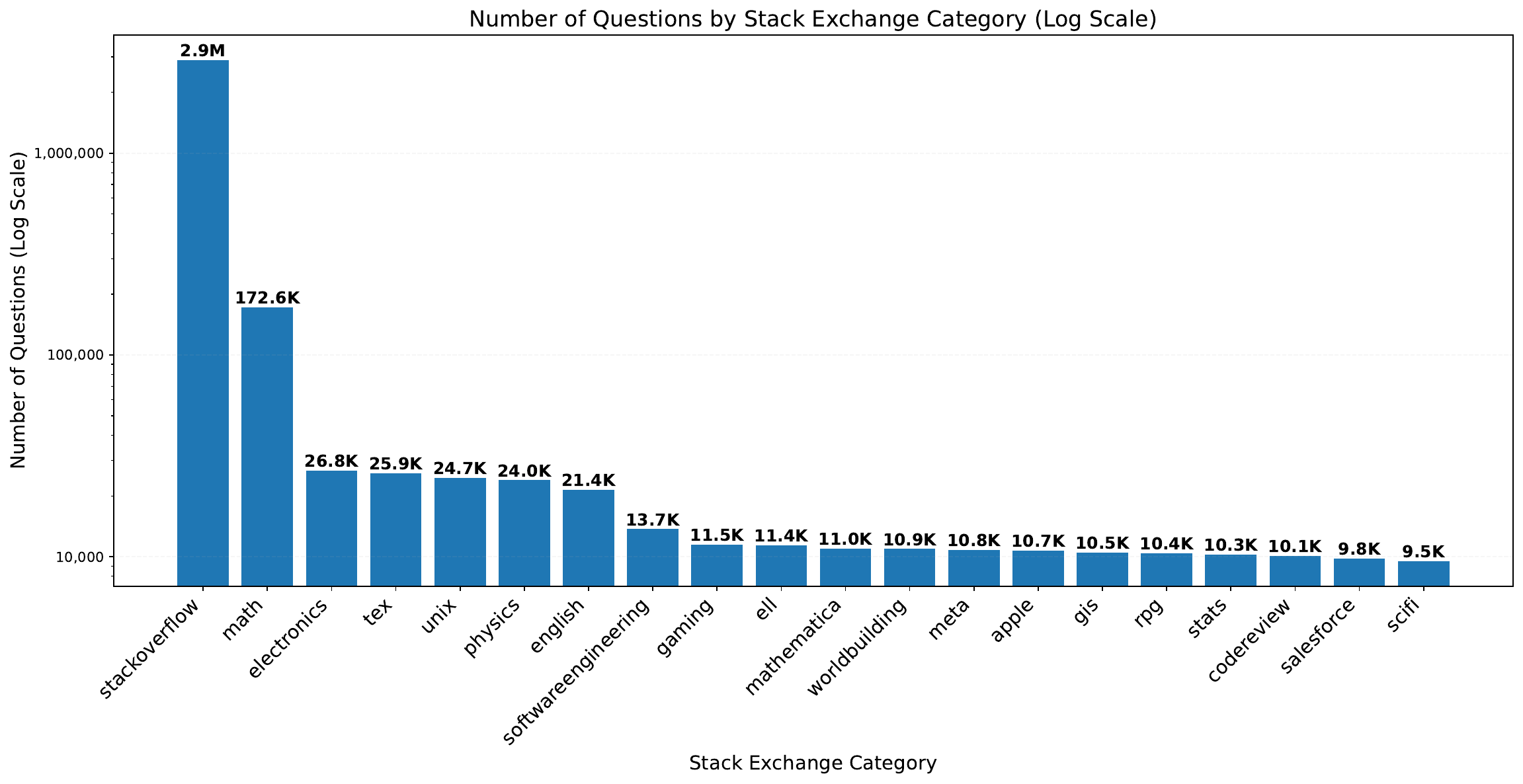}
\caption{Number of questions by StackExchange category (log scale).}
\label{fig:category_distribution}
\end{figure}

\begin{figure*}[h]
\centering
\begin{subfigure}[b]{0.32\textwidth}
    \includegraphics[width=\textwidth]{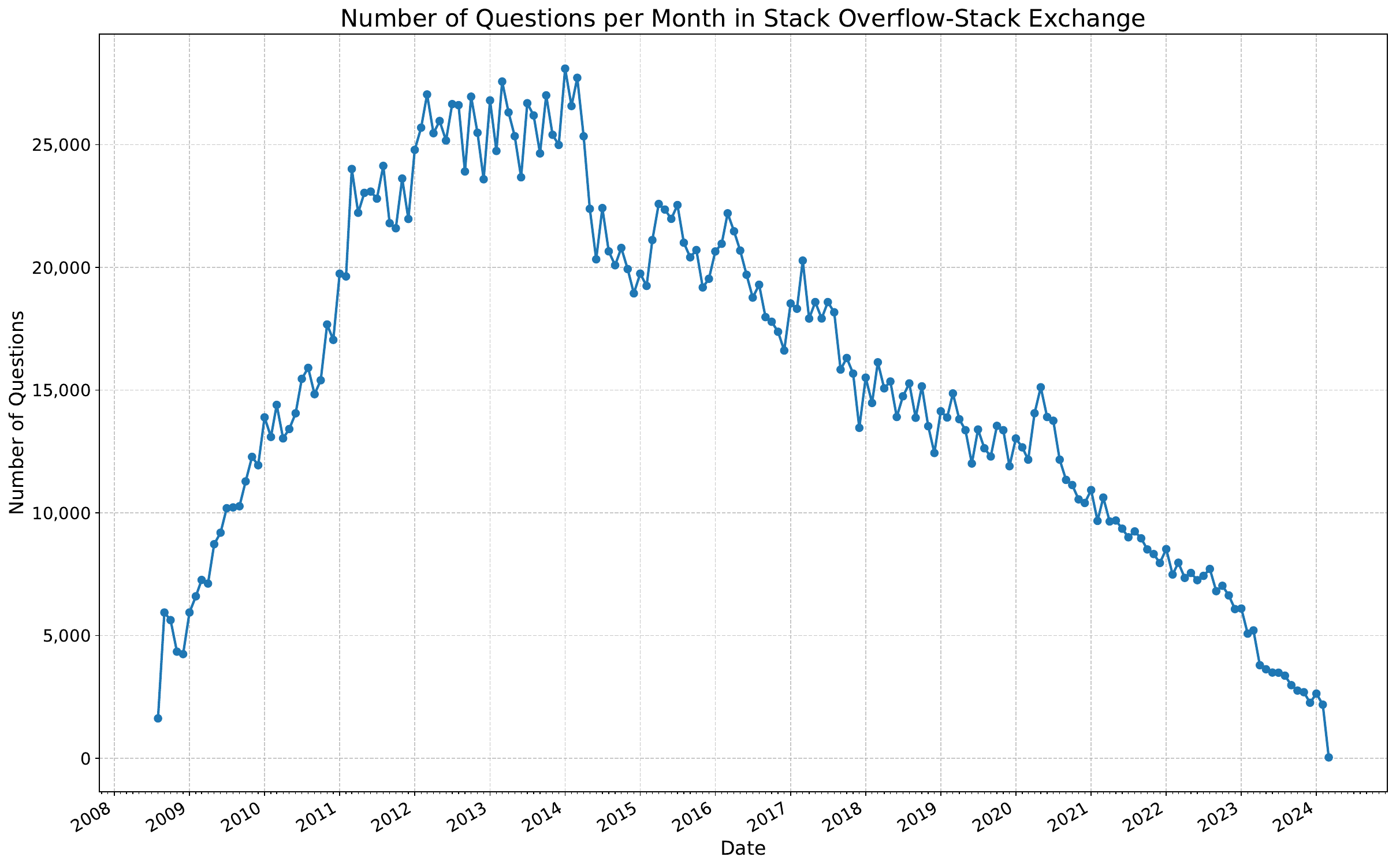}
    \caption{StackOverflow}
\end{subfigure}
\hfill
\begin{subfigure}[b]{0.32\textwidth}
    \includegraphics[width=\textwidth]{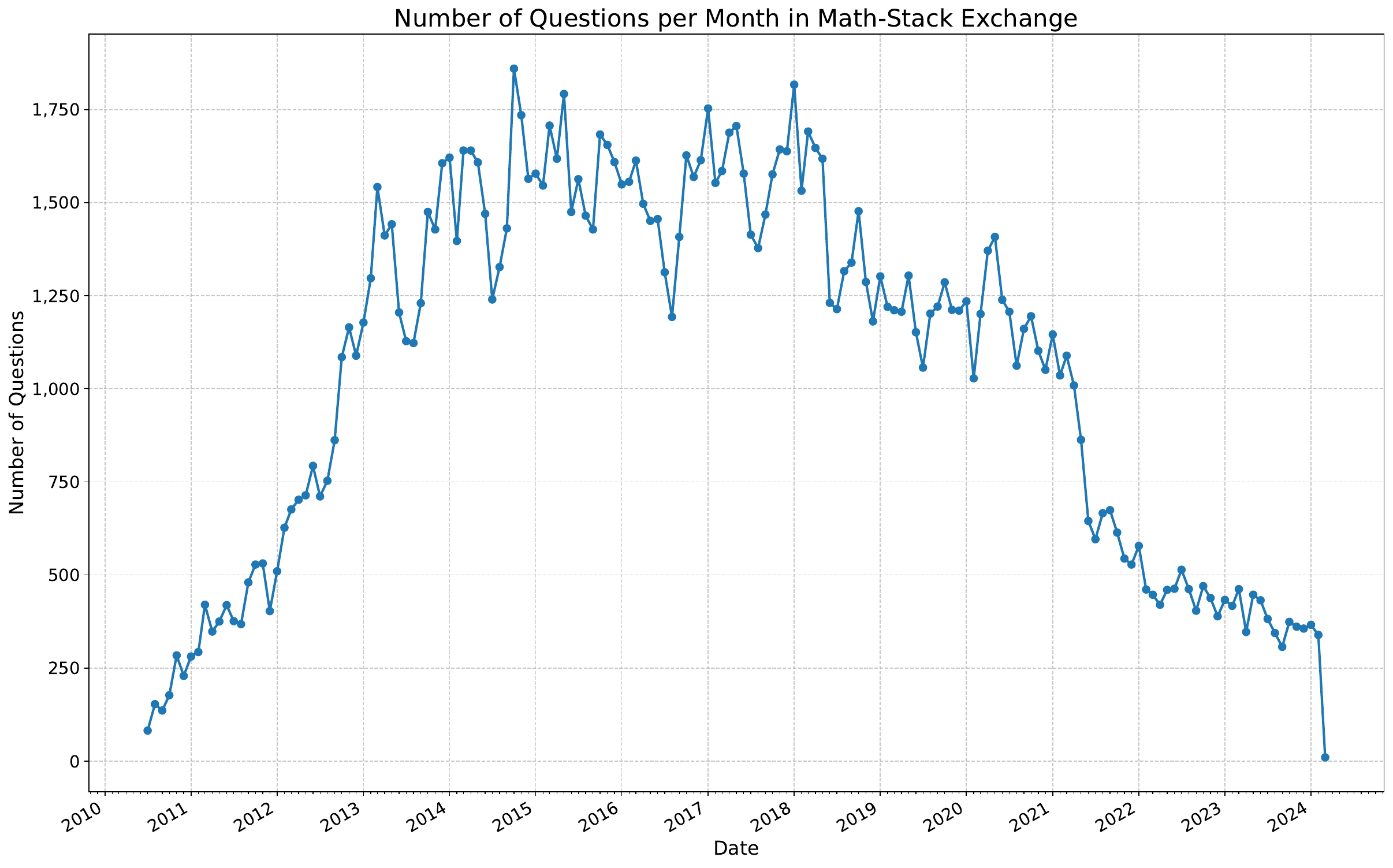}
    \caption{Mathematics}
\end{subfigure}
\hfill
\begin{subfigure}[b]{0.32\textwidth}
    \includegraphics[width=\textwidth]{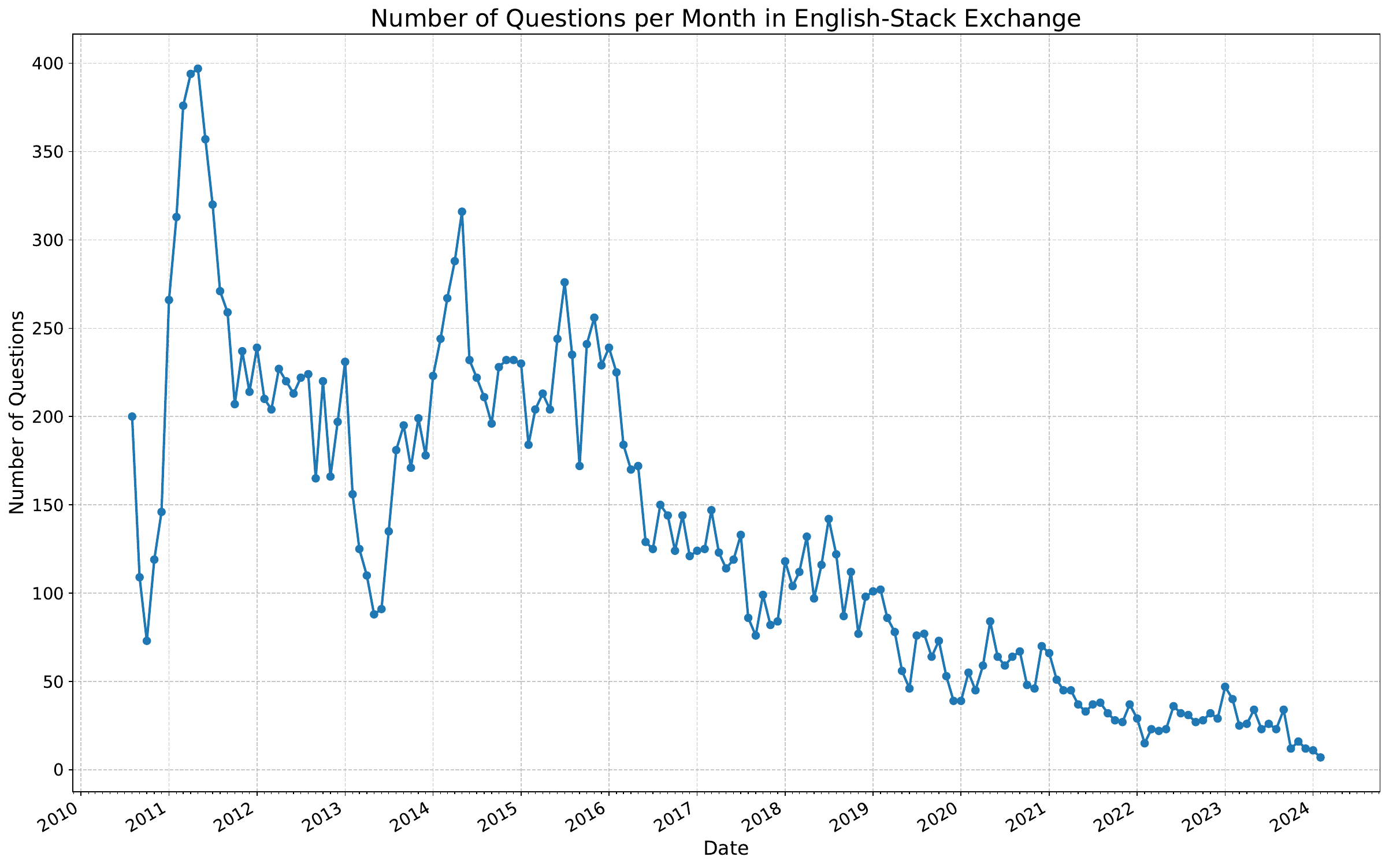}
    \caption{English Language \& Usage}
\end{subfigure}
\caption{Number of questions per month in StackOverflow, Mathematics and English Language \& 
    Usage.}
\label{fig:questions_per_month}

\end{figure*}

\begin{figure}[h]
\begin{subfigure}[b]{0.32\textwidth}
    \includegraphics[width=\textwidth]{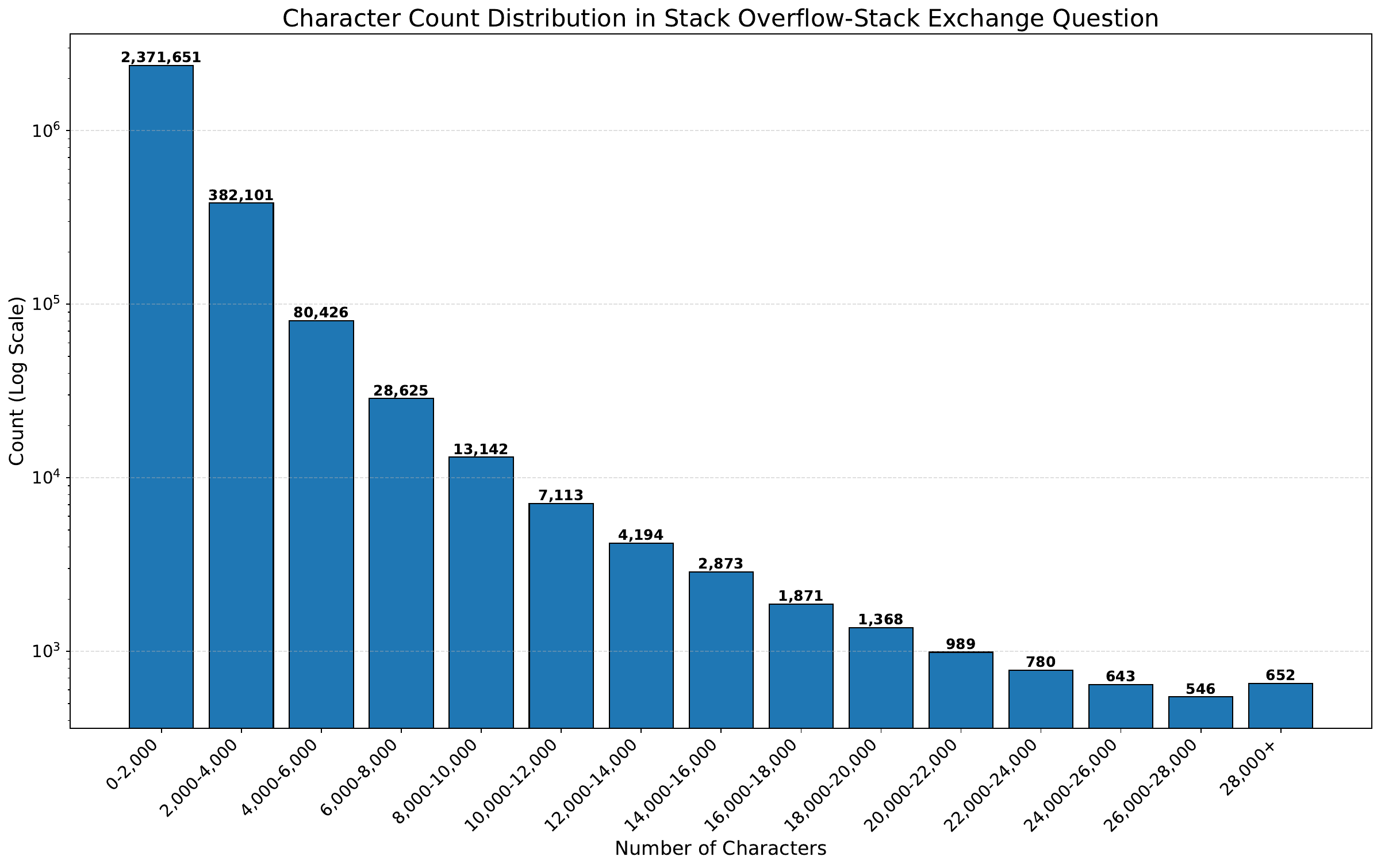}
    \caption{StackOverflow}
\end{subfigure}
\hfill
\centering
\begin{subfigure}[b]{0.32\textwidth}
    \includegraphics[width=\textwidth]{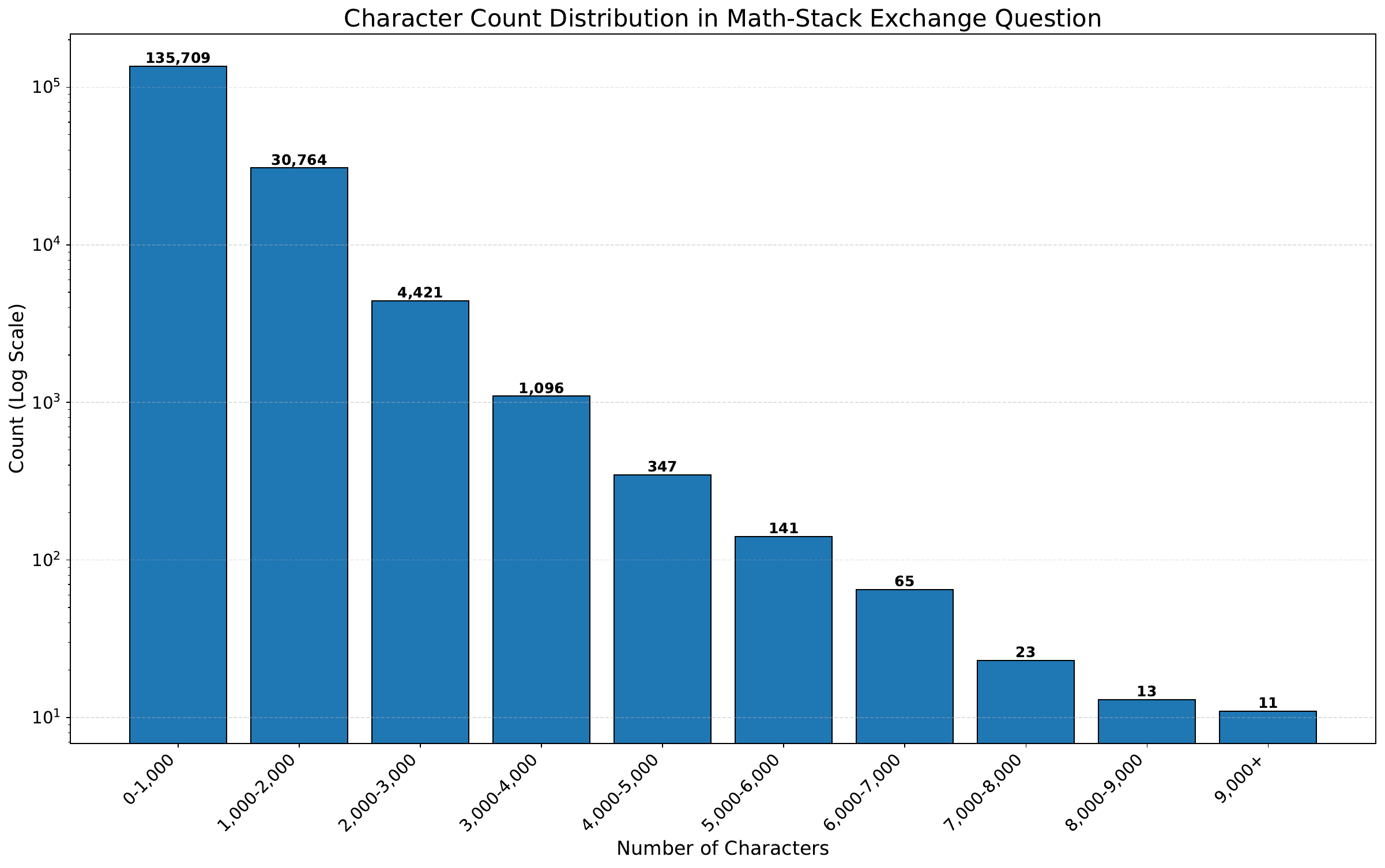}
    \caption{Mathematics}
\end{subfigure}
\hfill
\begin{subfigure}[b]{0.32\textwidth}
    \includegraphics[width=\textwidth]{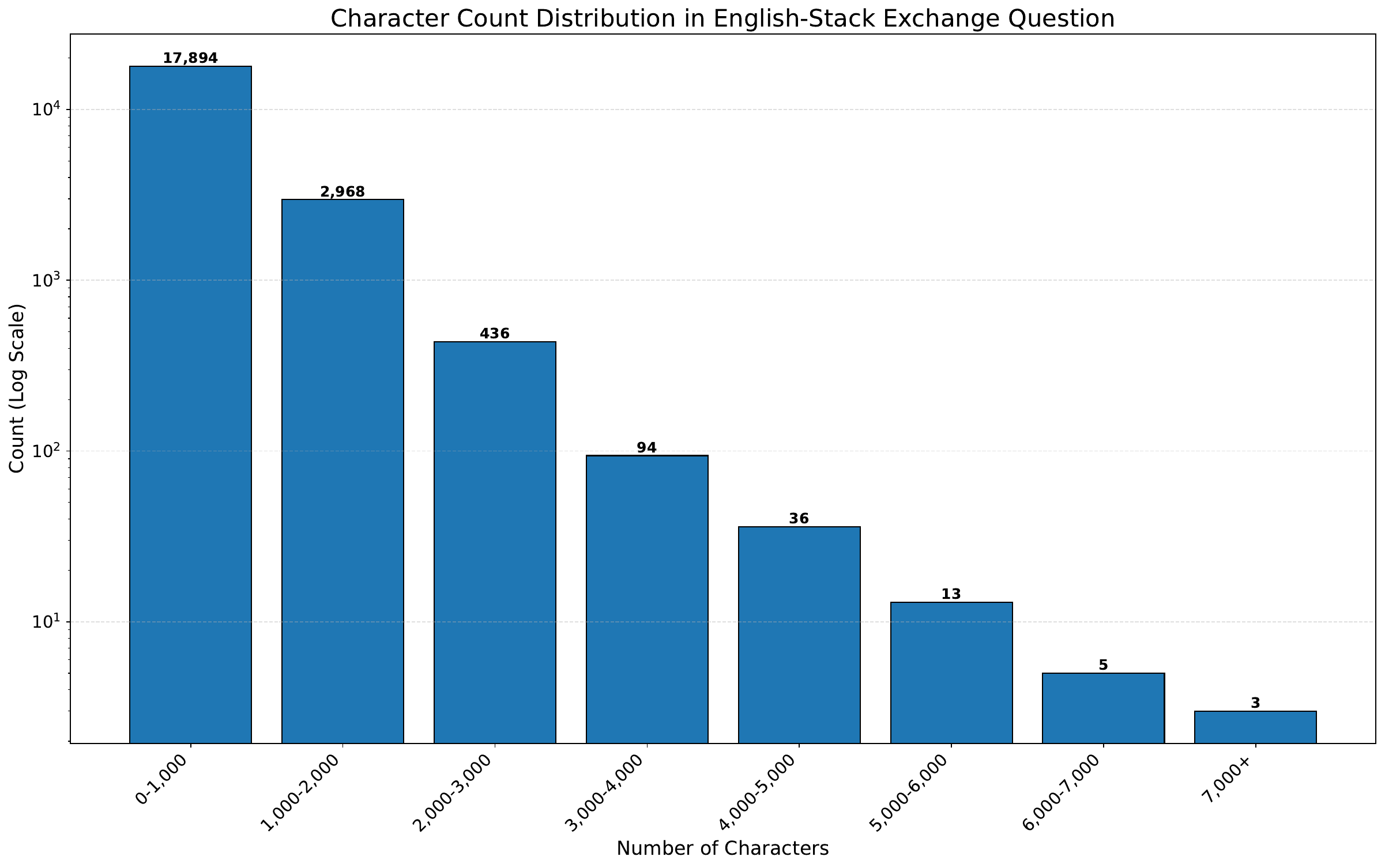}
    \caption{English Language \& Usage}
\end{subfigure}
\caption{Character Count Distribution in StackOverflow, Mathematics and English Language \& Usage Questions.}
\label{fig:question_length}

\end{figure}

\begin{figure}[h]
\centering
\begin{subfigure}[b]{0.32\textwidth}
    \includegraphics[width=\textwidth]{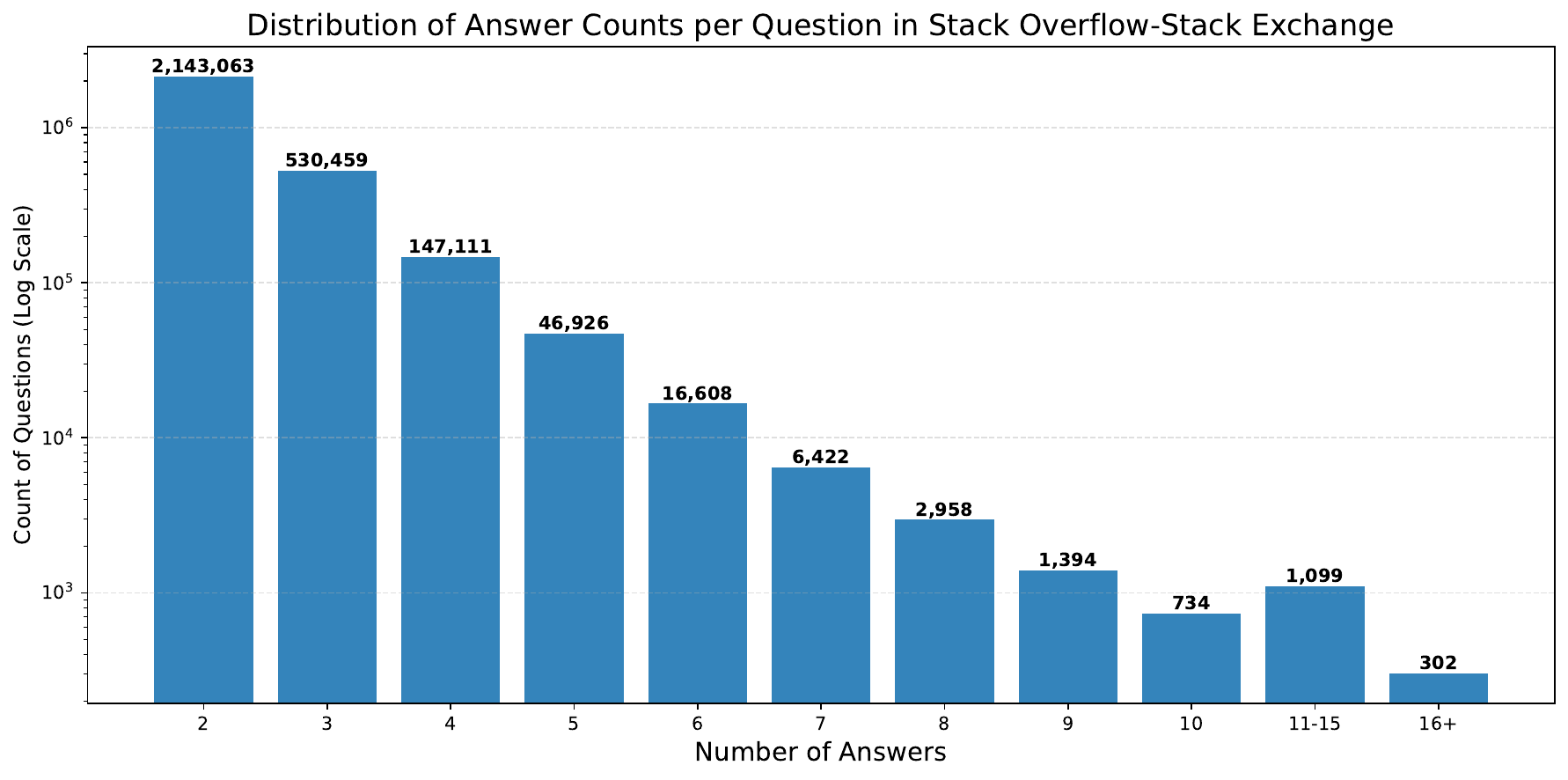}
    \caption{StackOverflow}
\end{subfigure}
\hfill
\begin{subfigure}[b]{0.32\textwidth}
    \includegraphics[width=\textwidth]{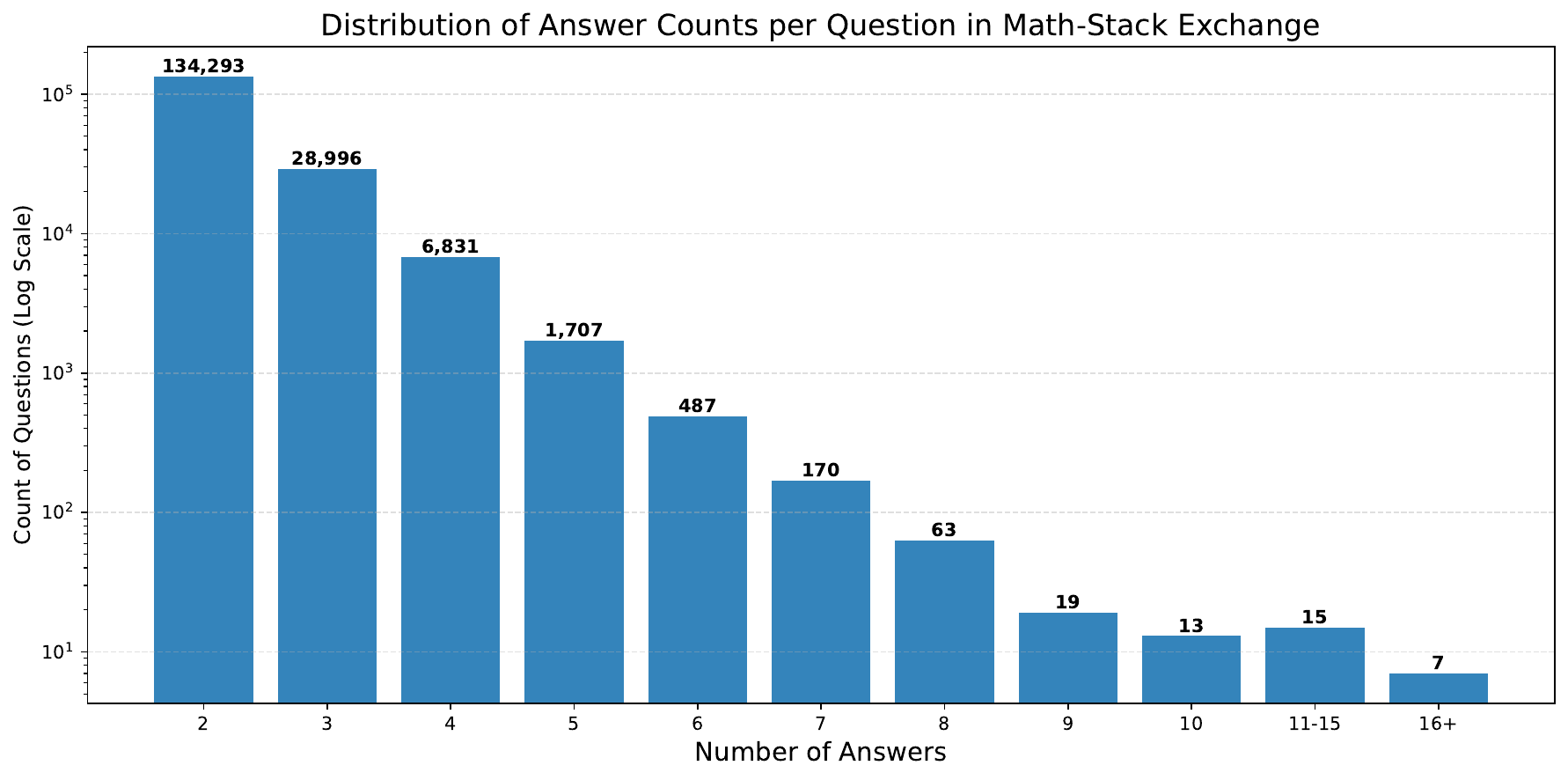}
    \caption{Mathematics}
\end{subfigure}
\hfill
\begin{subfigure}[b]{0.32\textwidth}
    \includegraphics[width=\textwidth]{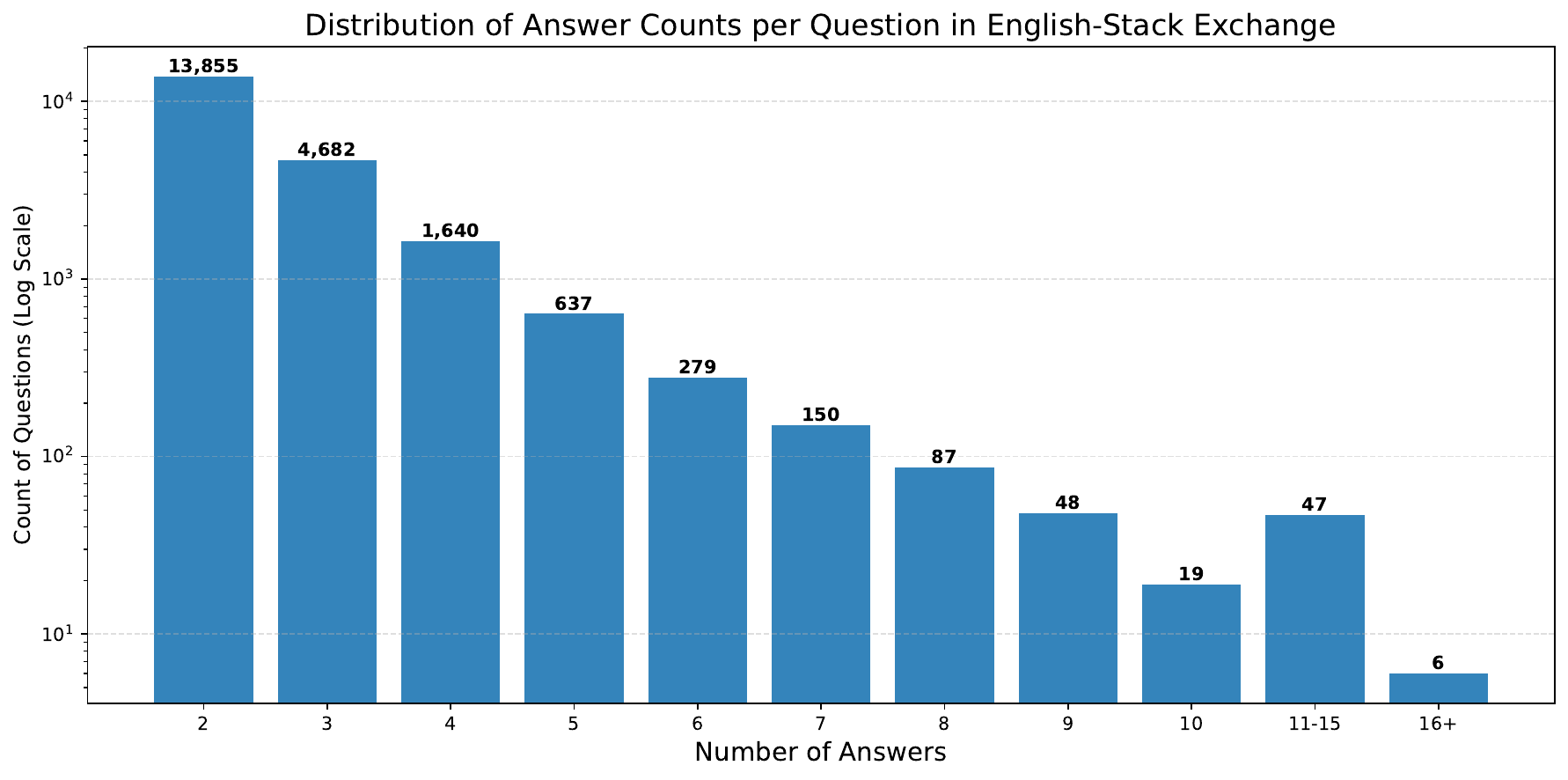}
    \caption{English Language \& Usage}
\end{subfigure}
\caption{Distribution of Answer Counts per Question in Mathematics and English Language \& Usage.}
\label{fig:answer_distribution}

\end{figure}

This section presents additional analysis of question-answer patterns across the top 20 
categories of StackExchange, with a focus on StackOverflow, Mathematics, and 
English Language \& Usage. \Cref{fig:category_distribution} shows the distribution of questions across the 
top 20 StackExchange categories. Then for our three chosen categories, \Cref{fig:questions_per_month} shows the number of questions asked per month, \Cref{fig:answer_distribution}
presents the distribution of answer counts per question, and \Cref{fig:question_length} illustrates the distribution of question lengths.

\clearpage

\subsection{\ticdocs{}}
\label{sec:tic_codedocs_sup}
Our \ticdocs{} dataset consists of official documentation extracted systematically from the Git repositories of NumPy~\citep{numpy} and PyTorch~\citep{Pytorch}. 

As shown in Table~\ref{tab:sup-ticcode-stats}, we include 16 major releases of NumPy (from \texttt{v1.13.0} in 2017 to \texttt{v2.1.0} in 2024) and 11 major releases of PyTorch (from \texttt{v1.8.0} in 2021 to \texttt{v2.4.0} in 2024). For each release, we revert the repository to the corresponding commit, build the library in a dedicated virtual environment, and generate HTML documentation via Sphinx. To extract clean plaintext content from these HTML documents, we use \texttt{readability.js}\footnote{\href{https://github.com/mozilla/readability}{https://github.com/mozilla/readability}}, which effectively preserves only the main textual content.

The processed documentation is then formatted into a structured JSONL file, where each entry consists of the document's title and textual content. This approach ensures consistency, reproducibility, and accurate representation of each library’s documentation, enabling reliable evaluation of model performance (measured by perplexity, $\ppltoken{}$) across library versions and their temporal evolution.

\begin{table}[h]
  \centering
  \caption{\ticdocs{}: Details of versions, release times, and number of extracted documents.}
  \label{tab:sup-ticcode-stats}
  \begin{minipage}[t]{0.4\textwidth}
    \centering
    \resizebox{\textwidth}{!}{%
      \begin{tabular}{@{}ccc@{}}
        \toprule
        \textbf{Numpy Version} & \textbf{Year/Month} & \textbf{\#Documents} \\ \midrule
        1.13.0                 & 2017/06             & 2072                 \\
        1.14.0                 & 2018/01             & 2097                 \\
        1.15.0                 & 2018/07             & 2111                 \\
        1.16.0                 & 2019/01             & 2112                 \\
        1.17.0                 & 2019/07             & 2201                 \\
        1.18.0                 & 2019/12             & 2450                 \\
        1.19.0                 & 2020/06             & 2462                 \\
        1.20.0                 & 2021/01             & 2419                 \\
        1.21.0                 & 2021/06             & 2502                 \\
        1.22.0                 & 2021/12             & 2455                 \\
        1.23.0                 & 2022/06             & 2470                 \\
        1.24.0                 & 2022/12             & 2538                 \\
        1.25.0                 & 2023/06             & 2550                 \\
        1.26.0                 & 2023/09             & 2550                 \\
        2.0.0                  & 2024/06             & 2510                 \\
        2.1.0                  & 2024/08             & 2528                 \\ \bottomrule
      \end{tabular}%
    }
  \end{minipage}%
  \hfill
  \begin{minipage}[t]{0.4\textwidth}
    \centering
    \resizebox{\textwidth}{!}{%
      \begin{tabular}{@{}ccc@{}}
        \toprule
        \textbf{PyTorch Version} & \textbf{Year/Month} & \textbf{\#Documents} \\ \midrule
        1.8.0                    & 2021/03             & 1373                 \\
        1.9.0                    & 2021/06             & 2998                 \\
        1.10.0                   & 2021/10             & 3127                 \\
        1.11.0                   & 2022/03             & 3533                 \\
        1.12.0                   & 2022/06             & 3591                 \\
        1.13.0                   & 2022/10             & 3665                 \\
        2.0.0                    & 2023/03             & 3944                 \\
        2.1.0                    & 2023/10             & 4050                 \\
        2.2.0                    & 2024/01             & 4096                 \\
        2.3.0                    & 2024/04             & 4295                 \\
        2.4.0                    & 2024/07             & 4385                 \\ \bottomrule
      \end{tabular}%
    }
  \end{minipage}
\end{table}

\subsection{Evaluation metrics}\label{app:eval_summary_metrics}

Here, we more comprehensively discuss details about our evaluation metrics. To recap, each run produces a $T_t \times T_e$ matrix of evaluations 
$E$ where $T_t,T_e$ are the number of training/evaluation timesteps, 
$E_{i,j}$ is the performance of the model trained after seeing data up to 
month $i$
and evaluated on the month $j$.

To reduce the computation costs of running evaluations, while we train and produce checkpoints on all 114 CC dumps in \ticcc{}, we evaluate on only 12 checkpoints which are roughly annually spaced. These include the first month from each year between 2014-2024 as well as the very last timestamp July-2024.

To 
control for inherent difficulty differences across evaluation months, we measure the \textit{regret} $R_{i,j} = E_{i,j} - E^*_{j}$ where 
$E^*_{j}$ is the performance of \textit{Oracle-2024-07}
on month $j$. We subtract $E^*_{j}$ instead of $E_{j,j}$ since if $E_{j,j}$ is bad it may lead to misleadingly good forward/backward metrics. 
As stated in \cref{sec:experiments}, we consider the following summary metrics, first defined assuming $T_t=T_e=T$ (as in the case of \ticcc{} where we have perfectly aligned training and evaluation data):

\begin{itemize}[leftmargin=*]
\vspace*{-2.5mm}
\itemsep0em
    \item In-distribution (ID) performance: averages along the matrix diagonal, i.e.,  $\sum_{i=1}^T = R_{i,i} / T$. 
    \item Backward transfer: averages the lower triangular of $R$, i.e., 
        $\sum_{i=1}^T\sum_{j<i}\frac{R_{i,j}}{T(T-1)/2}$.
    \item Forward transfer: averages the upper triangular of $R$ analogously to backward transfer.
\end{itemize}
\vspace*{-2.5mm}

For downstream evaluations where evaluation periods do not align with model checkpoint dates, we use an exclusive assignment strategy to define in-domain (ID) performance. For each checkpoint $i$, we identify its \textit{nearest preceding evaluation} -- the evaluation timestep that comes before or concurrent with checkpoint $i$ and is closest in time. However, since multiple checkpoints may share the same nearest preceding evaluation, we ensure each evaluation timestep is claimed as ``concurrent'' by only the temporally closest training timestep.

Formally, let $t_i$ and $e_j$ denote the timestamps of checkpoint $i$ and evaluation period $j$ respectively. Checkpoint $i$ contributes $R_{i,j}$ to the ID average only if evaluation period $j$ is both (1) the nearest preceding evaluation to checkpoint $i$, and (2) checkpoint $i$ is the closest checkpoint to evaluation period $j$ among all checkpoints that have $j$ as their nearest preceding evaluation. All other pairs $R_{i,j}$ contribute to forward transfer when $t_i < e_j$, and backward transfer when $t_i > e_j$.

%% file: sec/sup_hparams.tex
\clearpage
\section{Training and hyperparameter details}\label{app:hyperparamters}

\begin{figure}[h!]
    \centering
     \includegraphics[width=0.5\linewidth]{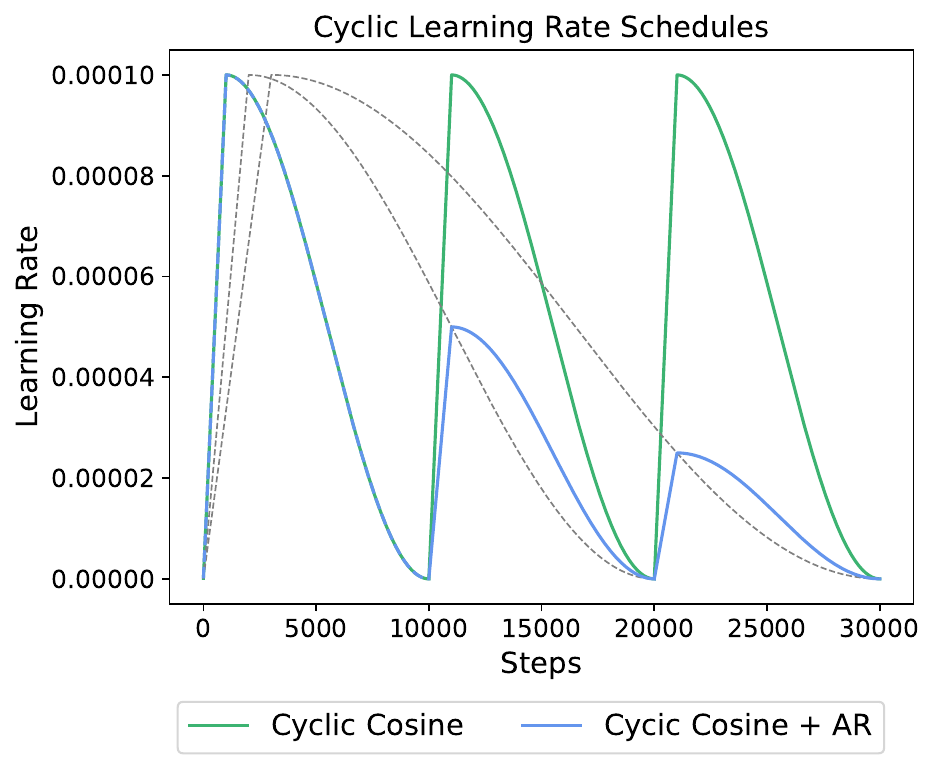}
    \caption{\textbf{Cyclic cosine learning rate schedules} compared with a single
    cosine learning rate schedule (gray) restart the schedule to either the
    initial maximum (green) or gradually decaying values (blue).}
    \label{fig:cyclic_lr}
    \vspace{-3mm}
\end{figure}

\subsection{General details}
 
 We follow the architectures and configurations used by DataComp-LM \citep{li2024datacomp} for their 3B model scale (unless further specified). For our Oracle and initialization trained on May-2013, we exactly follow their hyperparameters given that these were also standard pre-training runs from scratch. Most notably, this includes using a cosine schedule with maximum learning rate of 3e-3, a final learning rate of 3e-5, and a weight decay of 0.033. Each experiment was run using the OpenLM library \citep{openlm} on 64 H100 GPUs with a global batch size of 256 and sequence length of 2048 tokens (using the GPT-NeoX \citep{gpt-neox-20b} tokenizer). The continual phases of our training runs require 1.5K (for 220B scale) and 4.5K (for 440B scale) total H100 hours, with the exception of LwF/EWC, for which our implementations incurred about a 30\% additional cost.

\subsection{Hyperparameters for continual methods}
For our various continual methods, we do perform additional hyperparameter tuning using the first 10 TiC-CC training sets and held-out validation sets. Following  \citet{cha2024hyperparameterscontinuallearningreality}, we limit the tuning to an earlier limited set of training rounds given that it would be impossible for a practitioner to be able to tune based upon data they have not seen far in the future. We discuss the tuning and hyperparameter choices for specific methods in more detail below. All results are for the 220B token scale.
    
\textbf{Cyclic Cosine.} We mainly tuned the maximum learning rate in each cycle, trying values between 1e-3 and 3e-5, as shown in \cref{tab:lr_tuning}. On our tuning set, the best setting across the board was 1e-4. When carrying out these tuning runs to completion on all 113 timesteps, we do observe an important difference in behavior. While 1e-4 continues to offer the best ID performance and strictly dominates all higher settings, lowering it further can be used to trade-off Backward and ID performance. The smallest fixed max learning rate, 3e-5 results in a similar yet overall worse performance profile to using an an AR meta-schedule. This makes sense given the AR schedule roughly can be considered to decrease the maximum learning rate at a $1/t$ rate; since our setup involves over 100 months, AR schedules set the maximum learning rate very close to the minimum of 3e-5 in most rounds. Overall, we find that learning rates do need to be lowered by at least 30x compared to the the May-2013 initialization (which used 3e-3). This is in contrast to \citet{ibrahim2024simple, gupta2023continual} which both suggest re-warming up to a similar learning rate as the initial pre-training or \citet{parmar2024reusedontretrain} who start from the minimum learning rate of the pre-trained model. We suspect this is due to the difference in setup (i.e., these works use only 2 or 3 training rounds of comparable sizes and face distribution shifts related to data quality and language rather than temporal evolution). 

\begin{table}[h!]
    \scriptsize
    \centering
    \caption{\textbf{Tuning LR for Cyclic Cosine}}
    \label{tab:lr_tuning}
    \vspace*{-2mm}
    \begin{tabular}{c|ccc|ccc}
        \toprule[1.2pt]
        \multirow{2}{*}{\textbf{Max LR}} & \multicolumn{3}{c|}{\textbf{\ticcc{} (Tuning Months) }} & \multicolumn{3}{c}{ \textbf{\ticcc{} (All Months)}} \\
        & Backward & ID & Forward & Backward & ID & Forward \\
        \midrule
        \multirow{1}{*}{1e-3} &  0.103 & 0.086 & 0.118 & 0.197 & 0.083 & 0.209\\
        \multirow{1}{*}{3e-4} & 0.019 & 0.016 & 0.051 & 0.125 & 0.041 & 0.178 \\
        \multirow{1}{*}{1e-4} & \textbf{0.002} & \textbf{0.005} & \textbf{0.039} & 0.072 & \textbf{0.027} & \textbf{0.161} \\ 
        \multirow{1}{*}{5e-5} & \textbf{0.002} & 0.006 & \textbf{0.039} & 0.062 & 0.034 & 0.163 \\ 
        \multirow{1}{*}{3e-5} &  0.004 & 0.009 & 0.040 &  0.060 & 0.042 & 0.165 \\ \hline 
        \multirow{1}{*}{AR Schedule} & \textbf{0.002} & 0.008 & 0.043 &  \textbf{0.058} & 0.040 & 0.166\\ 
        \bottomrule[1.2pt]
 \end{tabular}
\end{table}

\textbf{Rsqrt.} We tuned both the maximum learning rate within the same range as Cyclic Cosine as well as the cooldown length, choosing between 50 and 400. Our final run continued to use 1e-4 for the maximum learning rate and 400 for the cooldown, though there did not appear to be much difference when compared to smaller values such as 200 or 100 on the tuning months.

\textbf{Schedule-Free.} We continued to use warmup but given that Schedule-Free makes more drastic changes to optimization (i.e., using a different optimizer versus simply a different learning rate schedule), we re-tuned both the learning rate and weight decay. Interestingly, 1e-4 as the maximum learning rate continued to work best for us, though we found it helped slightly to drop the weight decay from 0.033 to 0.01. 

\begin{table}[h!]
    \scriptsize
    \centering
    \caption{\textbf{Tuning for Schedule-Free}}
    \label{tab:schedfree_tuning}
    \vspace*{-2mm}
    \begin{tabular}{cc|ccc}
        \toprule[1.2pt]
        \multirow{2}{*}{\textbf{Max LR}} & \multirow{2}{*}{\textbf{WD}}  & \multicolumn{3}{c}{\textbf{\ticcc{} (Tuning Months) }}  \\
        & & Backward & ID & Forward \\
        \midrule
        \multirow{1}{*}{1e-3} & 0.033 &  0.1025 & 0.0856 & 0.1178 \\
        \multirow{1}{*}{5e-4} & 0.033 & 0.0448 & 0.0373 & 0.0713 \\ 
        \multirow{1}{*}{3e-4} & 0.033 &  0.0206 & 0.0183 & 0.0532 \\ 
        \multirow{1}{*}{5e-5} & 0.033 & 0.0053 & 0.0105 & 0.0406
        \\ \hline
        \multirow{1}{*}{1e-4} & 0.067 &  0.0049 & 0.0080 & 0.0406 \\
        \multirow{1}{*}{1e-4} & 0.033 &  0.0044 & 0.0077 & 0.0404 \\
        \multirow{1}{*}{1e-4} & 0.010 & \textbf{0.0042} & \textbf{0.0075} & \textbf{0.0403} \\
        \multirow{1}{*}{1e-4} & 0.005 & \textbf{0.0042} & \textbf{0.0075} & \textbf{0.0403} \\ 
        \multirow{1}{*}{1e-4} & 0.0001 & 0.0044 & 0.0077 & 0.0404 \\
        \bottomrule[1.2pt]
 \end{tabular}
\end{table}

\textbf{LwF.} Following the original paper \citep{li2018lwf}, we used a temperature parameter of $T=2$. We mainly tuned the regularization weight $\lambda$ trying values between 0.1 and 10.0 and settling upon 0.3. However, overall we found using LwF either resulted in little difference (when using a small $\lambda$) or started to decrease all metrics (when using a larger $\lambda$).

\textbf{EWC.} We fixed the number of iterations used to estimate the Fisher matrix to 100 and similar to LwF, we focused on tuning the weight given to the EWC regularization term. Overall, we found that fairly high values were needed to overcome the small values in the approximate Fisher matrix (coming from small second order moment terms). We found that $\lambda=10^7$ performed best when tuning between $10^1$ and $10^9$, as shown in \cref{tab:ewc_tuning}. The only other setting we tried that is not strictly dominated by this choice was $\lambda=10^6$, which resulted in slightly better ID performance but significantly worse backward transfer.
\vspace{-0em}
\begin{table}[h]
    \scriptsize
    \centering
    \caption{\textbf{Tuning $\lambda$ for EWC}}
    \label{tab:ewc_tuning}
    \vspace*{-2mm}
    \begin{tabular}{c|ccc}
        \toprule[1.2pt]
        \multirow{2}{*}{\textbf{$\lambda$}} & \multicolumn{3}{c}{\textbf{\ticcc{} (Tuning Months) }}\\
        & Backward & ID & Forward  \\
        \midrule
        \multirow{1}{*}{$10^0$} & 0.0025 & 0.0050 & 0.0394 \\
        \multirow{1}{*}{$10^1$} & 0.0025 & 0.0050 & 0.0394 \\
        \multirow{1}{*}{$10^4$} & 0.0025 & 0.0050 & 0.0394  \\
        \multirow{1}{*}{$10^5$} & 0.0025 & 0.0049 & 0.0394 \\
        \multirow{1}{*}{$10^6$} & 0.0021 & \textbf{0.0047} & 0.0391 \\
        \multirow{1}{*}{$10^7$} & \textbf{0.0013} & 0.0050 & \textbf{0.0389} \\
        \multirow{1}{*}{$10^8$} &  0.0107 & 0.0178 & 0.0462 \\
        \multirow{1}{*}{$10^9$} & 0.0286 & 0.0400 & 0.0586 \\ \bottomrule[1.2pt]
 \end{tabular}
\end{table}

\clearpage

%% file: sec/sup_results.tex
\section{Extended Results}\label{app:extended_results}

\subsection{TiC-CommonCrawl (\ticcc{}) evaluations}

\begin{figure}[H]
    \centering
    \includegraphics[width=0.9\linewidth]{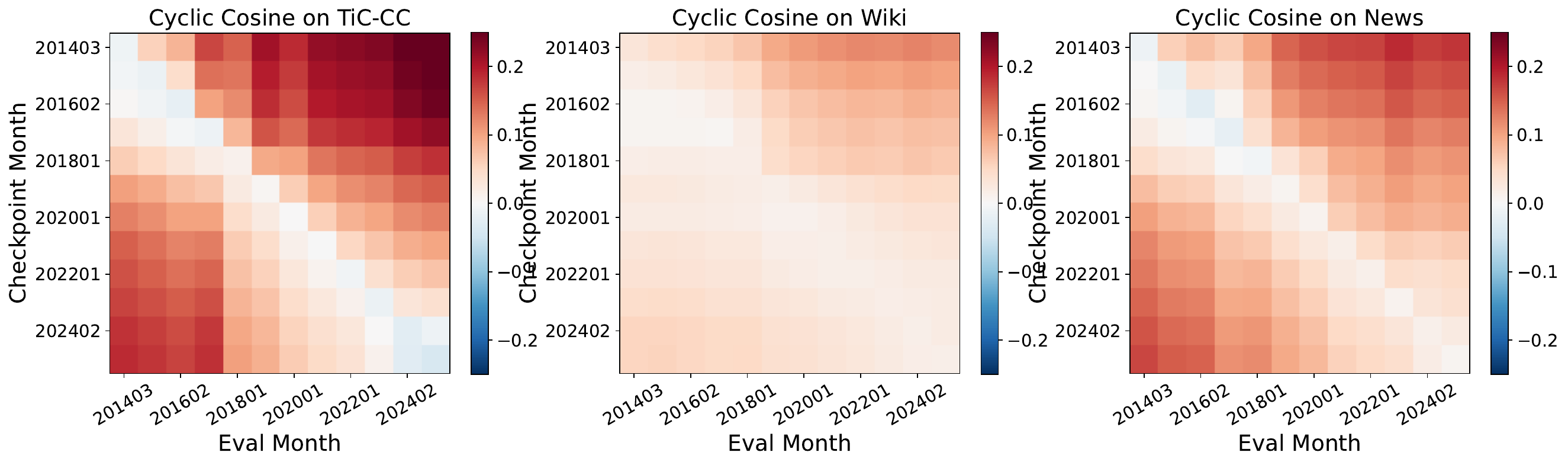}
    \includegraphics[width=0.9\linewidth]{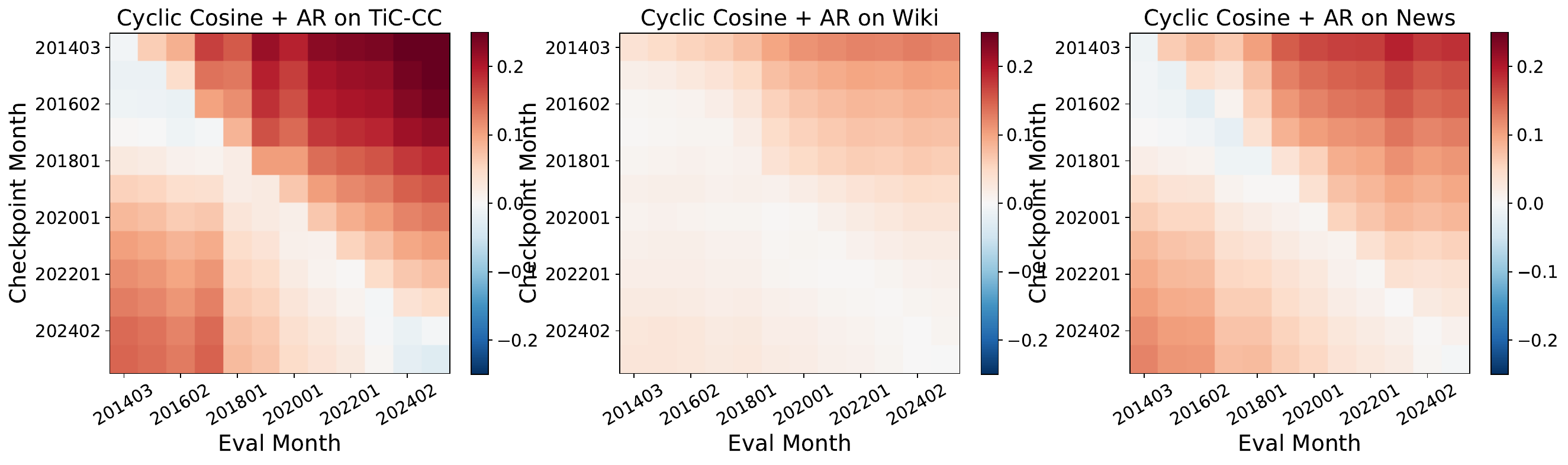}
    \includegraphics[width=0.9\linewidth]{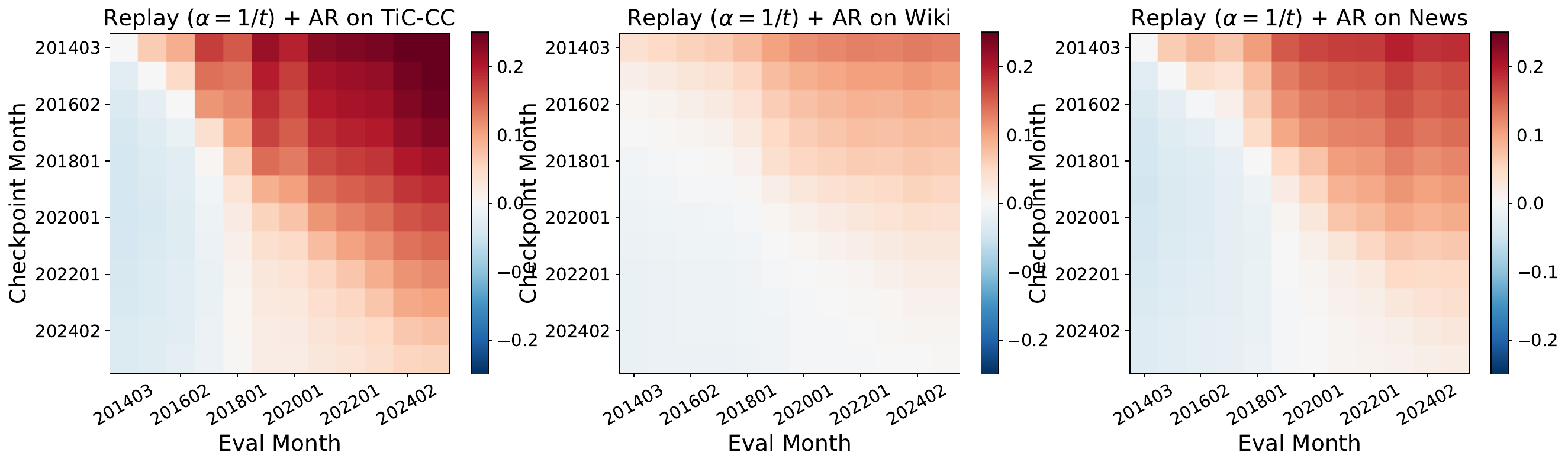}
    \includegraphics[width=0.9\linewidth]{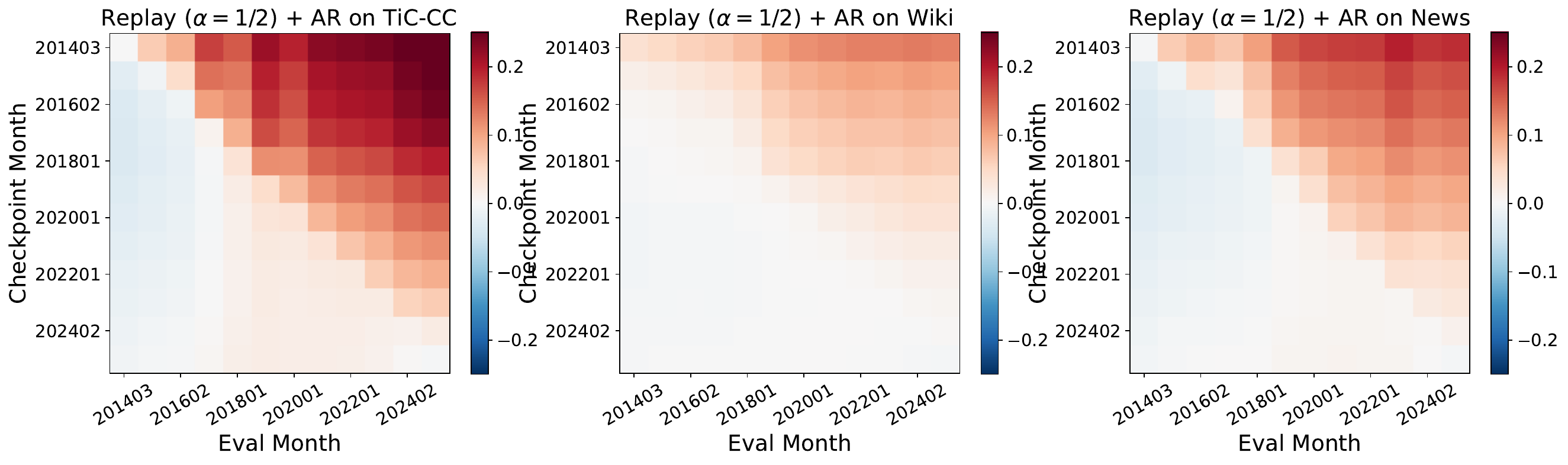}
    \includegraphics[width=0.9\linewidth]{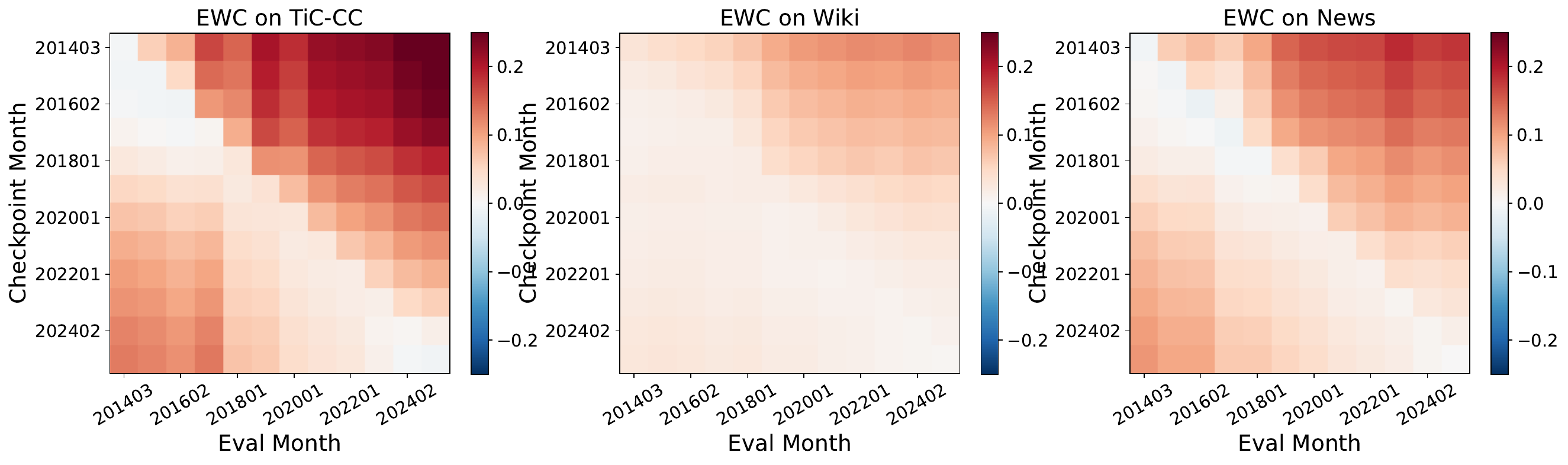}
    \caption{Evaluation matrix heatmaps for selected methods on our \ticcc{} evaluations (440B token scale). }
    \label{fig:cc_extended_heatmaps_440b}
\end{figure}

\begin{figure}[H]
    \centering
    \includegraphics[width=0.9\linewidth]{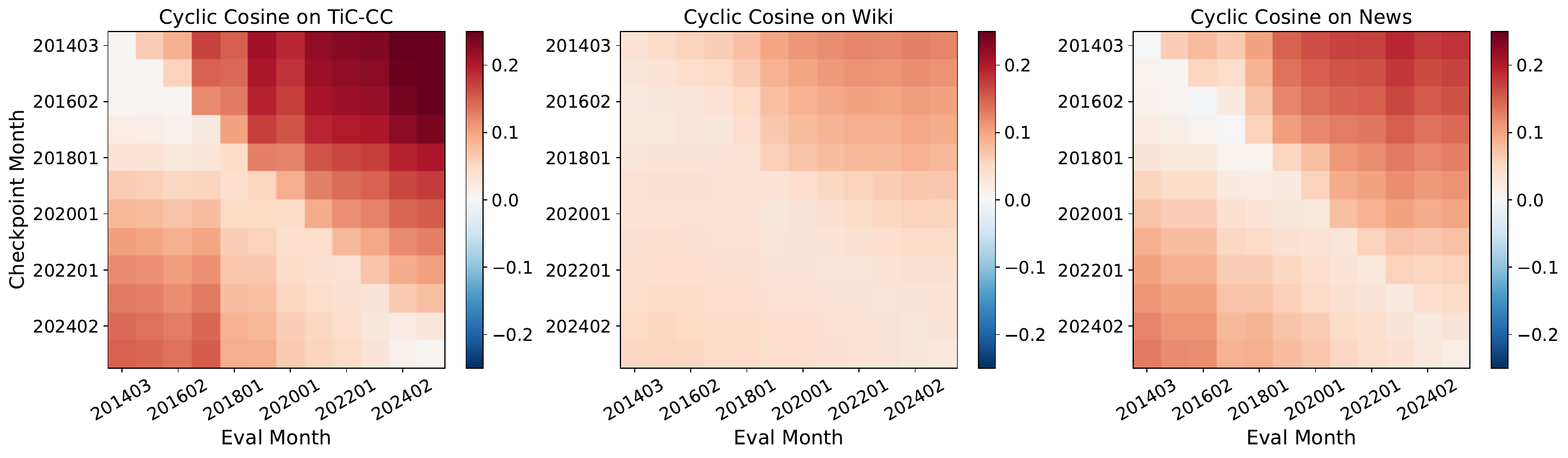}
    \includegraphics[width=0.9\linewidth]{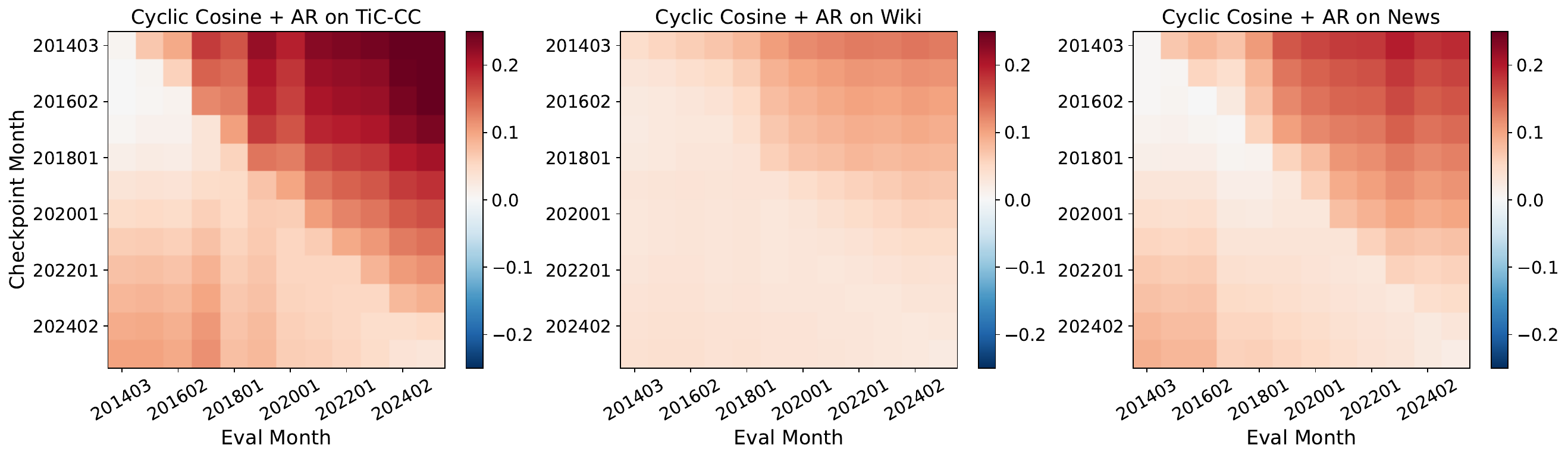}
    \includegraphics[width=0.9\linewidth]{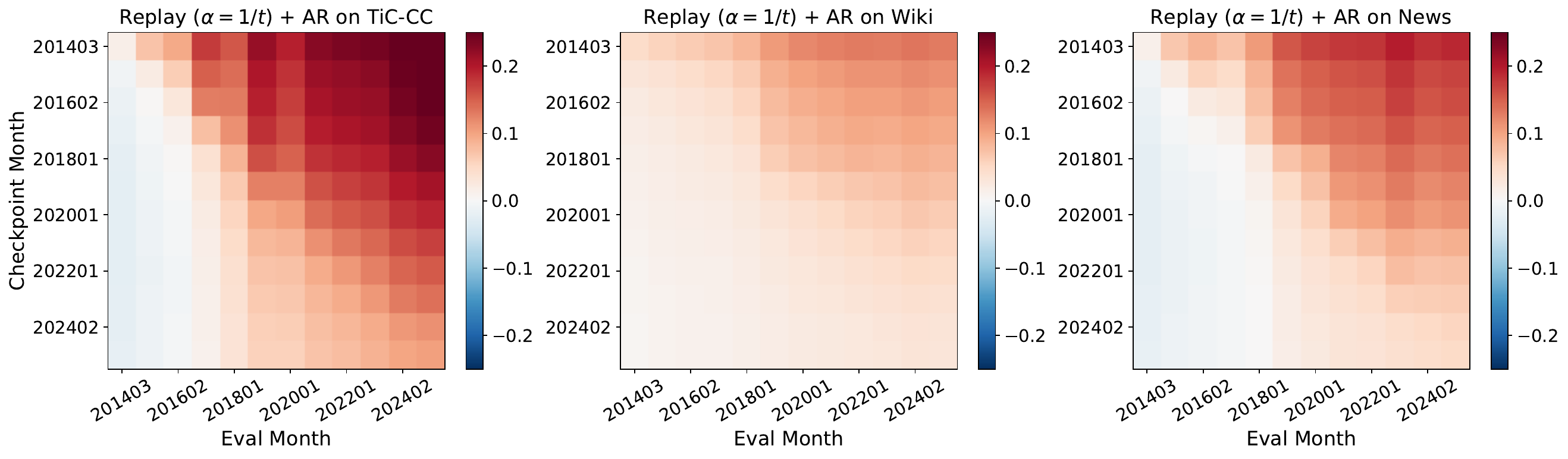}
    \includegraphics[width=0.9\linewidth]{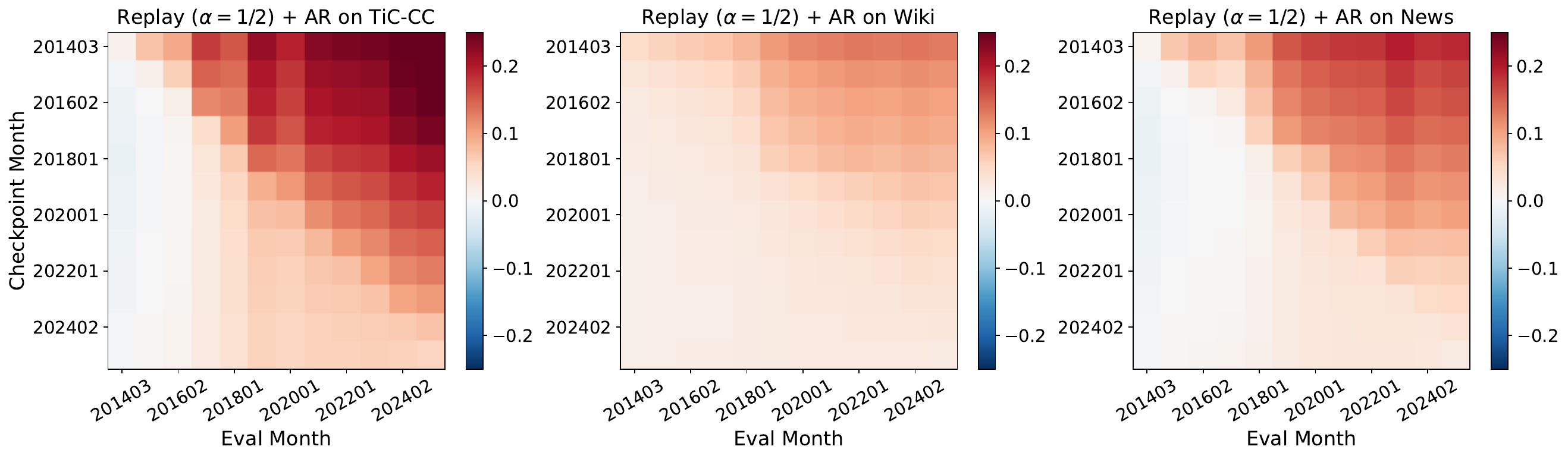}
    \includegraphics[width=0.9\linewidth]{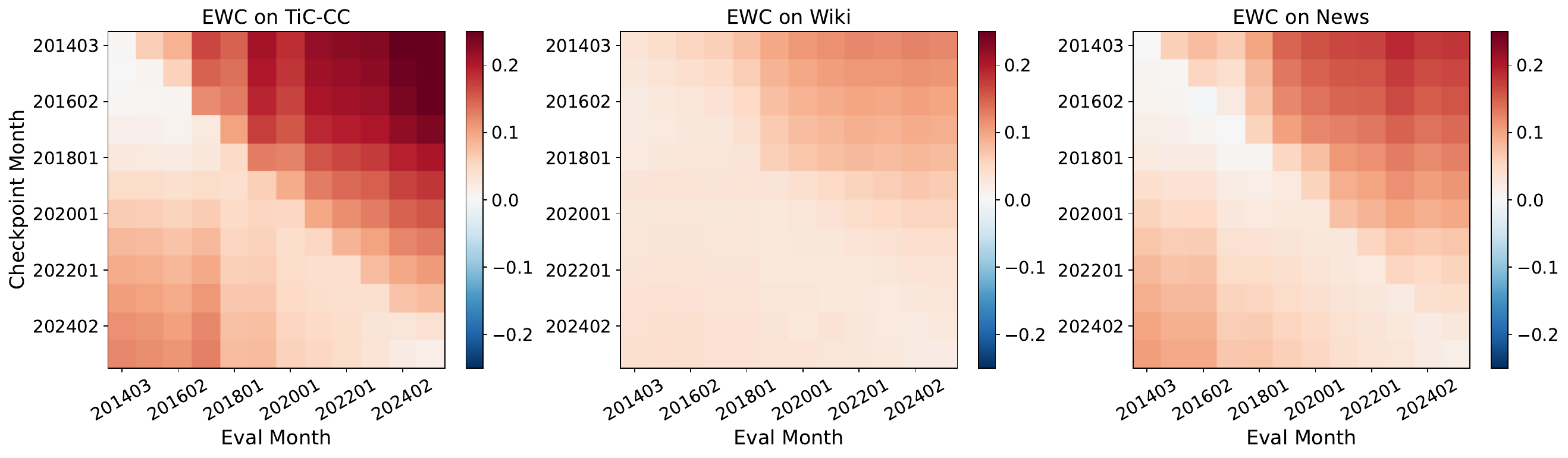}
    \caption{Evaluation matrix heatmaps for selected methods on our \ticcc{} evaluations (220B token scale). }
    \label{fig:cc_extended_heatmaps_220b}
\end{figure}

\clearpage

\subsection{TiC-Wikipedia (\ticwiki{})}

\begin{figure}[h!]
    \centering
    \includegraphics[width=0.85\linewidth]{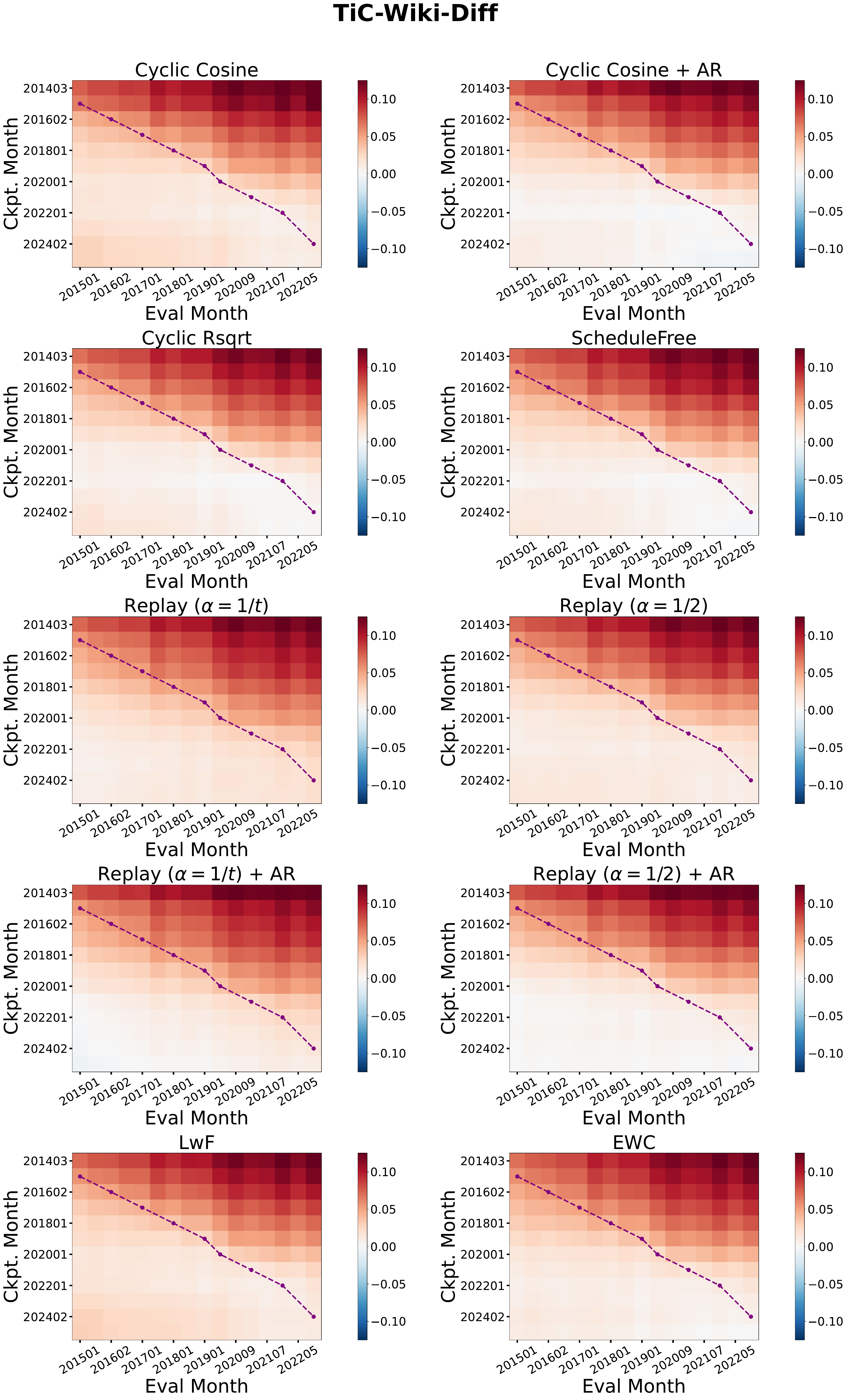}
    \caption{Evaluation matrix heatmaps for various methods on \ticwiki{}-Diff (440B token scale).}
    \label{fig:twiki_diff}
\end{figure}

\clearpage

\begin{figure}[h!]
    \centering
    \includegraphics[width=0.85\linewidth]{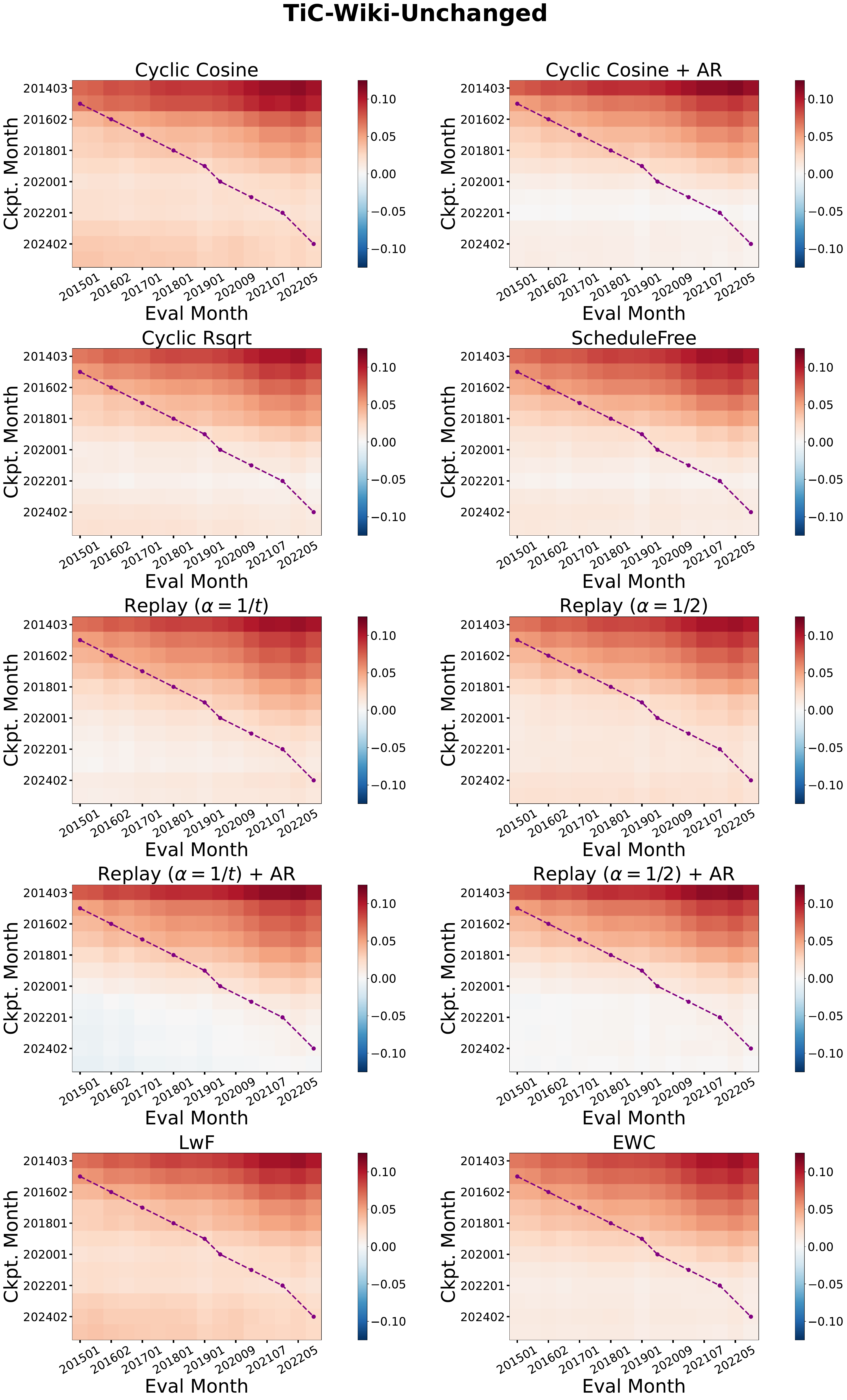}
    \caption{Evaluation matrix heatmaps for various methods on \ticwiki{}-Unchanged (440B token scale).}
    \label{fig:twiki_unchanged}
\end{figure}

\clearpage

\subsection{TiC-Stackexchange (\ticstack{})}
\begin{figure}[h!]
    \centering
    \includegraphics[width=0.85\linewidth]{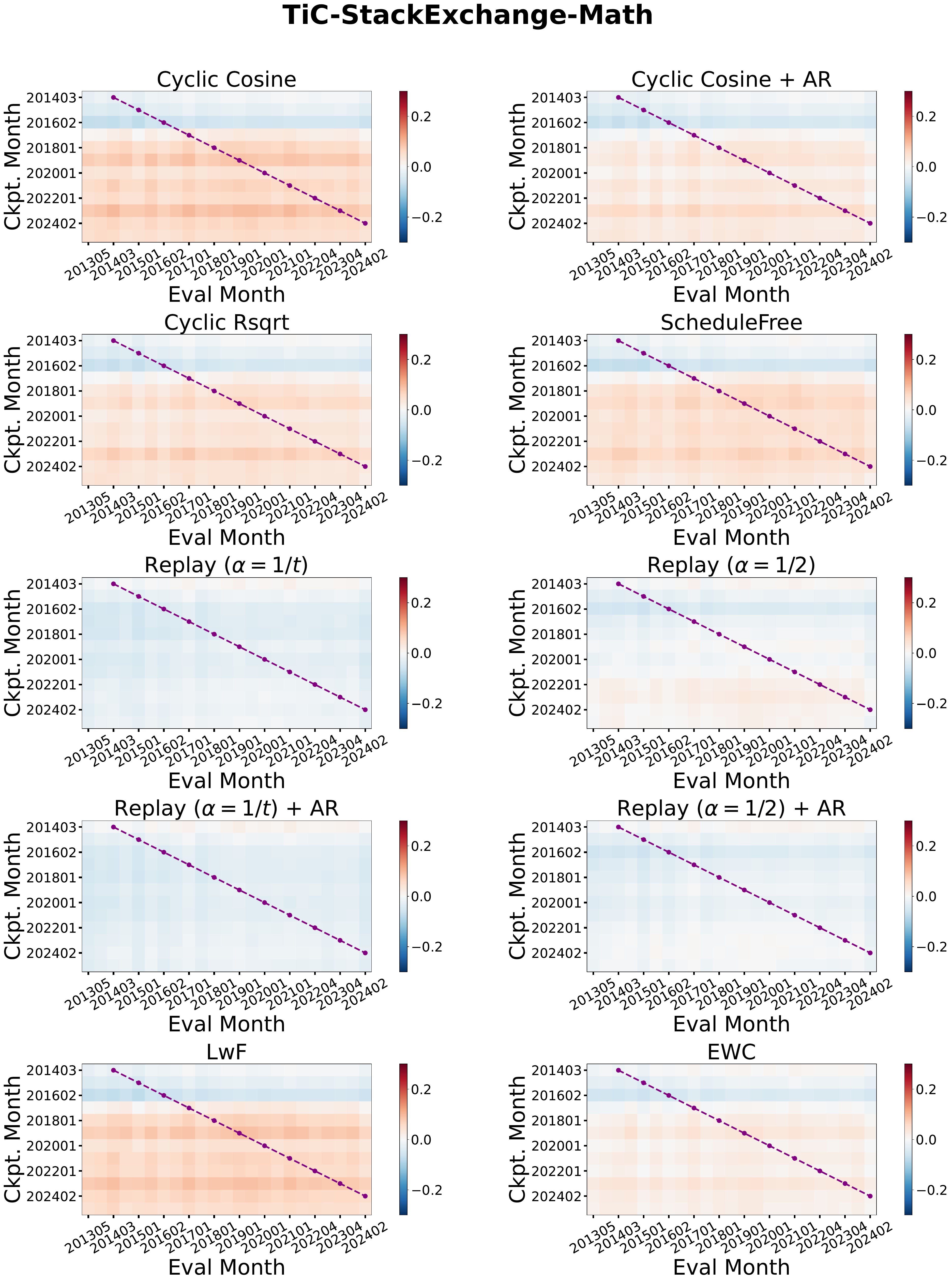}
    \caption{Evaluation matrix heatmaps for various methods on the Math site of \ticstack{}.}
    \label{fig:stackexchange_math}
\end{figure}

\begin{figure}[h!]
    \centering
    \includegraphics[width=0.85\linewidth]{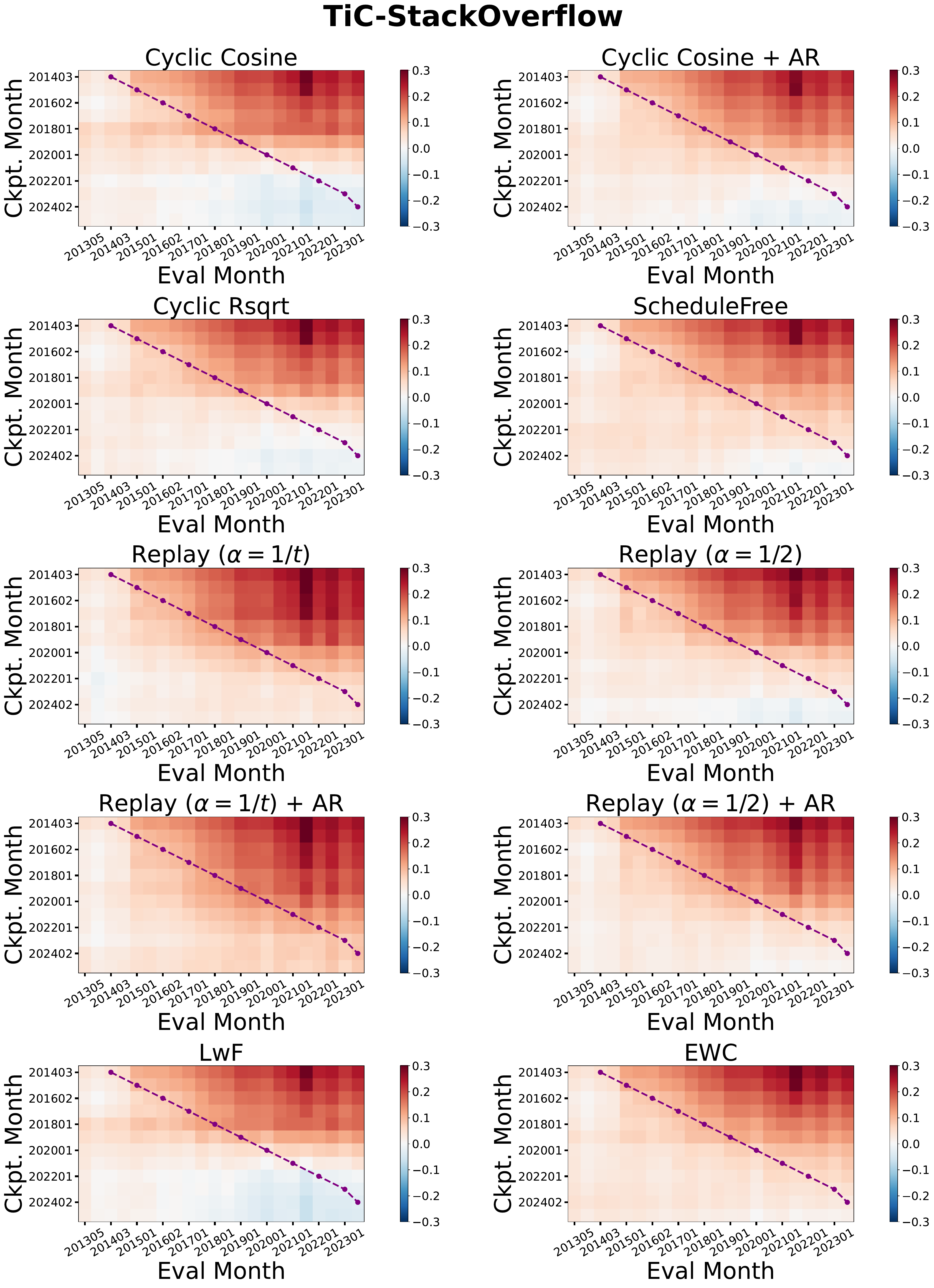}
    \caption{Heatmaps for various methods on the StackOverflow site of 
    \ticstack{} (440B token scale).}
    \label{fig:stackoverflow}
\end{figure}
\clearpage

\begin{figure}[h!]
    \centering
    \includegraphics[width=0.75\linewidth]{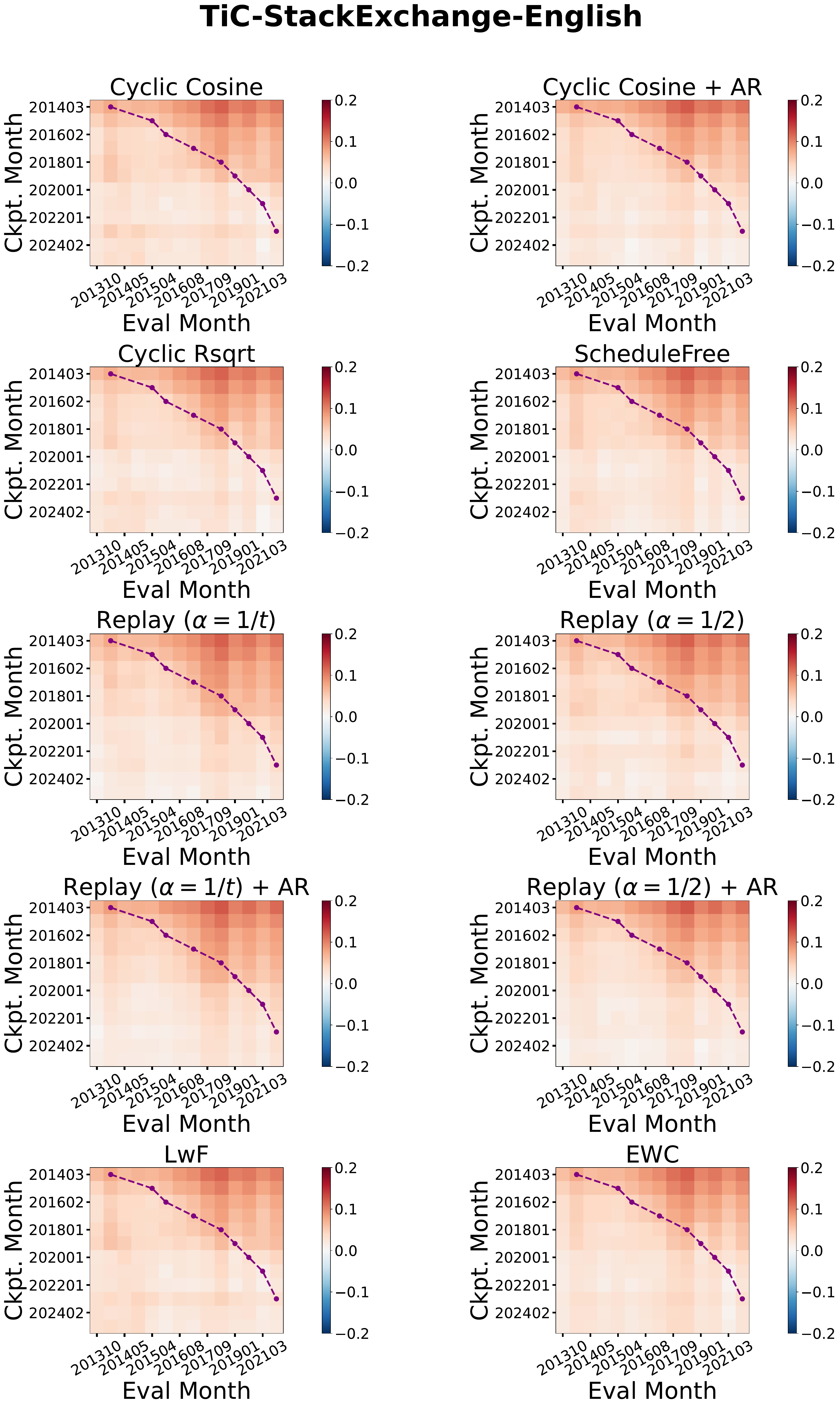}
    \caption{Heatmaps for various methods on the English site of 
    \ticstack{} (440B token scale).}
    \label{fig:stackexchange_english}
\end{figure}
\clearpage

\clearpage

\subsection{TiC-CodeDocs}

\begin{figure}[h!]
    \centering
    \includegraphics[width=0.75\linewidth]{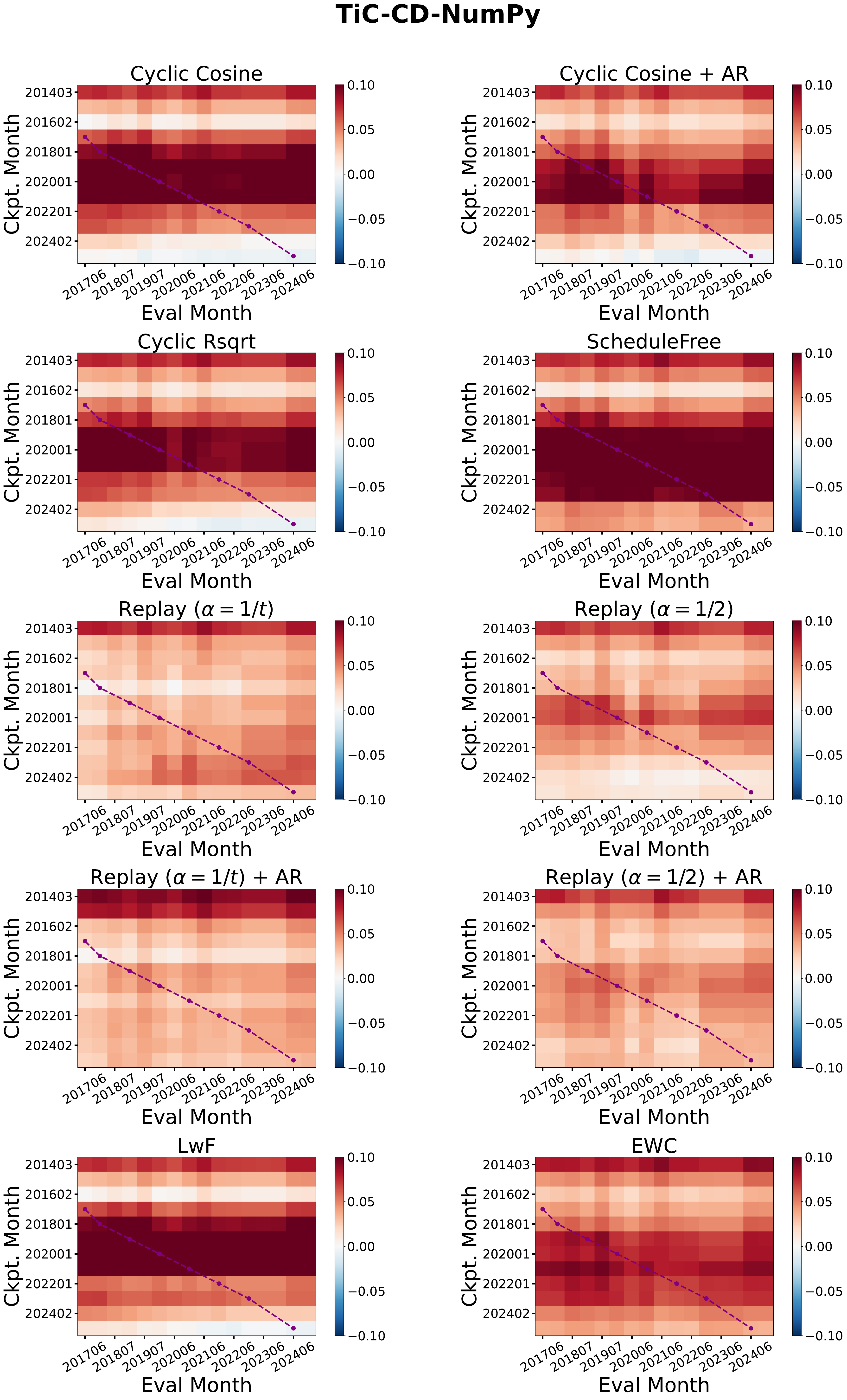}
    \caption{Heatmaps for various methods on \ticdocsnumpy{} (440B token scale).}
    \label{fig:torch_heatmaps}
\end{figure}

\begin{figure}[h!]
    \centering
    \includegraphics[width=0.75\linewidth]{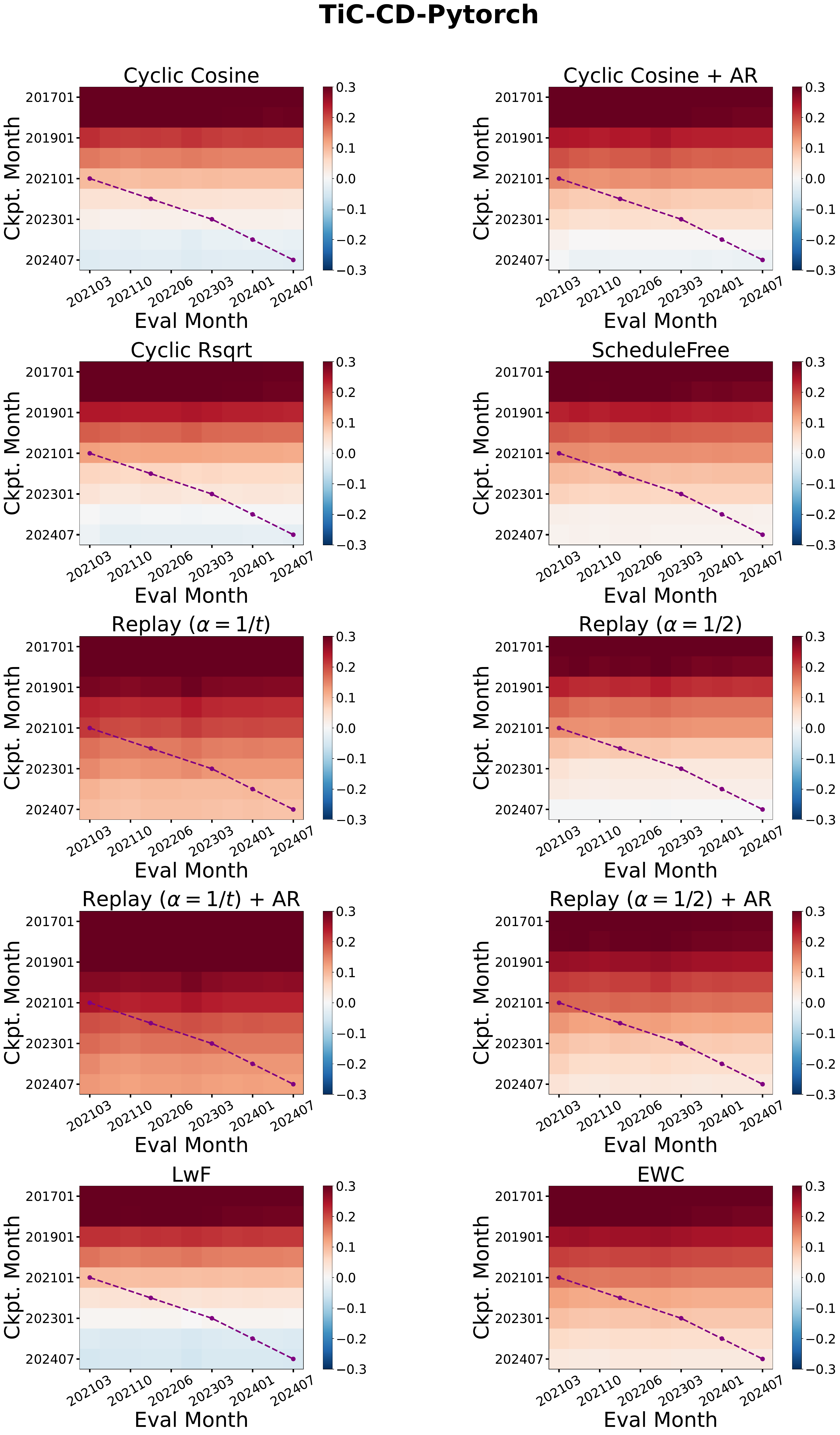}
    \caption{Heatmaps for various methods on \ticdocstorch{} (440B token scale).}
    \label{fig:numpy_heatmaps}
\end{figure}

\clearpage

\subsection{\ticstack{}-English}
\begin{table*}[h!]
    \scriptsize
    \centering
    \caption{Results for \ticstack{}-English (3B models, 440B training tokens)}
\begin{tabular}{c|c|ccc}
\toprule[1.2pt]
        \multirow{2}{*}{\textbf{Method}} & \multirow{2}{*}{\textbf{Tokens}} & \multicolumn{3}{c|}{\textbf{TiC-StackE-English $\downarrow$}} \\
        & & Bwd. & ID & Fwd. \\
        \midrule
        \multirow{2}{*}{Cyclic Cosine} & \multirow{2}{*}{440B} & 0.035 & {0.044} & \highlight{0.070} \\[-3pt] & & \stdev{(0.002)} & \stdev{(0.002)} & \stdev{(0.000)} \\[+1pt]
        \multirow{1}{*}{Cyclic Cosine + AR} & 440B & 0.028 & {\textbf{0.042}} & \highlight{0.070} \\
        \multirow{1}{*}{Cyclic Rsqrt} & 440B & 0.028 & {\textbf{0.041}} & \highlight{\textbf{0.068}}  \\
        \multirow{1}{*}{ScheduleFree} & 440B & 0.029 & {0.044} & \highlight{0.071} \\ \hline
        \multirow{1}{*}{Replay ($\alpha=1/t$)} & 440B & 0.030 & 0.050 & \highlight{0.074}  \\
        \multirow{1}{*}{Replay ($\alpha=1/2$)} & 440B & 0.030 & {0.045} & \highlight{0.072} \\
        \multirow{1}{*}{Replay ($\alpha=1/t$) + AR} & 440B & 0.030 & 0.052 & 0.075 \\
        \multirow{1}{*}{Replay ($\alpha=1/2$) + AR} & 440B & \textbf{0.026} & {0.045} & \highlight{0.070}  \\ \hline
         \multirow{1}{*}{LwF} & 440B  & 0.032 & \textbf{0.042} & 0.070  \\ 
                \multirow{1}{*}{EWC} & 440B & 0.030 & {\textbf{0.043}} & \highlight{0.069}  \\ \midrule[1pt]

        \rowcolor{lightgray!25} Oracle Series & 1.16T & 0.018 & 0.037 & 0.074 \\     

        \bottomrule[1.2pt]
    \end{tabular}
    \label{tab:stackexchange_english_440b}
    \vspace*{-9pt}
\end{table*}

\subsection{Extended replay exploration}

\begin{table}[h!]
    \scriptsize
    \centering
    \caption{\textbf{Further exploration of replay variants.}  Results are for 3B models trained for 220B total tokens.}
    \label{tab:replay_tuning}
    \vspace*{-2mm}
    \begin{tabular}{c|ccccccccc}
        \toprule[1.2pt]
        \multirow{2}{*}{Replay Variant} & \multicolumn{3}{c}{\textbf{\ticcc{}}} & \multicolumn{3}{c}{\textbf{\ticccwiki{}}} & \multicolumn{3}{c}{\textbf{ \ticccnews{} }}\\
        & Backward & ID & Forward  &  Backward & ID & Forward  & Backward & ID & Forward \\
        \midrule
        \multirow{1}{*}{Replay ($\alpha=1/t$)} & 0.023 & 0.074 & 0.178 &  0.020 & 0.036 & 0.078 & 0.005 & 0.035 & 0.117 \\
        \multirow{1}{*}{Replay (Exp)} & 0.037 & 0.038 & 0.165 & 0.030 & 0.032 & \textbf{0.074} & 0.027 & 0.018 & 0.111 \\        \multirow{1}{*}{Replay ($\alpha=0.1$)} & \textbf{0.022} & 0.066 & 0.175 &  0.021 & 0.034 & 0.077 & 0.006 & 0.030 & 0.116 \\
        \multirow{1}{*}{Replay ($\alpha=0.3$)} & \textbf{0.022} & 0.052 & 0.170 & 0.022 & 0.032 & 0.075 & 0.009 & 0.022 & 0.113 \\
        \multirow{1}{*}{Replay ($\alpha=0.5$)} & 0.024 & 0.042 & 0.167 & 0.024 & \textbf{0.031} & \textbf{0.074} & 0.013 & 0.019 & 0.111 \\
        \multirow{1}{*}{Replay ($\alpha=0.7$)} & 0.028 & 0.035 & 0.164 & 0.027 & \textbf{0.031} & \textbf{0.074} & 0.019 & 0.017 & 0.110 \\ 
        \multirow{1}{*}{Replay ($\alpha=0.9$)} & 0.039 & \textbf{0.029} & \textbf{0.162} & 0.032 & \textbf{0.031} & \textbf{0.074} & 0.030 & \textbf{0.015} & \textbf{0.109}  \\ 
        \hline
        \multirow{1}{*}{Replay ($\alpha=1/t$) + AR} & 0.026 & 0.083 & 0.181 & \textbf{0.019} & 0.037 & 0.079 & \textbf{0.004} & 0.039 & 0.119 \\
        \multirow{1}{*}{Replay (Exp) + AR} & 0.032 & 0.051 & 0.170 & 0.026 & 0.032 & 0.076 & 0.018 & 0.021 & 0.112 \\
        \multirow{1}{*}{Replay ($\alpha=0.1$) + AR} & 0.026 & 0.076 & 0.179 & \textbf{0.019 }& 0.036 & 0.079 & 0.005 & 0.035 & 0.118 \\
        \multirow{1}{*}{Replay ($\alpha=0.3$) + AR} & 0.025 & 0.064 & 0.174 & 0.020 & 0.034 & 0.077 & 0.006 & 0.027 & 0.114 \\
        \multirow{1}{*}{Replay ($\alpha=0.5$) + AR} & 0.025 & 0.055 & 0.171 & 0.022 & 0.032 & 0.076 & 0.009 & 0.022 & 0.112 \\
        \multirow{1}{*}{Replay ($\alpha=0.7$) + AR} & 0.027 & 0.048 & 0.169 & 0.023 & 0.032 & 0.075 & 0.012 & 0.020 & 0.112 \\
        \multirow{1}{*}{Replay ($\alpha=0.9$) + AR} & 0.034 & 0.043 & 0.167 & 0.026 & \textbf{0.031} & 0.075 & 0.019 & 0.018 & 0.111 \\
     \bottomrule[1.2pt]
     \end{tabular}

     \vspace{1em} %
    
    \begin{tabular}{c|cccccc}
        \toprule[1.2pt]
        \multirow{2}{*}{Replay Variant} & \multicolumn{3}{c}{\textbf{\ticwiki{}-Diff}} & \multicolumn{3}{c}{\textbf{\ticwiki{}-Unchanged}} \\
        & Backward & ID & Forward  & Backward & ID & Forward \\
        \midrule
        \multirow{1}{*}{Replay ($\alpha=1/t$)} & 0.038 & 0.057 & 0.091 & 0.035 & 0.050 & 0.074
 \\
        \multirow{1}{*}{Replay (Exp)} & 0.033 & 0.046 & 0.086 & 0.037 & 0.048 & \textbf{0.072} \\
        \multirow{1}{*}{Replay ($\alpha=0.1$)} & 0.036 & 0.054 & 0.090 & 0.035 & 0.050 & 0.075 \\
        \multirow{1}{*}{Replay ($\alpha=0.3$)} & 0.034 & 0.050 & 0.088 & 0.034 & 0.048 & 0.074  \\
        \multirow{1}{*}{Replay ($\alpha=0.5$)} & 0.034 & 0.050 & 0.088 & 0.034 & 0.048 & 0.074 \\
        \multirow{1}{*}{Replay ($\alpha=0.7$)} & \textbf{0.032} & 0.046 & \textbf{0.085} & 0.036 & 0.048 & \textbf{0.072} \\ 
        \multirow{1}{*}{Replay ($\alpha=0.9$)} & 0.033 & \textbf{0.044} & \textbf{0.085} & 0.039 & 0.048 & \textbf{0.072}
         \\ 
        \hline
        \multirow{1}{*}{Replay ($\alpha=1/t$) + AR} & 0.039 & 0.060 & 0.092 & 0.034 & 0.052 & 0.077  \\
        \multirow{1}{*}{Replay (Exp) + AR} & 0.033 & 0.048 & 0.088 & 0.034 & \textbf{0.046} & 0.073 \\
        \multirow{1}{*}{Replay ($\alpha=0.1$) + AR} & 0.038 & 0.056 & 0.091 & 0.034 & 0.049 & 0.076  \\
        \multirow{1}{*}{Replay ($\alpha=0.3$) + AR} & 0.036 & 0.052 & 0.090 & \textbf{0.033} & 0.047 & 0.075  \\
        \multirow{1}{*}{Replay ($\alpha=0.5$) + AR} & 0.034 & 0.050 & 0.088 & \textbf{0.033} & 0.047 & 0.074  \\
        \multirow{1}{*}{Replay ($\alpha=0.7$) + AR} &0.033 & 0.048 & 0.087 & \textbf{0.033} & \textbf{0.046} & 0.073  \\
        \multirow{1}{*}{Replay ($\alpha=0.9$) + AR} & 0.033 & 0.046 & 0.086 & 0.035 & 0.047 & \textbf{0.072}  \\
     \bottomrule[1.2pt]
     \end{tabular}

     \vspace{1em}

     \begin{tabular}{c|cccccc}
        \toprule[1.2pt]
        \multirow{2}{*}{Replay Variant} &  \multicolumn{3}{c}{\textbf{\ticstackoverflow{}}} & \multicolumn{3}{c}{\textbf{\ticstack{}-Math }}\\
        & Backward & ID & Forward  & Backward & ID & Forward \\
        \midrule
        \multirow{1}{*}{Replay ($\alpha=1/t$)} & 0.075 & 0.119 & 0.191 & -0.009 & -0.010 & 0.006 \\
        \multirow{1}{*}{Replay (Exp)} & 0.054 & 0.088 & 0.170 & 0.026 & 0.013 & 0.006 \\
        \multirow{1}{*}{Replay ($\alpha=0.1$)} & 0.062 & 0.104 & 0.183 & -0.007 & -0.008 & -0.004 \\
        \multirow{1}{*}{Replay ($\alpha=0.3$)} & 0.060 & 0.100 & 0.176 &  -0.001 & -0.002 & -0.001 \\
        \multirow{1}{*}{Replay ($\alpha=0.5$)} & 0.055 & 0.090 & 0.170 & 0.010 & 0.002 & 0.002 \\
        \multirow{1}{*}{Replay ($\alpha=0.7$)} & 0.050 & 0.086 & 0.169 & 0.018 & 0.009 & 0.005 \\ 
        \multirow{1}{*}{Replay ($\alpha=0.9$)} & 0.042 & \textbf{0.076} & \textbf{0.160} & 0.028 & 0.016 & 0.007
         \\ 
        \hline
        \multirow{1}{*}{Replay ($\alpha=1/t$) + AR} & 0.066 & 0.116 & 0.193 &\textbf{ -0.019} & \textbf{-0.015} & \textbf{-0.008} \\
        \multirow{1}{*}{Replay (Exp) + AR} & 0.044 & 0.088 & 0.174 & 0.005 & 0.000 & 0.000 \\
        \multirow{1}{*}{Replay ($\alpha=0.1$) + AR} & 0.057 & 0.107 & 0.188 &\textbf{ -0.018} & -0.013 & -0.006 \\
        \multirow{1}{*}{Replay ($\alpha=0.3$) + AR} & 0.052 & 0.099 & 0.179 & -0.012 & -0.010 & -0.004 \\
        \multirow{1}{*}{Replay ($\alpha=0.5$) + AR} & 0.047 & 0.091 & 0.176 & -0.006 & -0.006 & -0.002 \\
        \multirow{1}{*}{Replay ($\alpha=0.7$) + AR} & 0.042 & 0.086 & 0.170  & -0.001 & -0.003 & -0.002 \\
        \multirow{1}{*}{Replay ($\alpha=0.9$) + AR} & \textbf{0.033} & \textbf{0.076} & \textbf{0.161} & 0.005 & 0.000 & -0.001 \\
     \bottomrule[1.2pt]
     \end{tabular}

     \vspace{1em} %
    
    \begin{tabular}{c|cccccc}
        \toprule[1.2pt]
        \multirow{2}{*}{Replay Variant} & \multicolumn{3}{c}{\textbf{\ticdocsnumpy{}}} & \multicolumn{3}{c}{\textbf{\ticdocstorch{}}} \\
        & Backward & ID & Forward  &  Backward & ID & Forward \\
        \midrule
        \multirow{1}{*}{Replay ($\alpha=1/t$)} & 0.054 & 0.055 & 0.057 & 0.172 & 0.183 & 0.275 \\
        \multirow{1}{*}{Replay (Exp)} & 0.070 & 0.068 & 0.063
 & 0.069 & 0.087 & 0.233 \\
        \multirow{1}{*}{Replay ($\alpha=0.1$)} & 0.042 & 0.045 & \textbf{0.051 }& 0.163 & 0.171 & 0.268\\
        \multirow{1}{*}{Replay ($\alpha=0.3$)} & 0.056 & 0.059 & 0.057
 & 0.130 & 0.139 & 0.250 \\
        \multirow{1}{*}{Replay ($\alpha=0.5$)} & 0.058 & 0.062 & 0.060 & 0.097 & 0.110 & 0.237 \\
        \multirow{1}{*}{Replay ($\alpha=0.7$)} & 0.066 & 0.073 & 0.068 & 0.087 & 0.101 & 0.232 \\ 
        \multirow{1}{*}{Replay ($\alpha=0.9$)} & 0.069 & 0.072 & 0.064 & \textbf{0.066} & \textbf{0.083} & \textbf{0.225}
         \\ 
        \hline
        \multirow{1}{*}{Replay ($\alpha=1/t$) + AR} & 0.040 & 0.042 & 0.057 & 0.195 & 0.202 & 0.277 \\
        \multirow{1}{*}{Replay (Exp) + AR } & 0.053 & 0.052 & 0.062 & 0.099 & 0.114 & 0.240 \\
        \multirow{1}{*}{Replay ($\alpha=0.1$) + AR} & \textbf{0.031} & \textbf{0.032} & \textbf{0.049} & 0.184 & 0.190 & 0.271 \\
        \multirow{1}{*}{Replay ($\alpha=0.3$) + AR} & \textbf{0.035} & \textbf{0.036} & \textbf{0.050} & 0.155 & 0.161 & 0.254 \\
        \multirow{1}{*}{Replay ($\alpha=0.5$) + AR} & \textbf{0.035} & \textbf{0.038} & 0.052 & 0.127 & 0.138 & 0.246 \\
        \multirow{1}{*}{Replay ($\alpha=0.7$) + AR} & 0.044 & 0.046 & 0.056 & 0.111 & 0.121 & 0.238  \\
        \multirow{1}{*}{Replay ($\alpha=0.9$) + AR} & 0.048 & 0.045 & 0.054 & 0.098 & 0.112 & 0.234 \\
     \bottomrule[1.2pt]
     \end{tabular}
\end{table}

Here, we show results for additional fixed settings for $\alpha_t$ other on top of $1/t$ and $0.5$ which we focused on in the main paper. We also try another variant called Replay (Exp), which allocates 50\% of the budget to the latest month and then exponentially decreasing percentages to groups of earlier months. Specifically, we chunk all previous months into groups of 10. The most recent 10 months together receive 25\% of the overall monthly budget (each contributing 2.5\%). Each next oldest group of 10 is allocated half the budget of the previous 10 (with the exception of the last group which makes up the remainder). 

Overall, these results largely reinforce the conclusions conclusions discussed in \cref{sec:experiments}. Different evaluations benefit from more/less replay. On TiC-CC subsets, we see trade-offs between Backward and ID, where more replay helps on Backward but does worse on ID. On downstream evals, the optimal $\alpha_t$ varies. Even more replay helps more on the slower-changing domains (e.g., TiC-CodeDocs-NumPy) and less replay helps on faster moving ones (e.g., TiC-CodeDocs-PyTorch). 
Using $\alpha_t = 1/t$ is generally sub-optimal, as it is often dominated by $\alpha=0.1$ on all three metrics. Meanwhile, Replay (Exp) tends to fall somewhere between using $\alpha_t = 0.5 $ and $\alpha_t=0.9$ (as expected) but is most often dominated by simply using $\alpha_t=0.7$. Given the trade-off across evaluations, among fixed choices of $\alpha_t$, using a middling value like $0.5$ (or slightly higher) seems to be a reasonable practical recommendation to avoid performing poorly on any one evaluation/domain. 

\clearpage

\subsection{Deduplicating newer against older data}\label{app:deduplication}

In this section, we explore an alternative approach to data deduplication. As mentioned in \Cref{sec:tic_cc_data}, for all other experiments, we performed deduplication only within each month. Here, we consider the effects of also deduplicating newer months against older months. Specifically, we implement this by simply retaining the state of the Bloom Filter across months, only resetting it once every 10 months, which we refer to as a "windowed" deduplication.

\begin{table*}[h!]
    \scriptsize
    \centering
    \caption{\textbf{Exploration of windowed deduplication (3B models, 440B training tokens).} Here ``Loc.'' refers to the original within-month \textit{local} deduplication strategy and ``Wind.'' refers to our 10-month windowed deduplication.} 
    
    \vspace*{-2mm}

    \begin{tabular}{c|c|ccc|ccc|ccc|ccc}
        \toprule[1.2pt]
        \multirow{2}{*}{\textbf{Method}} & \multirow{2}{*}{\textbf{Dedup}} & \multicolumn{3}{c|}{\textbf{TiC-Wiki-Diff $\downarrow$}} & \multicolumn{3}{c|}{\textbf{TiC-Wiki-Unch. $\downarrow$}} & \multicolumn{3}{c|}{\textbf{ TiC-StackOverflow $\downarrow$}} & \multicolumn{3}{c}{\textbf{TiC-CD-PyTorch$\downarrow$}} \\ & 
        & Bwd. & ID & Fwd. & Bwd. & ID & Fwd. & Bwd. & ID & Fwd. & Bwd. & ID & Fwd.  \\
        \midrule[1pt]

        \multirow{1}{*}{Replay ($\alpha_t=1/2$)} & Loc. & 0.015 & 0.029 & 0.073 & 0.017 & 0.027 & 0.055 & \textbf{0.027} & 0.065 & 0.158 & \textbf{0.020} & \textbf{0.032} & \textbf{0.189}  \\

         \multirow{1}{*}{Replay ($\alpha_t=1/2$) + AR } & Loc. & 0.010 & 0.027 & 0.073 & 0.007 & 0.022 & 0.054 & 0.036 & 0.074 & 0.160 & 0.058 & 0.070 & 0.214 \\ \hline

        \multirow{1}{*}{Replay ($\alpha_t=1/2$)} & Wind. & 0.013 & 0.026 & 0.072 & 0.014 & 0.023 & 0.052 & 0.029 & \textbf{0.059} & \textbf{0.149 } & 0.036 & 0.048 & 0.199 \\

        \multirow{1}{*}{Replay ($\alpha_t=1/2$) + AR } & Wind. & \textbf{0.007} & \textbf{0.024} & \textbf{0.070} & \textbf{0.002} & \textbf{0.017} & \textbf{0.049} & 0.037 & 0.071 & 0.154 & 0.064 & 0.078 & 0.221
        \\ 
        
        \midrule[1pt]
        \rowcolor{lightgray!25}  \multirow{1}{*}{Oracle Series} & Loc. & 0.014 & 0.035 & 0.080 & 0.013 & 0.030 & 0.061 & 0.012 & 0.056 & 0.146 &  0.035 & 0.057 & 0.196 \\
        \bottomrule[1.2pt]
 \end{tabular}

\vspace{1em} %

\begin{tabular}{c|c|ccc|ccc|ccc}
\toprule[1.2pt]
        \multirow{2}{*}{\textbf{Method}} & \multirow{2}{*}{\textbf{Dedup}} & \multicolumn{3}{c|}{\textbf{TiC-StackE-Math $\downarrow$}} & \multicolumn{3}{c|}{\textbf{ TiC-CD-NumPy $\downarrow$}} & \multicolumn{1}{c}{\textbf{Static Evals. $\uparrow$}} \\
        & & Bwd. & ID & Fwd. & Bwd. & ID & Fwd. & \staticevals{} (DCLM)\\
        \midrule

        \multirow{1}{*}{Replay ($\alpha_t=1/2$)} & Loc. & 0.000 & -0.008 & -0.011 & \textbf{0.029} & \textbf{0.038 }& 0.044 & 50.1 \\

        \multirow{1}{*}{Replay ($\alpha_t=1/2$) + AR} & Loc. & \textbf{-0.015} & \textbf{-0.019} & \textbf{-0.017} & 0.036 & 0.039 & \textbf{0.043} & 49.2 \\ \hline 

        \multirow{1}{*}{Replay ($\alpha_t=1/2$) } & Wind. & 0.011 & 0.002 & -0.001 & 0.085 & 0.088 & 0.086 & 49.8 \\

        \multirow{1}{*}{Replay $\alpha_t=1/2$) + AR} & Wind. & -0.002 & -0.009 & -0.008 & 0.078 & 0.078 & 0.078 & 49.5 \\  \midrule[1pt]

        \rowcolor{lightgray!25} Oracle Series & Loc. & -0.025 & -0.028 & -0.022 & 0.008 & 0.008 & 0.015 & 50.6  \\ 
        \bottomrule[1.2pt]
    \end{tabular}

    \label{tab:dedup-exploration}
    \vspace*{-9pt}
\end{table*}

In \Cref{tab:dedup-exploration}, we apply the new deduplication strategies and measure performance on downstream evaluations only (as \ticcc{} evaluations will be biased towards the original within-month deduplication as it was used to construct \ticcc{} test sets). Overall, we observe that deduplicating across months, when applied with replay, can help improve performance on some evaluations but not all. We speculate that this is because on rarer domains, such as code, deduplicating more aggressively across months can reduce the already limited amount of relevant samples. On the other hand, for evaluations where more general data helps, such as \ticwiki{}, deduplication helps by letting the model see more diverse data given the same token budget. 

\clearpage

\subsection{Effects of model size on downstream evaluations.}\label{app:model_size}

\begin{table*}[h!]
    \scriptsize
    \centering
    \caption{\textbf{Comparing downstream evaluations for 1B and 3B models (220B training tokens).} For all dynamic evaluations, we report perplexity values relative to the \textit{Oracle-2024-07} model of the same size and with log-scaling. Meanwhile, \staticevals is an average of the accuracies of 22 downstream zero/few-shot tasks used by \citet{li2024datacomp}, evaluated only on the final model checkpoint (without scaling by an Oracle). \textbf{Bold} values are within one standard deviation (estimated with 3 runs of Cyclic Cosine) of the best in each column for a given model size. %
    }
    \vspace*{-2mm}

    \begin{tabular}{c|c|ccc|ccc|ccc|ccc}
        \toprule[1.2pt]
        \multirow{2}{*}{\textbf{Method}} & \multirow{2}{*}{\textbf{Params}} & \multicolumn{3}{c|}{\textbf{TiC-Wiki-Diff $\downarrow$}} & \multicolumn{3}{c|}{\textbf{TiC-Wiki-Unch. $\downarrow$}} & \multicolumn{3}{c|}{\textbf{ TiC-StackOverflow $\downarrow$}} & \multicolumn{3}{c}{\textbf{TiC-CD-PyTorch$\downarrow$}} \\ & 
        & Bwd. & ID & Fwd. & Bwd. & ID & Fwd. & Bwd. & ID & Fwd. & Bwd. & ID & Fwd.  \\
        
        \midrule[1pt]
        \multirow{2}{*}{Cyclic Cosine \stdev{(std)}} & \multirow{2}{*}{3B} & 0.033 & 0.045 & 0.085 & 0.039 & 0.048 & 0.072 & 0.041 & 0.073 & \textbf{0.156} & \textbf{0.057} & \textbf{0.072} & 0.219 \\[-3pt] & &  \stdev{(0.000)} & \stdev{(0.000)} & \stdev{(0.000)} & \stdev{(0.000)} &  \stdev{(0.000)} & \stdev{(0.000)} & \stdev{(0.002)} &  \stdev{(0.002)} & \stdev{(0.003)} & \stdev{(0.002)} & \stdev{(0.002)} & \stdev{(0.002)}  \\[+1pt]

        \multirow{1}{*}{Cyclic Cosine + AR} & 3B & 0.033 & 0.048 & 0.087 & 0.035 & 0.047 & 0.074 & \textbf{0.032} & \textbf{0.072} & 0.159 & 0.081 & 0.096 & 0.228  \\

        \multirow{1}{*}{Cyclic Rsqrt} & 3B & \textbf{0.031} & \textbf{0.043} & 0.084 & 0.035 & \textbf{0.045} & 0.070 & 0.035 & \textbf{0.071} & \textbf{0.158} & 0.059 & 0.074 & 0.220 \\

        \multirow{1}{*}{Schedule-Free} & 3B & 0.035 & 0.048 & 0.087 & 0.040 & 0.050 & 0.074 & 0.038 & 0.074 & 0.160 & 0.081 & 0.096 & 0.236 \\ \hline

        \multirow{1}{*}{Replay ($\alpha_t=1/t$)} & 3B & 0.038 & 0.057 & 0.091 & 0.035 & 0.050 & 0.074 & 0.075 & 0.119 & 0.191 & 0.172 & 0.183 & 0.275 \\

        \multirow{1}{*}{Replay ($\alpha_t=1/2$)} & 3B & 0.032 & 0.047 & 0.086 & 0.034 & 0.047 & 0.072 & 0.055 & 0.090 & 0.170 & 0.097 & 0.110 & 0.237 \\

        \multirow{1}{*}{Replay ($\alpha_t=1/t$) + AR} & 3B & 0.039 & 0.060 & 0.092 & 0.034 & 0.052 & 0.077 & 0.066 & 0.116 & 0.193 & 0.195 & 0.202 & 0.277 \\

        \multirow{1}{*}{Replay ($\alpha_t=1/2$) + AR} & 3B & 0.034 & 0.050 & 0.088 & \textbf{0.033} & 0.047 & 0.074 & 0.047 & 0.091 & 0.176 & 0.127 & 0.138 & 0.246  \\ \hline

        \multirow{1}{*}{LwF} & 3B & 0.033 & 0.045 & 0.085 & 0.039 & 0.048 & 0.072 & 0.037 & \textbf{0.070} & \textbf{0.155} & \textbf{0.055} & \textbf{0.070} & \textbf{0.216} \\

        \multirow{1}{*}{EWC} & 3B & \textbf{0.031} & 0.044 & \textbf{0.083} & 0.034 & \textbf{0.045} & \textbf{0.069} & \textbf{0.033} & \textbf{0.072} & 0.162 & 0.067 & 0.080 & 0.222 \\ \midrule[1pt]

        \rowcolor{lightgray!25} \multirow{1}{*}{Oracle Series} & 3B & 0.014 & 0.035 & 0.080 & 0.013 & 0.030 & 0.061 & 0.012 & 0.056 & 0.146 &  0.035 & 0.057 & 0.196
        
 \\ \bottomrule[1.2pt]
 \end{tabular}

\vspace{1em} %

\begin{tabular}{c|c|ccc|ccc|ccc}
\toprule[1.2pt]
        \multirow{2}{*}{\textbf{Method}} & \multirow{2}{*}{\textbf{Params}} & \multicolumn{3}{c|}{\textbf{TiC-StackE-Math $\downarrow$}} & \multicolumn{3}{c|}{\textbf{ TiC-CD-NumPy $\downarrow$}} & \multicolumn{1}{c}{\textbf{Static Evals. $\uparrow$}} \\
        & & Bwd. & ID & Fwd. & Bwd. & ID & Fwd. & \staticevals{} (DCLM)\\
        \midrule

        \multirow{2}{*}{Cyclic Cosine \stdev{(std)}} &   \multirow{2}{*}{3B} & 0.037 & 0.024 & 0.015 & 0.070 & 0.075 & 0.070  & 48.5 \\[-3pt] & & \stdev{(0.001)} & \stdev{(0.001)} & \stdev{(0.000)} & \stdev{(0.004)} & \stdev{(0.006)} & \stdev{(0.002)}  & \stdev{(0.4)} \\[+1pt]

        \multirow{1}{*}{Cyclic Cosine + AR} & 3B & 0.011 & 0.006 & 0.003 & 0.055 & 0.056 & 0.062  & 48.5  \\

        \multirow{1}{*}{Cyclic Rsqrt} & 3B & 0.024 & 0.016 & 0.009 & 0.067 & 0.072 & 0.071 & \textbf{49.0}\\

        \multirow{1}{*}{Schedule-Free} & 3B & 0.035 & 0.025 & 0.017 & 0.070 & 0.075 & 0.080 & \textbf{48.8} \\
        \hline

        \multirow{1}{*}{Replay ($\alpha_t=1/t$)} & 3B & -0.009 & -0.010 & -0.006 & 0.054 & 0.055 & 0.057 & \textbf{48.9} \\ 

        \multirow{1}{*}{Replay ($\alpha_t=1/2$)} & 3B & 0.010 & 0.002 & 0.002 & 0.058 & 0.062 & 0.060 & \textbf{49.0}\\

        \multirow{1}{*}{Replay ($\alpha_t=1/t$) + AR} & 3B & \textbf{-0.019} & \textbf{-0.015} & \textbf{-0.008} & 0.040 & \textbf{0.042} & 0.057 & \textbf{49.0} \\

        \multirow{1}{*}{Replay ($\alpha_t=1/2$) + AR} & 3B & -0.006 & -0.006 & -0.002 & \textbf{0.035} & \textbf{0.038} & \textbf{0.052} & \textbf{49.2}\\ \hline

        \multirow{1}{*}{LwF} & 3B & 0.034 & 0.022 & 0.013 & 0.073 & 0.078 & 0.073  & 48.5  \\ 

        \multirow{1}{*}{EWC} & 3B & 0.016 & 0.010 & 0.006 & 0.056 & 0.060 & 0.067 & \textbf{48.9} \\ \midrule[1pt]

        \rowcolor{lightgray!25} Oracle Series & 3B & -0.025 & -0.028 & -0.022 & 0.008 & 0.008 & 0.015 & 50.6  \\     
        \bottomrule[1.2pt]
    \end{tabular}

\vspace{1em} %

    \begin{tabular}{c|c|ccc|ccc|ccc|ccc}
        \toprule[1.2pt]
        \multirow{2}{*}{\textbf{Method}} & \multirow{2}{*}{\textbf{Params}} & \multicolumn{3}{c|}{\textbf{TiC-Wiki-Diff $\downarrow$}} & \multicolumn{3}{c|}{\textbf{TiC-Wiki-Unch. $\downarrow$}} & \multicolumn{3}{c|}{\textbf{ TiC-StackOverflow $\downarrow$}} & \multicolumn{3}{c}{\textbf{TiC-CD-PyTorch$\downarrow$}} \\ & 
        & Bwd. & ID & Fwd. & Bwd. & ID & Fwd. & Bwd. & ID & Fwd. & Bwd. & ID & Fwd.  \\
        \midrule[1pt]

        \multirow{2}{*}{Cyclic Cosine \stdev{(std)}} & \multirow{2}{*}{1B} &  0.035 & 0.044 & 0.079 & 0.040 & 0.045 & 0.064 & \textbf{0.074} & \textbf{0.113} & \textbf{0.211} & \textbf{0.061} & \textbf{0.079} & \textbf{0.232} \\[-3pt] & & \stdev{(0.000)} & \stdev{(0.000)} & \stdev{(0.000)} & \stdev{(0.000)} & \stdev{(0.000)} & \stdev{(0.000)} & \stdev{(0.001)} & \stdev{(0.001)} & \stdev{(0.001)} & \stdev{(0.004)} & \stdev{(0.004)} & \stdev{(0.004)}  \\[+1pt]

        \multirow{1}{*}{Cyclic Cosine + AR} & 1B & 0.035 & 0.045 & 0.081 & 0.035 & 0.043 & 0.065 & 0.089 & 0.135 & 0.222 & 0.104 & 0.120 & 0.252  \\

        \multirow{1}{*}{Cyclic Rsqrt} & 1B & 0.034 & 0.044 & 0.079 & 0.037 & 0.044 & 0.063 & 0.084 & 0.126 & 0.219 & 0.082 & 0.099 & 0.245 \\

        \multirow{1}{*}{ScheduleFree} & 1B & 0.040 & 0.049 & 0.083 & 0.044 & 0.050 & 0.068 & 0.095 & 0.138 & 0.232 & 0.111 & 0.127 & 0.262 \\ \hline

        \multirow{1}{*}{Replay ($\alpha=1/t$)} & 1B & 0.038 & 0.054 & 0.085 & 0.034 & 0.046 & 0.066 & 0.100 & 0.152 & 0.236 & 0.185 & 0.196 & 0.284 \\

        \multirow{1}{*}{Replay ($\alpha=1/2$)} & 1B & 0.036 & 0.048 & 0.081 & 0.036 & 0.045 & 0.064 & 0.089 & 0.129 & 0.218 & 0.115 & 0.129 & 0.248 \\

        \multirow{1}{*}{Replay ($\alpha=1/t$) + AR} & 1B &  0.039 & 0.055 & 0.085 & \textbf{0.032} & 0.044 & 0.066 & 0.113 & 0.167 & 0.245 & 0.214 & 0.225 & 0.299 \\

        \multirow{1}{*}{Replay ($\alpha=1/2$) + AR} & 1B &  0.036 & 0.049 & 0.083 & \textbf{0.032} & 0.043 & 0.064 & 0.097 & 0.145 & 0.231 & 0.152 & 0.165 & 0.272 \\ \hline

        \multirow{1}{*}{LwF} & 1B &   0.036 & 0.045 & 0.080 & 0.041 & 0.046 & 0.065 & 0.076 & 0.119 & 0.213 & \textbf{0.065} & \textbf{0.083} & 0.237 \\

        \multirow{1}{*}{EWC} & 1B  &\textbf{ 0.033} & \textbf{0.043} & \textbf{0.078} & 0.035 & \textbf{0.042} & \textbf{0.061} & 0.086 & 0.128 & 0.220 & 0.097 & 0.114 & 0.249
        \\ 
        \midrule[1pt]
        
        \rowcolor{lightgray!25} \multirow{1}{*}{Oracle Series} & 1B & 0.015 & 0.043 & 0.073 & 0.016 & 0.040 & 0.058 & 0.032 & 0.089 & 0.176 & 0.040 & 0.000 & 0.201  \\ 
        
        \bottomrule[1.2pt]
 \end{tabular}

\vspace{1em} %

\begin{tabular}{c|c|ccc|ccc|ccc|ccc}
\toprule[1.2pt]
        \multirow{2}{*}{\textbf{Method}} & \multirow{2}{*}{\textbf{Params}} & \multicolumn{3}{c|}{\textbf{TiC-StackE-Math $\downarrow$}} & \multicolumn{3}{c|}{\textbf{ TiC-CD-NumPy $\downarrow$}} & \multicolumn{1}{c}{\textbf{Static Evals. $\uparrow$}} \\
        & & Bwd. & ID & Fwd. & Bwd. & ID & Fwd. & \staticevals{} (DCLM)\\
        \midrule

        \multirow{2}{*}{Cyclic Cosine} & \multirow{2}{*}{1B} & 0.063 & 0.051 & 0.034 & 0.097 & 0.115 & 0.123 &  45.3 \\[-3pt] & &  \stdev{(0.001)} & \stdev{(0.001)} & \stdev{(0.001)} & \stdev{(0.003)} & \stdev{(0.004)} & \stdev{(0.006)} & \stdev{(0.003)} \\[+1pt]

        \multirow{1}{*}{Cyclic Cosine + AR} & 1B & 0.042 & 0.033 & 0.023 & 0.093 & 0.112 & 0.126 & \textbf{45.6} \\

        \multirow{1}{*}{Cyclic Rsqrt} & 1B & 0.054 & 0.043 & 0.029 &  0.096 & 0.115 & 0.127  & \textbf{45.7} \\

        \multirow{1}{*}{ScheduleFree} & 1B &  0.067 & 0.055 & 0.041 & 0.112 & 0.132 & 0.142  & \textbf{45.6} \\ \hline

        \multirow{1}{*}{Replay ($\alpha=1/t$)} & 1B & 0.013 & 0.012 & 0.011 & \textbf{0.082} & \textbf{0.095} & \textbf{0.110} & \textbf{45.6} \\

        \multirow{1}{*}{Replay ($\alpha=1/2$)} & 1B & 0.032 & 0.024 & 0.018 & \textbf{0.084} & \textbf{0.097} & \textbf{0.110} & \textbf{45.6 }\\

        \multirow{1}{*}{Replay ($\alpha=1/t$) + AR} & 1B & \textbf{0.008} & \textbf{0.010} & \textbf{0.011} & 0.090 & 0.103 & \textbf{0.116}  & \textbf{45.6} \\

        \multirow{1}{*}{Replay ($\alpha=1/2$) + AR} & 1B & 0.021 & 0.016 & 0.015  & \textbf{0.082} & \textbf{0.099} & \textbf{0.114} & \textbf{45.9} \\ \hline

        \multirow{1}{*}{LwF} & 1B  & 0.062 & 0.049 & 0.034 & 0.096 & 0.119 & 0.130 & \textbf{45.6} \\

        \multirow{1}{*}{EWC} & 1B &  0.041 & 0.033 & 0.023 &  0.107 & 0.123 & 0.128 & \textbf{45.6} \\ \midrule[1pt]

\rowcolor{lightgray!25} Oracle Series & 1B & -0.012 & -0.017 & -0.016 & 0.016 & 0.025 & 0.037 & 46.7  \\     
        \bottomrule[1.2pt]
    \end{tabular}
    \label{tab:1b-downstream-evals}
    \vspace*{-9pt}
\end{table*}

As seen in \Cref{tab:1b-downstream-evals} above, for the same token budget, changing model size can lead to continual methods being more suboptimal compared to \textit{Oracle-2024-07} on all evaluations besides \ticwiki{}. However, the overall conclusions do not majorly change (e.g., which evaluations benefit from replay versus not). 

%% file: sec/sup_related.tex
\section{Extended Related Work}\label{sec:related_sup}

\textbf{Temporal knowledge evaluations.}
Various 
benchmarks have been proposed to evaluate the temporal knowledge of LLMs.
TemporalWiki~\citep{jang2022temporalwiki} evaluates the capability of models to 
update factual knowledge.  TemporalWiki is constructed from the difference 
between four consecutive snapshots of Wikipedia and Wikidata. Our \ticwiki{} 
evaluation expands and improves on TemporalWiki in various ways (see 
\cref{sec:tic_wiki_sup}).
StreamingQA~\citep{liska2022streamingqa} consists of human written and 
generated questions from 14 years of news articles. The evaluation is either 
open-book where a model receives a collection of news articles that contain the 
answer, or closed-book where the model is first fine-tuned on the training set 
containing the documents and then tested. 
TempEL~\citep{zaporojets2022tempel} evaluates entity linking performance across 
{10} yearly snapshots of Wikipedia. Entity linking is the task of mapping 
anchor mentions to target entities that describe them in a knowledge base.  In 
comparison, our \ticwiki{} evaluates general language and knowledge 
understanding.
TempLAMA~\citep{dhingra2022time} constructs an evaluation for factual queries 
from Wikidata. They focus on temporally sensitive knowledge with known start 
and end dates in a specific Wikidata snapshot. Notably, they propose TempoT5 to 
jointly model text and timestamp which allows a language model to answer 
temporal questions that change over time such ``Who is the president''.
EvolvingQA~\citep{kim2023carpe} is also a benchmark for training and evaluating 
on Wikipedia over time where a LLM automatically generates question-answers 
from 6 months of articles in 2023. We avoid using any LLMs for generating our 
evaluations to prevent any transfer of biases.
TIQ~\citep{jia2024tiq} benchmark consists of 10k questions-answers based on 
significant events for the years 1801--2025.

\textbf{Temporal generalization.}
Beyond understanding the past, LLMs need to be prepared for the future.
\citet{li2024evaluating} observes performance deterioration of public LLMs on 
Wikipedia, news, code documentation, and arXiv papers after their training data cutoff date.  
They particularly use compression rate achieved by treating an LLM as a general 
input compressor using arithmetic coding~\citep{deletang2023language}.
Our comprehensive evaluations on CommonCrawl, Wikipedia, news articles, 
StackExchange, and code documentation evaluations verify their results and more 
comprehensively show that the rate of deterioration is domain-specific.
DyKnow~\citep{mousavi2024your} evaluations also reaffirm the finding that private and 
open-source LLMs have outdated knowledge by asking them questions constructed 
using Wikidata. They also observe LLMs output inconsistent answers in response 
to prompt variations and current knowledge editing methods do not reduce 
outdatedness.
TAQA~\citep{zhao2024set} further demonstrated that pretrained LLMs mostly answer 
questions using knowledge from years before their pretraining cutoff. They 
construct question/answers  from Wikipedia for years 2000--2023 and propose 
three methods to improve the temporal alignment of models.
Similar observations have been made in RealTimeQA~\citep{kasai2024realtime} and 
TempUN~\citep{beniwal2024remember}.
These works further solidify the need for continuously updating models with 
continual pretraining.

\textbf{Temporal understanding.}
General temporal understanding involves reasoning based on the relation between 
existing knowledge.
Test of Time benchmark~\citep{fatemi2024test} evaluates temporal reasoning, 
logic, and arithmetic by constructing a synthetic dataset. Their goal is to 
reduce the chance of factual inconsistency in the evaluation using synthetic 
data. Our benchmark is designed to be fully realistic based on real data and 
timestamps to understand the challenges of large-scale continual pretraining in 
practice.
\citet{gurnee2023language} find that LLMs learn a representation of space and 
time with individual neurons that encode spatial and temporal coordinates. They 
construct datasets of named entities and find that linear probing LLMs performs 
well on predicting spatial and temporal coordinates.
\citet{nylund2023timeencoded} proposed time vectors that specify a direction in 
the model's weight space that improve performance on text from a specific time 
period.

\textbf{Temporal domain-specific evaluations.}
We can further analyze the temporal understanding of a model based on the 
performance on specific domains with varying rates of change.
\citet{luu2021time} studied temporal misalignment such as quantifying temporal 
degradation of domain-specific finetuning in four domains: social media, 
science, news, and food reviews. They observed significant temporal degradation 
in domains such as news, social media, and science but less in food reviews.
\citet{gururangan2020dont} studied domain-adaptive pretraining and 
task-adaptive pretraining on unlabeled data for four domains in science, news, 
and reviews.  They observe domain/task-adaptive pretraining improves 
performance on the new domain but do not evaluate forgetting on previous 
domains.
\citet{agarwal2022temporal} studies the temporal model deterioration on future 
evaluations. They find that the deterioration is task-dependent and 
domain-adaptive pretraining does not help hypothesizing that limited 
pretraining data is detrimental in continual pretraining.
\citet{jin2022lifelong} domain-incremental pretraining for four scientific 
domains as well as temporal pretraining on social media over {6} years.  They 
focus on the impact on downstream performance after fine-tuning. They observe 
distillation-based approaches are the most effective in retaining dowstream 
performance for tasks related to earlier domains. Overall, the gap between 
different continual learning methods remained small that can be due to the 
small scale of pretraining. In comparison, our \ticcc{} training is simulating 
large-scale pretraining.

\textbf{Domain/task-continual learning for LLMs.}
In domain/task continual learning, the model is presented with a sequence of 
tasks with predefined labels~\citep{hsu2018re,van2019three, zhou2023pycil}. 
Each task comes with its training and test sets.  In contrast with continual 
pretraining, the model needs to support a growing set of labels while compared 
with temporal continual learning, the order of tasks are often arbitrary (e.g., 
Split-CIFAR, Perm-MNIST).
Prominent methods in this domain are regularization-based 
methods~\citep{kirkpatrick2017overcoming,
mirzadeh2020dropout,
mirzadeh2020understanding,farajtabar2020orthogonal},
replay-based methods that often perform 
superior~\citep{lomonaco2022cvpr,balaji2020effectiveness,prabhu2020gdumb},
and architecture-based methods that adapt the model over 
time~\citep{schwarz2018progress,rusu2016progressive}.
Continual learning for language models has also been dominated by domain/task 
continual works.
\citet{jin2022lifelong} proposed benchmarks for continually training models on 
a sequence of research paper domains as well as chronologically-ordered tweet 
streams.
\citet{razdaibiedina2023progressive} proposed learning a new soft prompt for 
each task and pass soft prompts for all seen tasks to the model which provides 
adaptability while preventing catastrophic forgetting.
\citet{luo2023empirical} studied continual learning for instruction tuning and 
observed catastrophic forgetting, especially for larger models.
\citet{mehta2023empirical} showed that generic pretraining implicitly reduces 
catastrophic forgetting during task incremental finetuning.

\textbf{Continual pretraining of LLMs.}
Recent work have studied continual pretraining of foundation models at 
large-scale.
TiC-CLIP~\citep{garg2024tic} proposed a benchmark of training and evaluation of 
image-text foundation models and demonstrated the deterioration of existing 
foundation models on new data.
\citet{lazaridou2021mind} studied time-stratified language pretraining on WMT, 
news, and arXiv up to 2019 and observed the models become outdated quickly on 
news data that holds even for models of various sizes.
They study dynamic evaluation as a continual pretraining method that trains on 
a stream of chronologically ordered documents and observed that models can be 
updated. However, they did not explore the impact on forgetting and scalability 
of the method to more generic pretraining data over years.
\citet{jang2021towards} proposed continual knowledge learning as a new problem 
and suggested that parameter expansion is necessary to retain and learn 
knowledge. They focus on one-step continual pretraining where models are 
pretrained on C4/Wikipedia data up to 2020 and then trained once more on recent 
news articles. They find adapter methods perform better than regularization and 
replay methods. Adapter methods are not directly applicable in our multi-year 
continual pretraining setup where we train in more than 100 steps on 
large-scale data.
\citet{gupta2023continual} proposed simple recipes for continual pretraining of 
LLMs such as utilizing cyclical learning rate schedules with warmup and ablated 
on hyperparameters such as warmup duration when continuing the pretraining on 
a fixed pair of pretraining datasets.

\textbf{Time-aware training.}
Orthogonal to continual pretraining, one can modify the training or fine-tuning 
of a model to include explicit information about time.
TempLAMA~\citep{dhingra2022time} proposed prepending a time prefix to each example 
during training which gives the model the flexibility to respond to 
time-sensitive questions. They train models on news articles where the time can 
be reliably extracted.
\citet{drinkall2024time} proposed training a series of models with sequential 
data cutoffs dates to avoid data contamination with benchmark and private data.  
The observe no difference across time on static downstream evaluations when 
training models on news and Wikipedia

\textbf{Factual editing and retrieval augmented generation (RAG).}
Another set of works aim to address the staleness of pretrained LLMs without 
further standard pretraining. One approach is to surgically edit the facts 
a model ``knows'' by identifying and updating the relevant weights within 
a model~\citep{mitchell2021fast}. Another is to store edits in an explicit 
memory and learn to reason over them~\citep{mitchell2022memory}.
Retrieval augmented generation (RAG) pairs an LLM with new data sources to
retrieve the most relevant document for a query.
Generally, continual pretraining and RAG are orthogonal approaches to generate 
up to date responses. RAG methods increase the cost at inference time without 
changing the model while continual pretraining is the opposite.
FreshLLMs~\citep{vu2023freshllms} proposes a QA benchmark and argues that 
fast-changing knowledge requires a retrieval-based solution compared with 
slow-changing knowledge.
Continual pretraining can be crucial in reducing the cost of RAG by utilizing 
retrieval only on knowledge that changes faster than the rate of continual 
pretraining.